\documentclass{article}
\usepackage[preprint]{neurips_2026}
\makeatletter
\renewcommand{\@notice}{}
\makeatother

\usepackage[utf8]{inputenc}
\usepackage[T1]{fontenc}
\usepackage{hyperref}
\usepackage{url}
\usepackage{booktabs}
\usepackage{amsfonts}
\usepackage{nicefrac}
\usepackage{microtype}
\usepackage[table]{xcolor}

\definecolor{darkblue}{rgb}{0, 0, 0.5}
\hypersetup{colorlinks=true, citecolor=darkblue, linkcolor=darkblue, urlcolor=darkblue}
\usepackage{caption}
\usepackage{amsmath}
\usepackage{array}
\usepackage{rotating}
\usepackage{needspace}
\newcommand{\ie}{\emph{i.e.}}
\newcommand{\eg}{\emph{e.g.}}

\usepackage{multirow}
\usepackage{enumitem}
\setlist[itemize]{label=\textbullet}
\usepackage{algpseudocode}
\usepackage{float}
\usepackage{makecell}%
\definecolor{skipgreen}{RGB}{51,116,35}
\definecolor{casbg}{RGB}{242,252,254}%
\usepackage{graphicx}
\usepackage{subcaption}
\usepackage{wrapfig}%
\usepackage[percent]{overpic}

\usepackage{dsfont}
\usepackage{tabularx}
\newcommand{\cots}{\texttt{CoTS}}
\newcommand{\ecots}{\texttt{E-CoTS}}
\usepackage[ruled,vlined]{algorithm2e}

\setlist[itemize]{left=10pt}

\newlength{\arrowwidth}
\title{Bridging the Confidence Gap: Temperature Scaling for Calibrating Test-Time Prompt Tuning}
\author{
Yuwei Liang$^{1,2}$ \quad
Jian Liang$^{1,2}$\thanks{To whom correspondence should be addressed.} \quad
Dapeng Hu$^{3}$ \quad
Yinuo Xu$^{1,2}$ \quad
Ran He$^{1,2}$
\\
$^{1}$ School of Artificial Intelligence, University of Chinese Academy of Sciences, Beijing, China\\
$^{2}$ NLPR \& MAIS, Institute of Automation, Chinese Academy of Sciences, Beijing, China\\
$^{3}$ Independent Researcher\\
\texttt{liangyuwei911@gmail.com, liangjian92@gmail.com}
}

\begin{document}

\maketitle

\begin{abstract}
Test-time prompt tuning (TPT) enables adaptation on a single test instance, achieving improved accuracy but often sacrificing calibration performance.
Most existing calibration methods introduce additional regularization terms to promote dispersion across text embeddings and reduce calibration error, yet these methods often suffer from a drop in accuracy.
Motivated by the well‑calibrated nature of zero‑shot predictions, we propose \cots, a simple yet effective post‑hoc calibration method that preserves accuracy.
Specifically, \cots~applies temperature scaling to minimize the confidence gap between adapted and zero‑shot predictions.
To fully exploit the potential of multiple augmentations during adaptation, we introduce a weak‑strong ensemble strategy that further boosts accuracy.
We then apply \cots~to this ensemble, termed \ecots, to maintain its well-calibrated property.
Extensive experiments on diverse datasets and backbones show that our approaches effectively mitigate miscalibration without compromising primary accuracy.
For instance, \ecots~reduces the average expected calibration error of TPT from 11.90\% to 5.38\% on ImageNet variants, while even increasing accuracy from 60.74\% to 62.95\%.
Moreover, when integrated with existing calibration methods, \ecots~usually enhances both accuracy and calibration simultaneously. Code is available at \url{https://github.com/yuweiliang911/CoTS}.
\end{abstract}

\section{Introduction}
Vision-language models (VLMs), such as CLIP~\cite{radford2021learning} and ALIGN~\cite{jia2021scaling}, are pre-trained to align visual and textual features in a shared embedding space, enabling strong zero-shot performance.
To efficiently adapt VLMs to diverse downstream tasks, pioneering works~\cite{zhou2022learning,zhou2022conditional} replace hand-crafted prompts with learnable continuous vectors optimized on training data.
For unsupervised settings, earlier methods typically require access to multiple samples from either the entire test set~\cite{huang2022unsupervised,tanwisuth2023pouf,liang2024realistic} or streaming data~\cite{ma2023swapprompt,xiao2025dynaprompt}.
In contrast, episodic test-time adaptation (TTA) methods~\cite{shu2022testtime,feng2023diverse,sheng2025r} perform prompt tuning using a single test instance, which is more challenging and has received increasing attention.
A classic approach, test-time prompt tuning (TPT)~\cite{shu2022testtime} adapts CLIP during inference by minimizing the marginal entropy over multiple augmented predictions.

\begin{figure}[t]
    \centering
    \begin{overpic}[width=0.9\linewidth]{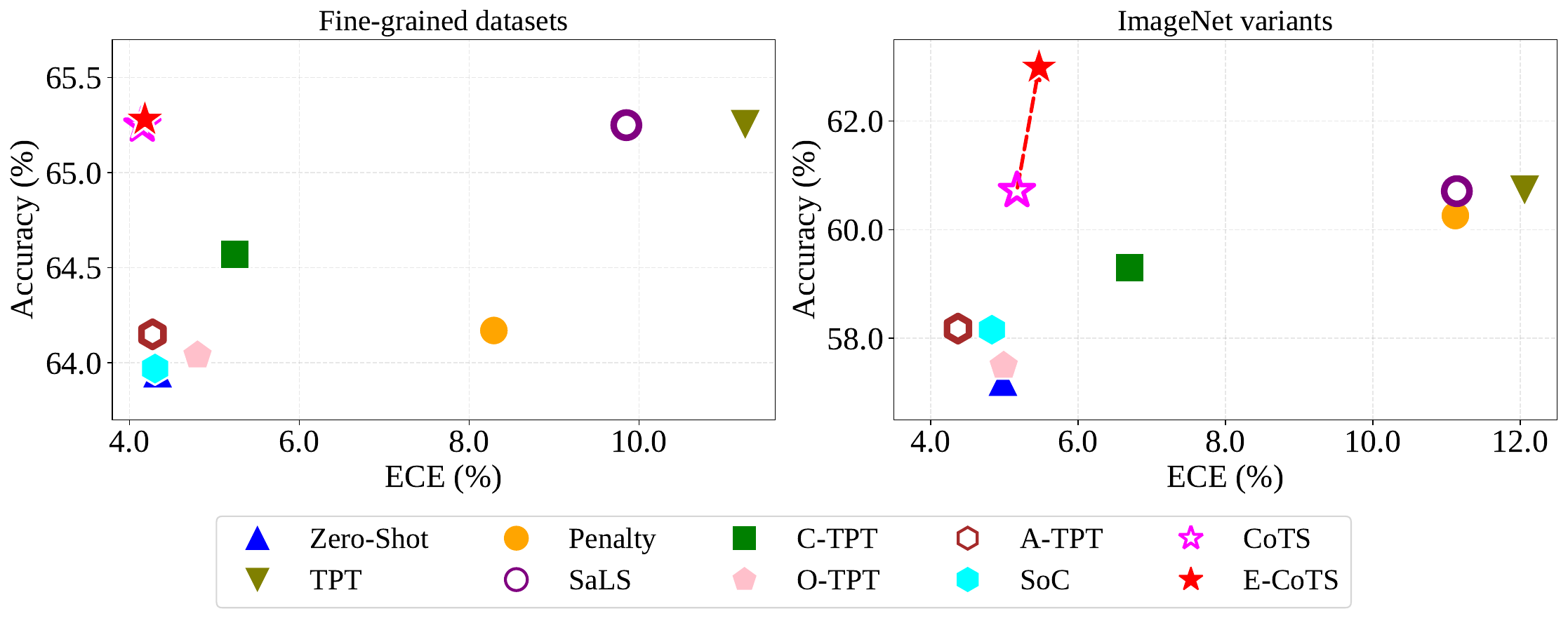}
    \put(27.8,4.8){\scriptsize\cite{radford2021learning}}%
    \put(42.3,4.8){\scriptsize\cite{murugesan2024robust}}%
    \put(56.5,4.8){\scriptsize\cite{yoon2024ctpt}}%
    \put(70.8,4.8){\scriptsize\cite{ahamed2026atpt}}%

    \put(23.5,2.2){\scriptsize\cite{shu2022testtime}}%
    \put(40.6,2.2){\scriptsize\cite{murugesan2024robust}}%
    \put(56.5,2.2){\scriptsize\cite{sharifdeen2025otpt}}%
    \put(68.6,2.2){\scriptsize\cite{fillioux2026soc}}%
    \end{overpic}
    \caption{
    Average accuracy (\%) and expected calibration error (ECE, \%) of zero-shot CLIP, TPT, and various calibration methods on fine-grained datasets and ImageNet variants using ViT-B/16.
    }
    \label{fig:scatter}
    \vspace{-20pt}
\end{figure}
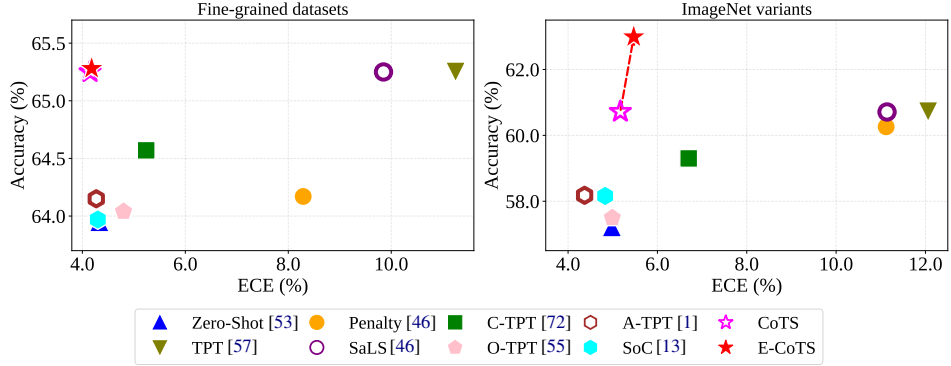

Although TPT~\cite{shu2022testtime} improves accuracy over zero-shot CLIP, it is prone to generating overconfident predictions, raising reliability concerns for VLMs in safety-critical applications~\cite{khandelwal2022simple,pan2024vlp,koleilat2025biomedcoop,silva2025few}.
A pioneering work, C-TPT~\cite{yoon2024ctpt}, shows that well-calibrated prompts tend to produce text embeddings with broader class-wise dispersion and thus introduces a distance-dispersion regularization term to push prompt embeddings away from the class centroid.
Building on this insight, subsequent works~\cite{sharifdeen2025otpt,ahamed2026atpt,fillioux2026soc} further propose various regularization terms to encourage angular separation between textual features.
Although these regularization-based methods achieve lower expected calibration error (ECE) than the original TPT~\cite{shu2022testtime}, we find they suffer from a noticeable accuracy drop, as shown in Fig.~\ref{fig:scatter}.
We argue that reduced ECE is less meaningful when it comes at the expense of accuracy, especially as the underlying calibration mechanisms remain poorly interpretable.

Alternatively, SaLS~\cite{murugesan2024robust} presents the only post-hoc calibration method that preserves accuracy by adjusting logits to remain within the zero-shot range after inference, but its calibration improvement remains limited.
It is well known that CLIP~\cite{radford2021learning} is generally well-calibrated~\cite{minderer2021revisiting,galil2023what}, with a relatively small gap between accuracy and confidence.
Taking a closer look at the accuracy of CLIP and TPT, we find that prompt tuning using a single test instance does not substantially alter accuracy, suggesting that the increased miscalibration stems from a shift in confidence.
To validate this hypothesis, we conduct a pilot study that assigns zero-shot confidence to TPT predictions.
We find that this straightforward replacement significantly reduces calibration error, surpassing~\cite{murugesan2024robust,yoon2024ctpt} while falling short of~\cite{sharifdeen2025otpt,fillioux2026soc}.
However, this naive confidence replacement introduces a critical inconsistency: the predicted class comes from TPT's softmax distribution, whereas the confidence comes from the zero-shot distribution, potentially leading to unstable decisions.

Based on these insights, we propose \textbf{Co}nfidence-based \textbf{T}emperature \textbf{S}caling (\cots), a simple yet effective calibration method that follows the classic idea of temperature scaling~\cite{guo2017calibration}.
Specifically, \cots~optimizes a learnable temperature parameter to align TPT‑predicted confidence with zero‑shot confidence, thereby mitigating calibration error while retaining accuracy.
Furthermore, inspired by prior work~\cite{farina2024frustratingly,dafnis2025testtime} that ensembles augmented views to enhance prediction robustness, we introduce a new weak-strong ensemble strategy to further boost accuracy.
In this strategy, we retain the weakly augmented view and balance its predictions with those from selected strong augmentations via a weighted trade-off.
Empirically, we find that this ensemble strategy indeed boosts accuracy but may disrupt the model's well-calibrated properties.
To mitigate this issue, we further integrate the ensemble strategy with \cots, termed \ecots, achieving a better trade-off between accuracy and calibration.
We validate the proposed methods on both fine-grained datasets and ImageNet variants.

To summarize, our contributions are as follows:
\begin{itemize}
\item We propose \cots, a novel post-hoc confidence-based calibration method that uses temperature scaling to align TPT-predicted confidence with zero-shot confidence.
\item We devise a simple weak-strong ensemble strategy integrated with \cots, dubbed \ecots, that achieves reliable classification against TPT with superior accuracy and lower calibration error.
\item Experiments across diverse datasets and backbones show that our methods reduce ECE to near the well-calibrated zero-shot baseline, while maintaining or even improving the adapted model's classification accuracy.
Both \cots~and \ecots~are plug-and-play and usually complement existing regularization-based calibration approaches, improving either accuracy or ECE.
\end{itemize}

\section{Preliminaries and Motivation}
\textbf{CLIP-based zero-shot classification.}
CLIP~\cite{radford2021learning} consists of a visual encoder $f_v(\cdot)$ and a text encoder $f_t(\cdot)$.
For a $K$-class classification task with label space $\mathcal{Y} = \{y_1, y_2, \dots, y_K\}$, let $x$ denote an input image and $y \in \mathcal{Y}$ its ground-truth label.
The visual encoder maps $x$ to an image embedding $v = f_v(x) \in \mathbb{R}^d$, while each class label $y_k \in \mathcal{Y}$ is converted into a textual prompt $c_k$ (\eg, using the template "a photo of a [CLASS]") and then encoded into a text embedding $t_k = f_t(c_k)$.
Both the image and text embeddings are $l_2$-normalized, \ie, $\|v\|=\|t_k\|=1$.
Then, the $k$-th values of the logit vector $\textbf{z}$ and softmax probability vector $\textbf{p}(x)$ corresponding to class $y_k$ are defined as follows:
\begin{equation}
    \textbf{z}_k=(v^T\cdot t_k)/\tau_{\text{clip}}, \quad \textbf{p}_k(x) = \textbf{p}(y=k|x) = \frac{\exp(\textbf{z}_k)}{\sum_{j=1}^{K} \exp(\textbf{z}_j)},
\end{equation}
where the scaling factor $\tau_{\text{clip}}=0.01$ is learned during pre-training \cite{radford2021learning}.

\textbf{Test-time prompt tuning (TPT).}
TPT~\cite{shu2022testtime} adapts CLIP~\cite{radford2021learning} to a single test image by optimizing text prompts at test time, following the test-time adaptation paradigm~\cite{sun2020test,zhang2022memo,liang2025comprehensive}.
Besides the weak augmented view $\mathcal{A}_1(x)$, TPT generates $(N-1)$ randomly augmented views of the test image using AugMix~\cite{hendrycks2020augmix}, and optimizes the input text prompts by minimizing the  marginal entropy loss:
\begin{equation}
    \mathcal{L}_{\text{TPT}} = -\sum_{k=1}^K \bar{\textbf{p}}_k(x)\log \bar{\textbf{p}}_k(x), \quad \bar{\textbf{p}}_k(x)=\frac{1}{\rho N}\sum_{i=1}^{N} \mathds{1}[\mathbf{H}(\textbf{p}(\mathcal{A}_{i}(x))) \leq \gamma] \ \textbf{p}_k(\mathcal{A}_{i}(x)).
\end{equation}
Here $\bar{\textbf{p}}_k(x)$ is the average softmax probability over selected augmentations that correspond to small entropy values, $\gamma$ is the threshold corresponding to a cutoff percentile $\rho$ (defaulting to 0.1 \cite{shu2022testtime}), and $\mathbf{H}(\textbf{p}(\mathcal{A}_i(x)))$ measures the self-entropy of the prediction on the $i$-th augmented view.

\textbf{Temperature scaling.}
Temperature Scaling (TS)~\cite{guo2017calibration} is a classic post-hoc calibration technique~\cite{platt1999probabilistic} that adjusts model confidence without altering its predictions.
It introduces a single temperature parameter $\tau>0$ to rescale the logits before applying the softmax:
\begin{equation}
    \hat{\textbf{p}}_k(x;\tau) = \frac{\exp(\textbf{z}_k / \tau)}{\sum_{j=1}^{K} \exp(\textbf{z}_j / \tau)}, \quad k \in [1, \dots, K],
\end{equation}
where $\textbf{z}_k$ is the logit for class $k$.
A larger $\tau$ softens the distribution (reducing overconfidence), while a smaller $\tau$ sharpens it.
The optimal temperature $\tau^*$ is typically obtained by minimizing the negative log-likelihood (NLL) on a labeled validation set $V = \{x_i, y_i\}_{i=1}^{|V|}$:
\begin{equation}
    \tau^* = \arg \min_{\tau} - \sum\limits_{i=1}^{|V|} \log \hat{\textbf{p}}_{y_i}(x_i;\tau).
\end{equation}
However, no labeled validation set is available in the episodic test-time adaptation problem, motivating alternative strategies for determining the temperature.

\subsection{A Critical Examination of Prior Calibration Methods}
Although TPT~\cite{shu2022testtime} improves accuracy compared to zero-shot CLIP~\cite{radford2021learning}, recent studies have shown that it produces overconfident predictions~\cite{murugesan2024robust,yoon2024ctpt}.
To address this, a line of work has focused on calibrating TPT~\cite{yoon2024ctpt,sharifdeen2025otpt,han2025d,ahamed2026atpt,fillioux2026soc}.
These methods introduce regularization terms during prompt optimization to encourage greater dispersion among text embeddings, \eg, distance-based dispersion~\cite{yoon2024ctpt}, orthogonal constraints~\cite{sharifdeen2025otpt}, angular diversity~\cite{ahamed2026atpt}, and semantic-aware separation~\cite{fillioux2026soc}.
Alternatively, SaLS~\cite{murugesan2024robust} identifies the increase in logit range as the cause of miscalibration and adjusts logits to remain within the zero-shot range.
To investigate whether these methods truly work as expected, we examine their ECE values and accuracies alongside those of CLIP and TPT in Fig.~\ref{fig:scatter}.

\textbf{Observation 1: Calibration methods reduce TPT's calibration error but do not significantly outperform zero-shot CLIP.}
As depicted in Fig.~\ref{fig:scatter}, zero-shot predictions maintain low ECE values, confirming that CLIP models are generally well-calibrated~\cite{minderer2021revisiting,galil2023what}.
While TPT~\cite{shu2022testtime} clearly boosts accuracy, it leads to a substantial increase in ECE.
Existing calibration methods significantly reduce the ECE of TPT, but most remain comparable to or slightly worse than zero-shot CLIP; only SoC~\cite{fillioux2026soc} marginally outperforms it.
The only post-hoc approach, SaLS~\cite{murugesan2024robust}, achieves notably weaker ECE reduction than its regularization-based counterparts.

\textbf{Observation 2: Calibration methods improve accuracy over CLIP but often degrade the accuracy of TPT.}
As depicted in Fig.~\ref{fig:scatter}, all regularization-based methods decrease the accuracy of TPT.
Specifically, the accuracy of O-TPT~\cite{sharifdeen2025otpt} and SoC~\cite{fillioux2026soc} degrade substantially, even approaching that of zero-shot CLIP.
In contrast, the post-hoc approach SaLS~\cite{murugesan2024robust} fully preserves TPT's accuracy.

Taken together, zero-shot CLIP serves as a strong calibration baseline, while TPT provides a clear accuracy gain.
Existing regularization-based methods trade accuracy for calibration, whereas SaLS~\cite{murugesan2024robust} preserves accuracy but offers limited calibration improvement.
This reveals an unresolved gap: \textbf{no existing method simultaneously preserves TPT's accuracy advantage and reduces calibration error to the well-calibrated zero-shot level.}

\subsection{Confidence Replacement: A Pilot Study}
\label{sec:pilot}
\begin{wrapfigure}[19]{r}{0.6\textwidth}%
  \centering
  \scriptsize
  \vspace{-10pt}%
  \setlength{\abovecaptionskip}{5pt}%
  \setlength{\belowcaptionskip}{5pt}%
  \begin{overpic}[width=0.98\linewidth]{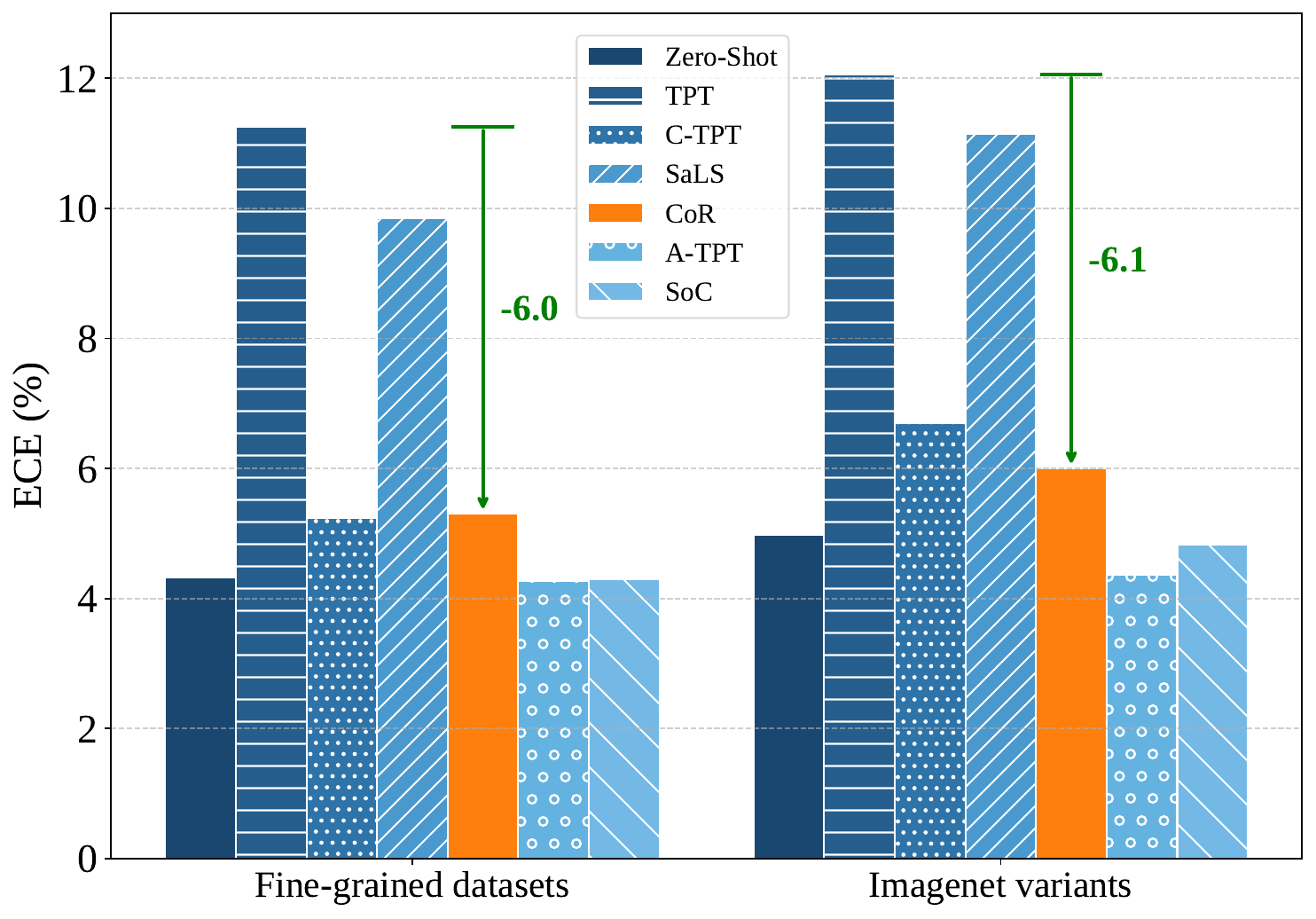}
  \put(55.0, 61.6){\tiny\cite{shu2022testtime}}%
  \put(56.8, 58.8){\tiny\cite{yoon2024ctpt}}%
  \put(55.5, 55.6){\tiny\cite{murugesan2024robust}}%
  \put(56.8, 49.9){\tiny\cite{ahamed2026atpt}}%
  \put(55.0, 46.8){\tiny\cite{fillioux2026soc}}%
  \end{overpic}
  \caption{Comparison of average ECE (\%) between \texttt{CoR} and existing calibration methods on fine-grained datasets and ImageNet variants using ViT-B/16.
  }
  \label{fig:replace}
\end{wrapfigure}
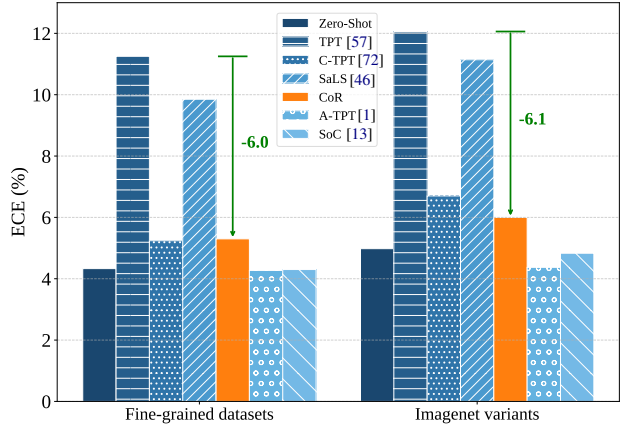

Comparing the accuracy of zero-shot CLIP and TPT, we find that adaptation using a single test instance typically yields limited accuracy gains, suggesting that TPT's miscalibration stems from a shift in confidence rather than a change in predictive performance.
To test this hypothesis, we conduct a simple experiment: for each test sample, we retain the predicted class from TPT but replace its confidence (maximum predicted probability) with the zero-shot confidence of the same sample, and then measure ECE.
We refer to this procedure as Confidence Replacement (\texttt{CoR}).

As shown in Fig.~\ref{fig:replace}, this simple replacement leads to a notable reduction in ECE, even outperforming SaLS~\cite{murugesan2024robust} and C-TPT~\cite{yoon2024ctpt}.
However, \texttt{CoR} remains inferior to recent calibration methods such as SoC~\cite{fillioux2026soc}.
We attribute this to a fundamental inconsistency: the predicted class is derived from the TPT distribution, whereas the confidence comes from the zero-shot distribution, which may lead to incoherent decisions.
Nevertheless, this experiment confirms two key findings: (1) the miscalibration of TPT is largely caused by confidence inflation, and (2) zero-shot confidence provides a useful reference for calibration.
These observations motivate a principled approach that aligns TPT confidence with zero-shot confidence while maintaining a consistent probability distribution.

\section{Methodology}
Section~\ref{sec:cots} describes the proposed confidence-based temperature scaling method, while Section~\ref{sec:ecots} then presents its enhanced variant under a new weak-strong ensemble strategy.

\subsection{Confidence-based Temperature Scaling}
\label{sec:cots}
Based on the findings in Section~\ref{sec:pilot}, we propose \textbf{Co}nfidence-based \textbf{T}emperature \textbf{S}caling (\cots), a post-hoc calibration method that follows the classic idea of temperature scaling~\cite{guo2017calibration}.
Unlike standard temperature scaling, which requires a labeled validation set to optimize the NLL, \cots~leverages the well-calibrated zero-shot predictions as a label-free reference.
Specifically, \cots~optimizes a learnable temperature parameter $\tau$ to align TPT-predicted confidence with zero-shot confidence, thereby reducing calibration error while retaining accuracy.

Let $\textbf{z}^{\texttt{tpt}}(x)$ and $\textbf{z}^{\texttt{zs}}(x)$ denote the logits of the TPT-adapted model and the zero-shot model for the test image $x$, respectively.
For a single augmented view, the objective is to minimize the squared difference between the temperature-scaled TPT confidence and the zero-shot confidence:
$\left(\max_k \hat{\textbf{p}}^{\texttt{tpt}}_k(x;\tau)- \max_k \textbf{p}_k^{\texttt{zs}}(x)\right)^2$,
where $\hat{\textbf{p}}^{\texttt{tpt}}$ and $\textbf{p}^{\texttt{zs}}$ are the corresponding softmax probabilities.
To fully exploit the multiple augmentations available in TPT and reduce the sensitivity of relying on the weak view itself, we optimize over the set of selected strongly augmented views:
\begin{equation}
    \tau^* = \arg\min_{\tau} \frac{1}{|\Omega|}\sum_{i\in \Omega} \left(\max_k \hat{\textbf{p}}^{\texttt{tpt}}_k(\mathcal{A}_i(x);\tau)- \max_k \textbf{p}_k^{\texttt{zs}}(\mathcal{A}_i(x))\right)^2,
    \label{eq:cots}
\end{equation}
where $\Omega=\{i \mid i>1,\; \mathbf{H}(\textbf{p}^{\texttt{tpt}}(\mathcal{A}_{i}(x))) \leq \gamma'\}$ denotes the set of strongly augmented views selected by having low entropy after adaptation under the threshold $\gamma'$, excluding the weakly augmented view.
Once the optimal parameter $\tau^*$ is obtained, the calibrated probability is computed as $\hat{\textbf{p}}^{\texttt{tpt}}(\mathcal{A}_1(x);\tau^*)$.

Compared to existing regularization-based calibration methods~\cite{yoon2024ctpt,sharifdeen2025otpt,ahamed2026atpt,fillioux2026soc}, which modify the prompt optimization objective and may interfere with discriminative performance, \cots~operates entirely after adaptation and thus avoids any accuracy degradation.

\subsection{Ensemble with Confidence-Based Temperature Scaling}
\label{sec:ecots}
While \cots~reduces calibration error without degrading the accuracy of TPT, it mainly preserves rather than improves the adapted prediction.
We wonder whether we can further improve accuracy while maintaining good calibration, a goal that is especially important since existing calibration methods (\eg, O-TPT~\cite{sharifdeen2025otpt}, SoC~\cite{fillioux2026soc}) often suffer from a notable accuracy degradation.
Orthogonal to these calibration methods, ensembling over augmented views provides an effective way to boost accuracy during CLIP adaptation~\cite{farina2024frustratingly,sheng2025r,dafnis2025testtime}, and ZERO~\cite{farina2024frustratingly} theoretically shows that marginalizing predictions across augmented views can reduce the error bound compared with single-view inference.

Inspired by these insights, we propose a weak-strong ensemble strategy that further exploits the predictions from selected strongly augmented views to boost accuracy.
Unlike prior work that uses majority voting~\cite{farina2024frustratingly} or uniform averaging~\cite{dafnis2025testtime} over augmented views, we balance the predictions from weak and strong augmentations using an adaptive weight:
\begin{equation}
    \textbf{p}_{\text{ens}} = \alpha \cdot \textbf{p}^{\texttt{tpt}}(\mathcal{A}_1(x)) + (1-\alpha) \cdot \frac{1}{|\Omega|}\sum_{i\in \Omega} \textbf{p}^{\texttt{tpt}}(\mathcal{A}_i(x)),
    \label{eq:ens}
\end{equation}
where $\alpha = \frac{1}{K(K-1)} \sum_{i \neq j} t_i^T\cdot t_j$ is the average pairwise cosine similarity among the $K$ normalized text embeddings $\{t_i\}_{i=1}^{K}$.
Intuitively, a higher $\alpha$ indicates greater inter-class similarity in the text embedding space, which makes entropy-based view selection less discriminative (\ie, the average of strong views becomes relatively unreliable).
In such cases, the weakly augmented view (original image) is more reliable, and the ensemble assigns it a larger weight.
Conversely, when text embeddings are well-separated ($\alpha$ is small), the entropy-filtered strongly augmented views provide more reliable complementary information.
Note that $\alpha$ depends only on the text embeddings and thus varies across tasks and prompts, providing a task-adaptive weighting without per-sample tuning.

Through a preliminary study, we find that directly applying this ensemble on top of calibrated models indeed enhances accuracy but may disrupt their well-calibrated properties.
To address this, we integrate the proposed confidence-based temperature scaling into the ensemble strategy.
Specifically, we substitute each probability vector in Eq.~(\ref{eq:ens}) with its temperature-calibrated version:
\begin{equation}
    \label{eq:ecots}
    \hat{\textbf{p}}_{\text{ens}} = \alpha \cdot \hat{\textbf{p}}^{\texttt{tpt}}(\mathcal{A}_1(x);\tau^*) + (1-\alpha) \cdot \frac{1}{|\Omega|}\sum_{i\in \Omega} \hat{\textbf{p}}^{\texttt{tpt}}(\mathcal{A}_i(x);\tau^*).
\end{equation}
The combined approach, termed \ecots, achieves a better trade-off between accuracy and calibration: the ensemble component enhances classification performance, while temperature scaling reduces calibration error.
Both methods are post-hoc, and their pseudocode is provided in Appendix~\ref{code}.
\nocite{yan2026if}

\section{Experiment}

\subsection{Experimental Setup}
\textbf{Datasets.}
To comprehensively evaluate the performance of \cots~and \ecots, we first use a fine-grained benchmark consisting of eleven image classification datasets, including ImageNet~\cite{deng2009imagenet}, DTD~\cite{cimpoi2014describing}, Flowers102~\cite{nilsback2008automated}, Food101~\cite{bossard2014food}, SUN397~\cite{xiao2010sun}, Aircraft~\cite{maji2013fine}, OxfordPets~\cite{parkhi2012cats}, Caltech101~\cite{fei2004learning}, UCF101~\cite{soomro2012ucf101}, EuroSAT~\cite{helber2018introducing}, and StanfordCars~\cite{krause20133d}.
To further evaluate robustness under natural distribution shifts, we conduct experiments on four ImageNet variants, including ImageNet-A (natural adversarial examples)~\cite{hendrycks2021natural}, ImageNet-V (re-collected images)~\cite{recht2019imagenet}, ImageNet-R (artistic renditions)~\cite{hendrycks2021many}, and ImageNet-K (sketch-style images with domain shifts)~\cite{wang2019learning}.
In the episodic test-time adaptation setting \cite{dong2025adapting}, each data instance is adapted independently of the others.

\textbf{Baselines.}
We first compare \cots~with representative calibration methods designed for fine-tuned CLIP~\cite{radford2021learning}.
These include regularization-based methods that encourage textual dispersion, such as C-TPT~\cite{yoon2024ctpt}, O-TPT~\cite{sharifdeen2025otpt}, A-TPT~\cite{ahamed2026atpt}, and SoC~\cite{fillioux2026soc}; logit-adjustment methods including ZS-Norm~\cite{murugesan2024robust}, Penalty~\cite{murugesan2024robust}, and SaLS~\cite{murugesan2024robust}.
In addition to direct comparison, we also evaluate whether our advanced \ecots~can serve as a plug-and-play module on top of existing calibration methods \cite{yoon2024ctpt,sharifdeen2025otpt,ahamed2026atpt,fillioux2026soc}.
To examine generality beyond TPT methods, we further apply \cots~to other episodic test-time adaptation methods, including TTL~\cite{imam2025test} and TPS~\cite{sui2025just}.

\textbf{Implementation details.}
We use ViT-B/16~\cite{radford2021learning} as the backbone model and follow the standard TPT~\cite{shu2022testtime} protocol.
The initial text prompt is set to "a photo of a [CLASS]" and optimized using AdamW~\cite{loshchilov2018decoupled} with a single gradient step and a learning rate of 0.005.
For each test sample, we generate 64 augmented views using AugMix~\cite{hendrycks2020augmix} and select confident views following TPT \cite{shu2022testtime}.
For temperature learning, we select the top 10\% ($\rho$=0.1) most confident samples, excluding the weak augmentation (original image), and train the temperature parameter for 50 steps.
We implement all baselines using their official code and report the average results over three random seeds.
\textbf{We provide additional results (\eg, for ResNet-50, standard deviations, and other metrics) in the Appendix.}

\subsection{Results}

\begin{table*}[t]
\centering
\caption{Performance comparison of different methods on fine-grained datasets using ViT-B/16. Accuracy (\%) and ECE (\%) are reported. In the Average column, \textcolor{green!100}{green} indicates performance improvement over TPT, while \textcolor{red!100}{red} indicates performance degradation.}
\label{tab:main_standard}
\scriptsize
\setlength{\tabcolsep}{3.2pt}
\resizebox{\textwidth}{!}{
\begin{tabular}{llcccccccccccr}
\toprule
(\%) & Method & ImgNet & DTD & Flowers & Food101 & SUN397 & Aircraft & Pets & Caltech & UCF101 & EuroSAT & Cars & Average \\
\midrule
\multirow{10}{*}{\rotatebox[origin=c]{90}{Accuracy}}
& Zero-Shot~\cite{radford2021learning}  & 66.72 & 44.33 & 67.32 & 83.66 & 62.58 & 23.85 & 88.20 & 93.96 & 65.19 & 42.05 & 65.56 & 63.95$_\texttt{{\textcolor{white!100}{+0.00}}}$ \\
& TPT~\cite{shu2022testtime} & 68.91 & 47.12 & 68.64 & 84.65 & 65.50 & 23.27 & 87.25 & 94.08 & 68.20 & 42.91 & 66.58 & 65.19$_\texttt{{\textcolor{white!100}{+0.00}}}$ \\
& C-TPT~\cite{yoon2024ctpt} & 68.45 & 45.17 & 69.58 & 83.14 & 64.53 & 24.10 & 88.20 & 93.75 & 65.05 & 42.45 & 65.80 & 64.57$_\texttt{{\textcolor{red!100}{-0.63}}}$ \\
& Penalty~\cite{murugesan2024robust} & 68.84 & 45.80 & 68.36 & 84.43 & 65.44 & 22.75 & 82.94 & 93.27 & 67.09 & 41.20 & 66.19 & 64.21$_\texttt{{\textcolor{red!100}{-0.98}}}$ \\
& SaLS~\cite{murugesan2024robust} & 68.91 & 47.12 & 68.67 & 84.65 & 65.49 & 23.29 & 87.25 & 94.08 & 68.21 & 42.91 & 66.57 & 65.20$_\texttt{{\textcolor{green!100}{+0.00}}}$ \\
& O-TPT~\cite{sharifdeen2025otpt} & 67.30 & 45.49 & 69.03 & 82.78 & 63.12 & 23.65 & 88.13 & 93.54 & 63.82 & 42.44 & 65.24 & 64.05$_\texttt{{\textcolor{red!100}{-1.14}}}$ \\
& A-TPT~\cite{ahamed2026atpt} & 67.92 & 45.17 & 69.79 & 83.06 & 63.20 & 23.80 & 87.89 & 93.24 & 64.03 & 42.05 & 65.20 & 64.12$_\texttt{{\textcolor{red!100}{-1.07}}}$ \\
& SoC~\cite{fillioux2026soc} & 67.57 & 42.79 & 68.01 & 83.60 & 62.37 & 24.01 & 88.30 & 94.01 & 65.47 & 42.04 & 65.31 & 63.95$_\texttt{{\textcolor{red!100}{-1.24}}}$ \\
& \cellcolor{casbg} \texttt{\hspace*{-2pt}CoTS} & \cellcolor{casbg}68.91 & \cellcolor{casbg}47.12 & \cellcolor{casbg}68.64 & \cellcolor{casbg}84.65 & \cellcolor{casbg}65.50 & \cellcolor{casbg}23.27 & \cellcolor{casbg}87.25 & \cellcolor{casbg}94.08 & \cellcolor{casbg}68.20 & \cellcolor{casbg}42.91 & \cellcolor{casbg}66.58 & \cellcolor{casbg}65.19$_\texttt{{\textcolor{green!100}{+0.00}}}$ \\
& \cellcolor{casbg} \texttt{\hspace*{-2pt}E-CoTS} & \cellcolor{casbg}69.49 & \cellcolor{casbg}47.32 & \cellcolor{casbg}68.25 & \cellcolor{casbg}84.54 & \cellcolor{casbg}65.67 & \cellcolor{casbg}23.59 & \cellcolor{casbg}86.94 & \cellcolor{casbg}94.13 & \cellcolor{casbg}68.25 & \cellcolor{casbg}41.68 & \cellcolor{casbg}67.57 & \cellcolor{casbg}65.22$_\texttt{{\textcolor{green!100}{+0.03}}}$ \\
\midrule
\multirow{10}{*}{\rotatebox[origin=c]{90}{ECE}}
& Zero-Shot~\cite{radford2021learning} & 1.98  & 8.24  & 2.73  & 2.08  & 2.24  & 5.53  & 4.45  & 5.91  & 2.95  & 7.10  & 4.40  & 4.33$_\texttt{{\textcolor{white!100}{+0.00}}}$  \\
& TPT~\cite{shu2022testtime} & 10.56 & 20.96 & 13.59 & 4.31  & 11.23 & 16.96 & 5.50  & 4.36  & 11.57 & 20.15 & 5.08  & 11.30$_\texttt{{\textcolor{white!100}{+0.00}}}$ \\
& C-TPT~\cite{yoon2024ctpt} & 5.14  & 12.94 & 5.16  & 3.39  & 5.10  & 4.16  & 1.63  & 4.41  & 2.49  & 11.66 & 1.47  & 5.23$_\texttt{{\textcolor{green!100}{-6.07}}}$ \\
& Penalty~\cite{murugesan2024robust} & 10.38 & 14.88 & 12.08 & 2.67  & 10.98 & 15.65 & 1.84  & 4.89  & 9.14  & 4.62  & 3.94  & 8.28$_\texttt{{\textcolor{green!100}{-3.02}}}$ \\
& SaLS~\cite{murugesan2024robust} & 9.72  & 18.61 & 11.93 & 4.35  & 11.01 & 15.76 & 5.12  & 4.37  & 10.66 & 13.83 & 3.94  & 9.94$_\texttt{{\textcolor{green!100}{-1.36}}}$ \\
& O-TPT~\cite{sharifdeen2025otpt} & 2.03  & 7.66  & 3.68  & 4.28  & 8.42  & 3.76  & 2.13  & 4.53  & 2.70  & 11.52 & 1.81  & 4.77$_\texttt{{\textcolor{green!100}{-6.52}}}$  \\
& A-TPT~\cite{ahamed2026atpt}& 2.38  & 8.49  & 4.23  & 3.19  & 4.17  & 6.25  & 2.18  & 5.30  & 2.63  & 7.11  & 1.68  & 4.33$_\texttt{{\textcolor{green!100}{-6.97}}}$  \\
& SoC~\cite{fillioux2026soc} & 3.71  & 7.63  & 2.98  & 2.84  & 2.84  & 5.41  & 1.83  & 5.80  & 2.71  & 7.08  & 4.63  & 4.31$_\texttt{{\textcolor{green!100}{-6.98}}}$  \\
& \cellcolor{casbg} \texttt{\hspace*{-2pt}CoTS} &
\cellcolor{casbg}3.39  & \cellcolor{casbg}6.67  & \cellcolor{casbg}2.48  & \cellcolor{casbg}1.85  & \cellcolor{casbg}3.51  & \cellcolor{casbg}5.21  & \cellcolor{casbg}2.91  & \cellcolor{casbg}6.13  & \cellcolor{casbg}3.11  & \cellcolor{casbg}7.58  & \cellcolor{casbg}4.29  & \cellcolor{casbg}4.28$_\texttt{{\textcolor{green!100}{-7.01}}}$  \\
& \cellcolor{casbg} \texttt{\hspace*{-2pt}E-CoTS} & \cellcolor{casbg}3.00  & \cellcolor{casbg}6.94  & \cellcolor{casbg}4.10  & \cellcolor{casbg}1.29  & \cellcolor{casbg}2.59  & \cellcolor{casbg}5.73  & \cellcolor{casbg}3.43  & \cellcolor{casbg}4.74  & \cellcolor{casbg}3.50  & \cellcolor{casbg}8.23  & \cellcolor{casbg}3.84  & \cellcolor{casbg}4.31$_\texttt{{\textcolor{green!100}{-6.98}}}$ \\
\bottomrule
\end{tabular}
}
\vspace{-10pt}
\end{table*}

\textbf{Performance on fine-grained datasets.}
We evaluate the performance on fine-grained datasets with ViT-B/16 in Table~\ref{tab:main_standard}.
\cots~preserves average accuracy of TPT while substantially improving calibration, reducing average ECE from 11.30\% to 4.28\%, which is the best among all methods.
On Flowers, \cots~attains the lowest ECE overall, even outperforming zero-shot.
Compared to SaLS~\cite{murugesan2024robust}, the only post-hoc calibration baseline, \cots~achieves much lower ECE.
Furthermore, \ecots~improves the average accuracy and maintains a competitive ECE, demonstrating a better balance between classification performance and calibration.

\begin{table}[t]
\centering
\caption{Performance comparison of different methods on ImageNet variants using ViT-B/16. Accuracy (\%) and ECE (\%) are reported.}
\label{tab:imagenet_variants}
\scriptsize
\setlength{\tabcolsep}{5pt}
\renewcommand{\arraystretch}{1.12}
\resizebox{\linewidth}{!}{
\begin{tabular}{lccccrccccr}
\toprule
\multirow{2}{*}{Method}
& \multicolumn{5}{c}{Accuracy (\%)}
& \multicolumn{5}{c}{ECE (\%)} \\
\cmidrule(lr){2-6} \cmidrule(lr){7-11}
& -A & -V & -R & -Sk & Average
& -A & -V & -R & -Sk & Average \\
\midrule
Zero-Shot~\cite{radford2021learning}
& 47.83 & 60.94 & 73.99 & 46.10 & 57.22$_\texttt{{\textcolor{white!100}{+0.00}}}$
& 8.34  & 3.18  & 3.55  & 4.87  & 4.98$_\texttt{{\textcolor{white!100}{+0.00}}}$ \\
TPT~\cite{shu2022testtime}
& 54.63 & 63.45 & 77.05 & 47.82 & 60.74$_\texttt{{\textcolor{white!100}{+0.00}}}$
& 15.20 & 11.88 & 4.84  & 15.69 & 11.90$_\texttt{{\textcolor{white!100}{+0.00}}}$ \\
C-TPT~\cite{yoon2024ctpt} & 51.19 & 62.58 & 75.79 & 47.43 & 59.25$_\texttt{{\textcolor{red!100}{-1.49}}}$
& 8.18  & 6.55  & 1.74  & 10.20 & 6.67$_\texttt{{\textcolor{green!100}{-5.24}}}$ \\
Penalty~\cite{murugesan2024robust}
& 53.07 & 63.40 & 76.68 & 47.79 & 60.24$_\texttt{{\textcolor{red!100}{-0.50}}}$
& 13.33 & 11.64  & 3.60  & 15.40 & 10.99$_\texttt{{\textcolor{green!100}{-0.91}}}$ \\
SALS~\cite{murugesan2024robust}
& 54.63 & 63.44 & 77.05 & 47.82 & 60.74$_\texttt{{\textcolor{green!100}{+0.00}}}$
& 13.95 & 10.81 & 3.87  & 15.29 & 10.98$_\texttt{{\textcolor{green!100}{-0.92}}}$ \\
O-TPT~\cite{sharifdeen2025otpt}
& 48.12 & 61.35 & 73.98 & 46.59 & 57.51$_\texttt{{\textcolor{red!100}{-3.23}}}$
& 6.68  & 3.45  & 3.98  & 5.76  & 4.97$_\texttt{{\textcolor{green!100}{-6.94}}}$ \\
A-TPT~\cite{ahamed2026atpt}
& 49.16 & 61.77 & 74.99 & 46.89 & 58.20$_\texttt{{\textcolor{red!100}{-2.54}}}$
& 5.55  & 2.85  & 4.09  & 4.87  & 4.34$_\texttt{{\textcolor{green!100}{-7.56}}}$ \\
SoC ~\cite{fillioux2026soc}
& 49.57 & 61.36 & 75.11 & 46.45 & 58.12$_\texttt{{\textcolor{red!100}{-2.62}}}$
& 8.15  & 3.65  & 4.50  & 3.16  & 4.87$_\texttt{{\textcolor{green!100}{-7.04}}}$ \\
\rowcolor{casbg} \texttt{\hspace*{0pt}CoTS}
& 54.63 & 63.45 & 77.05 & 47.82 & 60.74$_\texttt{{\textcolor{green!100}{+0.00}}}$
& 7.85 & 4.37 & 5.09 & 3.87 & 5.30$_\texttt{{\textcolor{green!100}{-6.61}}}$ \\
\rowcolor{casbg} \texttt{\hspace*{0pt}E-CoTS}
& 60.72 & 64.49 & 77.88 & 48.72 & 62.95$_\texttt{{\textcolor{green!100}{+2.22}}}$
& 8.71 & 3.77 & 2.94 & 6.11 & 5.38$_\texttt{{\textcolor{green!100}{-6.52}}}$ \\
\bottomrule
\end{tabular}
}
\end{table}

\textbf{Performance on ImageNet variants.}
The results on four ImageNet variants under distribution shifts are demonstrated in Table~\ref{tab:imagenet_variants}.
It is clear that regularization-based methods (O-TPT~\cite{sharifdeen2025otpt}, A-TPT~\cite{ahamed2026atpt}, SoC~\cite{fillioux2026soc}) reduce ECE but at the cost of accuracy.
In contrast, \cots~preserves TPT's average accuracy while more than halving the average ECE, demonstrating its effectiveness without altering predictions.
With the weak-strong ensemble strategy, \ecots~achieves the best average accuracy of 62.95\% while maintaining a competitive ECE of 5.38\%.
These results indicate that our approach not only mitigates miscalibration but also enhances the robustness of test-time prompt tuning.

\begin{table}[t]
\centering
\caption{
Compatibility of \texttt{CoTS}, and \texttt{E-CoTS} with different base methods using ViT-B/16.
Average accuracy (\%) and ECE (\%) are reported on fine-grained datasets and ImageNet variants, respectively.
}
\label{tab:cots_ablation_full}
\renewcommand{\arraystretch}{1.1}
\setlength{\tabcolsep}{4.5pt}
\resizebox{\linewidth}{!}{
\begin{tabular}{llcccccccccc}
\toprule
\multirow{3}{*}{(\%)} & \multirow{3}{*}{Method}
& \multicolumn{5}{c}{Fine-grained datasets} & \multicolumn{5}{c}{ImageNet variants} \\
\cmidrule(lr){3-7} \cmidrule(lr){8-12}
& & C-TPT & Penalty & O-TPT & A-TPT & SoC & C-TPT & Penalty & O-TPT & A-TPT & SoC \\
& & \cite{yoon2024ctpt} & \cite{murugesan2024robust} & \cite{sharifdeen2025otpt} & \cite{ahamed2026atpt} & \cite{fillioux2026soc} & \cite{yoon2024ctpt} & \cite{murugesan2024robust} & \cite{sharifdeen2025otpt} & \cite{ahamed2026atpt} & \cite{fillioux2026soc} \\
\midrule
\multirow{2}{*}{\rotatebox[origin=c]{90}{ACC}}
& base & 64.57 & 64.21 & 64.05 & 64.12 & 63.95 & 59.25 & 60.24 & 57.51 & 58.20 & 58.12\\
& \cellcolor{casbg}+ \texttt{E-CoTS} & \cellcolor{casbg}64.64 & \cellcolor{casbg}65.25 & \cellcolor{casbg}64.93 & \cellcolor{casbg}65.07 & \cellcolor{casbg}65.01 & \cellcolor{casbg}62.74 & \cellcolor{casbg}62.65 & \cellcolor{casbg}62.01 & \cellcolor{casbg}62.32 & \cellcolor{casbg}61.84\\
\midrule
\multirow{4}{*}{\rotatebox[origin=c]{90}{ECE}}
& base & 5.23 & 8.28 & 4.77 & 4.33 & 4.31 & 6.67 & 10.99 & 4.97 & 4.34 & 4.87 \\
& + SaLS~\cite{murugesan2024robust} & 4.97 & 9.04 & 4.56 & 4.25 & 4.26 & 6.17 & 11.06 & 4.38 & 4.80 & 5.02 \\
& \cellcolor{casbg}+ \texttt{CoTS} & \cellcolor{casbg}4.53 & \cellcolor{casbg}3.90 & \cellcolor{casbg}4.55 & \cellcolor{casbg}4.38 & \cellcolor{casbg}4.44 & \cellcolor{casbg}4.75 & \cellcolor{casbg}5.14 & \cellcolor{casbg}4.14 & \cellcolor{casbg}4.43 & \cellcolor{casbg}4.37\\
& \cellcolor{casbg}+ \texttt{E-CoTS} & \cellcolor{casbg}4.16 & \cellcolor{casbg}3.86 & \cellcolor{casbg}4.30 & \cellcolor{casbg}3.85 & \cellcolor{casbg}4.20 & \cellcolor{casbg}5.14 & \cellcolor{casbg}5.39 & \cellcolor{casbg}4.55 & \cellcolor{casbg}4.87 & \cellcolor{casbg}4.58 \\
\bottomrule
\end{tabular}
}
\end{table}

\textbf{Plug-and-play compatibility with calibration methods.}
Table~\ref{tab:cots_ablation_full} evaluates whether the proposed \cots~and \ecots~can be integrated with different base methods.
Overall, \ecots~ consistently improves accuracy across all base methods, showing that the weak-strong ensemble strategy is broadly compatible with prior approaches.
For example, when combined with O-TPT~\cite{sharifdeen2025otpt}, \ecots~increases the accuracy on ImageNet variants from 57.51\% to 62.01\%, showing its effectiveness under distribution shifts.
For calibration, \cots~and \ecots~generally reduce ECE compared with the corresponding base methods, especially for Penalty~\cite{murugesan2024robust} and C-TPT~\cite{yoon2024ctpt}.
These results demonstrate that our method is not restricted to TPT, but can serve as a plug-and-play component to improve the accuracy-calibration trade-off of existing calibration methods.

\subsection{Ablation Studies}
We conduct ablation studies on both temperature scaling and the ensemble design.
For temperature scaling in \cots, we analyze the choice of augmented views and the choice of confidence alignment loss.
For the ensemble strategy in \ecots, we study the effect of different ensemble weights.
We also provide a component-level ablation in Appendix~\ref{sec:ensemble}, showing that ensemble-only improves accuracy but may harm calibration, whereas \ecots~achieves a better accuracy-calibration trade-off.

\begin{figure}[t]
    \centering
    \begin{subfigure}{0.58\linewidth}
        \centering
        \includegraphics[width=\linewidth,height=4cm]{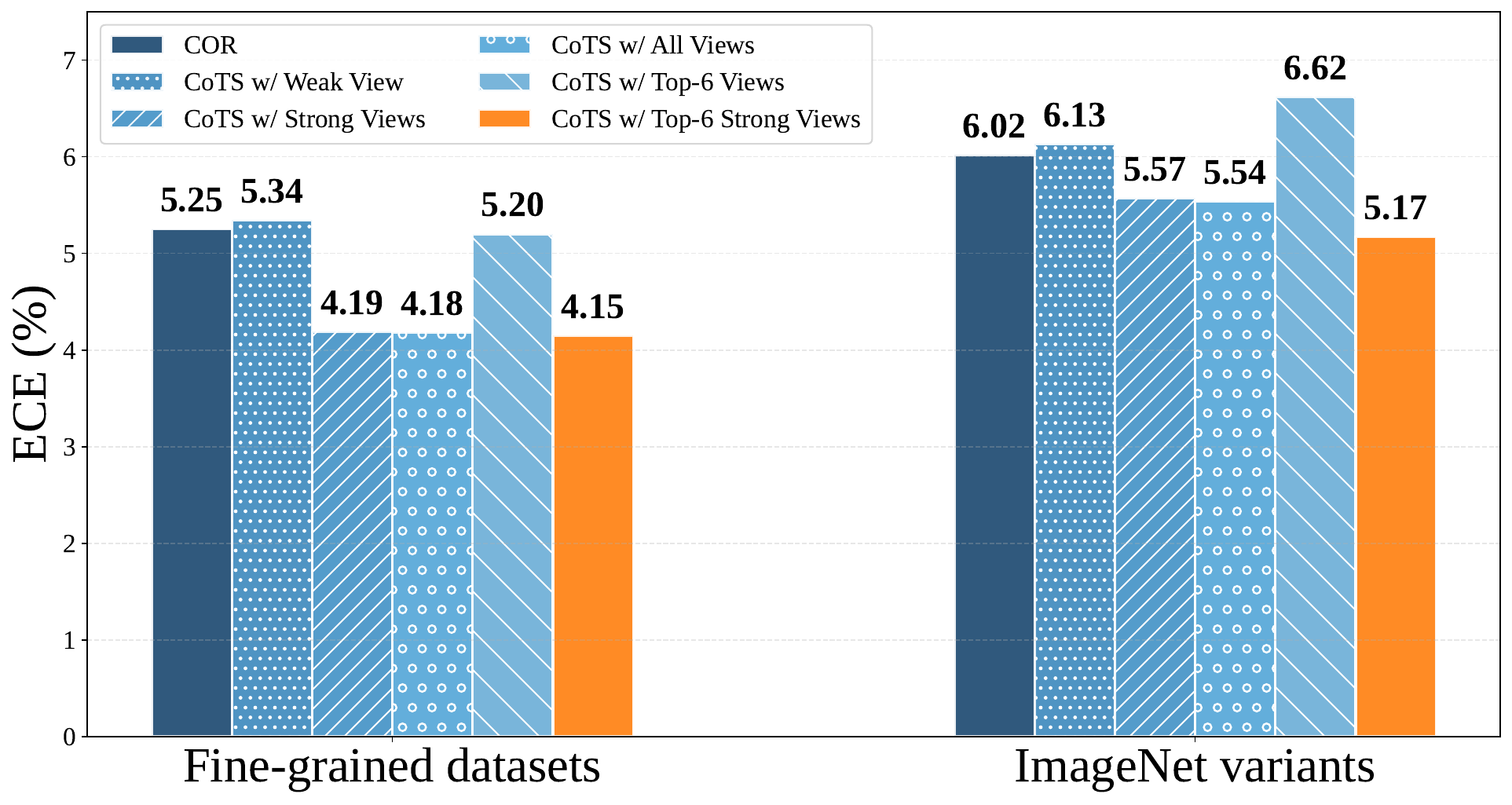}
        \caption{View selection for \cots.}
        \label{fig:view_num}
    \end{subfigure}
    \hfill
    \begin{subfigure}{0.40\linewidth}
        \centering
        \includegraphics[width=\linewidth,height=4cm]{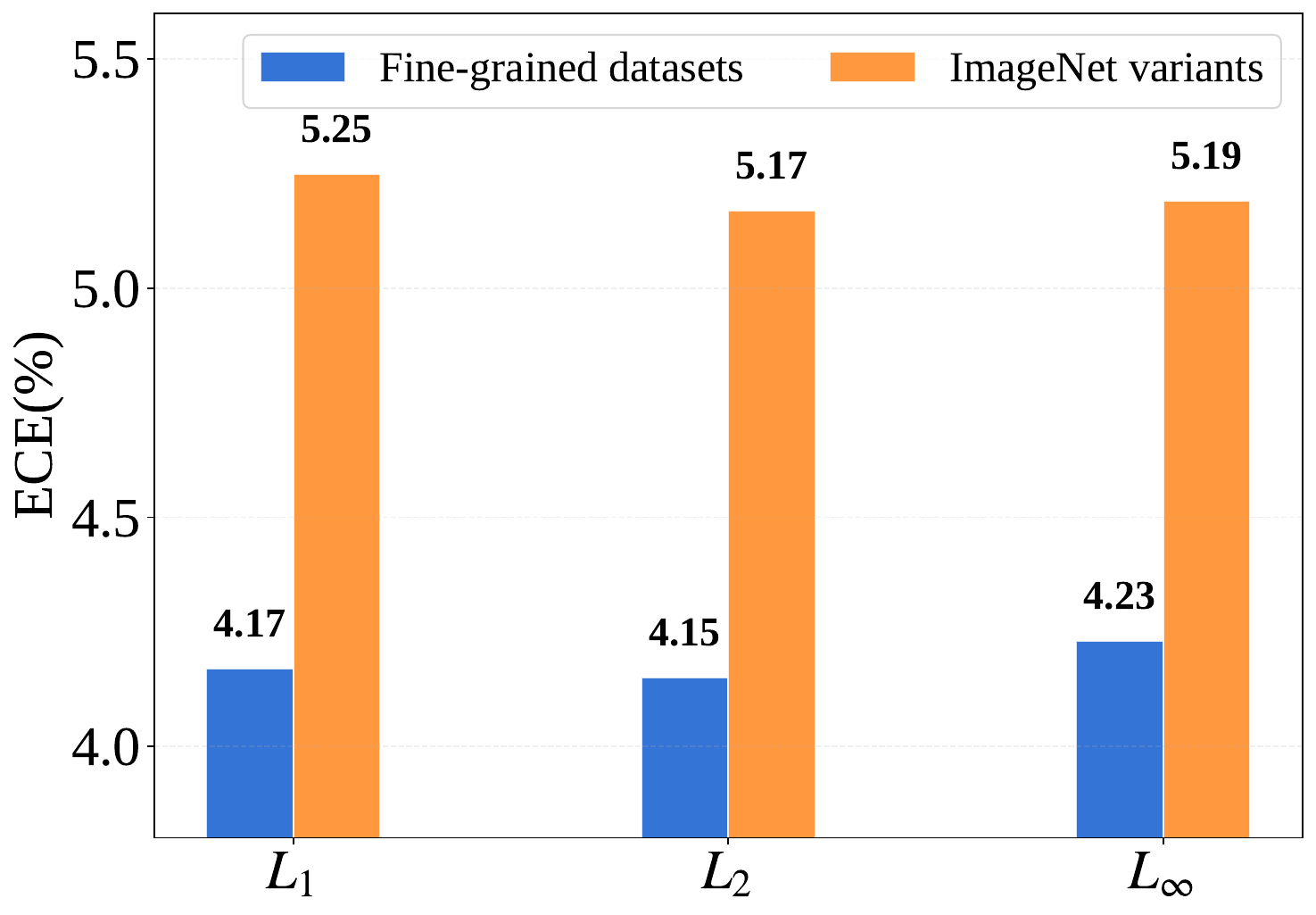}
        \caption{Loss function for \cots.}
        \label{fig:loss_function}
    \end{subfigure}
    \caption{Ablation studies on key components (\ie, view selection and loss function) of \cots~with ViT-B/16. Average ECE (\%) is reported on fine-grained datasets and ImageNet variants.}
\end{figure}

\textbf{View selection in \cots.}
We analyze how different augmented view selections affect temperature learning, as illustrated in Fig.~\ref{fig:view_num}.
Using only the weak view yields limited calibration improvement, with ECE even higher than \texttt{CoR}, indicating that a single weak augmentation does not provide enough confidence variation for reliable temperature estimation.
In contrast, using strong augmented views substantially reduces ECE, indicating that strong augmentations offer more informative confidence variations for alignment.
However, simply using the Top-6 views is not optimal, since the selected set may still contain weak or overly confident views that provide a limited calibration signal.
Among all settings, \cots~with the Top-6 strong views achieves the best calibration performance, obtaining the lowest ECE on both dataset groups.

\textbf{Loss function in \cots.}
We further compare different loss functions for confidence alignment in Fig.~\ref{fig:loss_function}.
The results show that all three losses achieve comparable calibration performance, indicating that \cots~is not highly sensitive to the specific loss formulation.
Among them, the $\mathcal{L}_2$ loss obtains the lowest ECE on both fine-grained datasets and ImageNet variants.
Therefore, we adopt $\mathcal{L}_2$ as the default objective for temperature learning in our experiments.

\textbf{Ensemble weight $\alpha$ in \ecots.}
We study the influence of the ensemble weight $\alpha$ in \ecots, with results shown in Fig.~\ref{fig:alpha_fg_imagenet}.
Using fixed weights underperforms the proposed adaptive weighting strategy, particularly on ImageNet variants (see Appendix~\ref{sec:alpha_ig}).
The adaptive strategy achieves a favorable balance between accuracy and calibration, avoiding the need for manual weight selection.

\subsection{Robustness Analysis}

\begin{table*}[t]
\centering
\caption{Performance comparison of different methods initialized with CoOp~\cite{zhou2022learning} using ViT-B/16. We report average accuracy (\%) and ECE (\%) on fine-grained datasets and ImageNet variants, and evaluate the effect of integrating \texttt{CoTS} and \texttt{E-CoTS}.}
\label{tab:coop_cots_ablation_full}
\renewcommand{\arraystretch}{1.1}
\setlength{\tabcolsep}{3.8pt}
\small
\begin{tabular}{llcccccccccc}
\toprule
\multirow{3}{*}{(\%)} & \multirow{3}{*}{Method}
& \multicolumn{5}{c}{Fine-grained datasets}
& \multicolumn{5}{c}{ImageNet variants} \\
\cmidrule(lr){3-7} \cmidrule(lr){8-12}
&
& \shortstack[c]{TPT}
& \shortstack[c]{C-TPT}
& \shortstack[c]{O-TPT}
& \shortstack[c]{A-TPT}
& \shortstack[c]{SoC}
& \shortstack[c]{TPT}
& \shortstack[c]{C-TPT}
& \shortstack[c]{O-TPT}
& \shortstack[c]{A-TPT}
& \shortstack[c]{SoC}  \\
& &\cite{shu2022testtime}& \cite{yoon2024ctpt} & \cite{sharifdeen2025otpt} & \cite{ahamed2026atpt} & \cite{fillioux2026soc} &\cite{shu2022testtime} & \cite{yoon2024ctpt} & \cite{sharifdeen2025otpt} & \cite{ahamed2026atpt} & \cite{fillioux2026soc} \\
\midrule
\multirow{2}{*}{\rotatebox[origin=c]{90}{ACC}}
& base & 65.67 & 65.12 & 64.58 & 64.68 & 64.23 & 63.08 & 61.51 & 59.16 & 60.48 & 61.45\\
& \cellcolor{casbg}+ \texttt{E-CoTS} & \cellcolor{casbg}65.47 & \cellcolor{casbg}65.70 & \cellcolor{casbg}65.60 & \cellcolor{casbg}65.54 & \cellcolor{casbg}65.19 & \cellcolor{casbg}64.45 & \cellcolor{casbg}64.15 & \cellcolor{casbg}63.36 & \cellcolor{casbg}63.88 & \cellcolor{casbg}64.24 \\
\midrule
\multirow{4}{*}{\rotatebox[origin=c]{90}{ECE}}
& base & 17.46 & 10.56 & 8.43  & 9.29  & 5.61  & 19.95 & 15.38 & 9.92  & 12.20 & 10.22 \\
& + SALS~\cite{murugesan2024robust} & 13.83 & 6.76  & 5.38  & 6.05  & 5.44  & 16.00 & 9.94  & 5.46  & 7.26  & 8.13  \\
& \cellcolor{casbg}+ \texttt{CoTS}   & \cellcolor{casbg}5.61  & \cellcolor{casbg}5.22  & \cellcolor{casbg}5.18  & \cellcolor{casbg}5.11  & \cellcolor{casbg}5.23  & \cellcolor{casbg}5.06  & \cellcolor{casbg}4.62  & \cellcolor{casbg}4.68  & \cellcolor{casbg}4.65  & \cellcolor{casbg}4.44  \\
& \cellcolor{casbg}+ \texttt{E-CoTS} & \cellcolor{casbg}6.10  & \cellcolor{casbg}5.37  & \cellcolor{casbg}5.30  & \cellcolor{casbg}5.12  & \cellcolor{casbg}4.98  & \cellcolor{casbg}6.44  & \cellcolor{casbg}5.26  & \cellcolor{casbg}4.26  & \cellcolor{casbg}4.80  & \cellcolor{casbg}4.80 \\
\bottomrule
\end{tabular}
\vspace{-10pt}
\end{table*}

\textbf{Compatibility with CoOp initialization.}
We evaluate our method initialized with CoOp~\cite{zhou2022learning}, with the results shown in Table~\ref{tab:coop_cots_ablation_full}.
Compared with the base results of existing regularization-based methods, \cots~ consistently achieves much lower ECE across both fine-grained datasets and ImageNet variants.
Meanwhile, \ecots~ generally improves accuracy over the corresponding base methods, showing that the weak-strong ensemble remains beneficial under CoOp~\cite{zhou2022learning} initialization.
Overall, these results demonstrate that our method is compatible with stronger prompt initialization and provides stronger calibration performance than existing regularization-based approaches.
We also analyze the sensitivity of our method to hand-crafted prompt initialization in Appendix~\ref{sec:sens_to_prompt}.

\begin{table*}[t]
\centering
\scriptsize
\renewcommand{\arraystretch}{1.1}

\begin{minipage}[t]{0.53\linewidth}
\centering
\vspace{0pt}

\includegraphics[width=\linewidth]{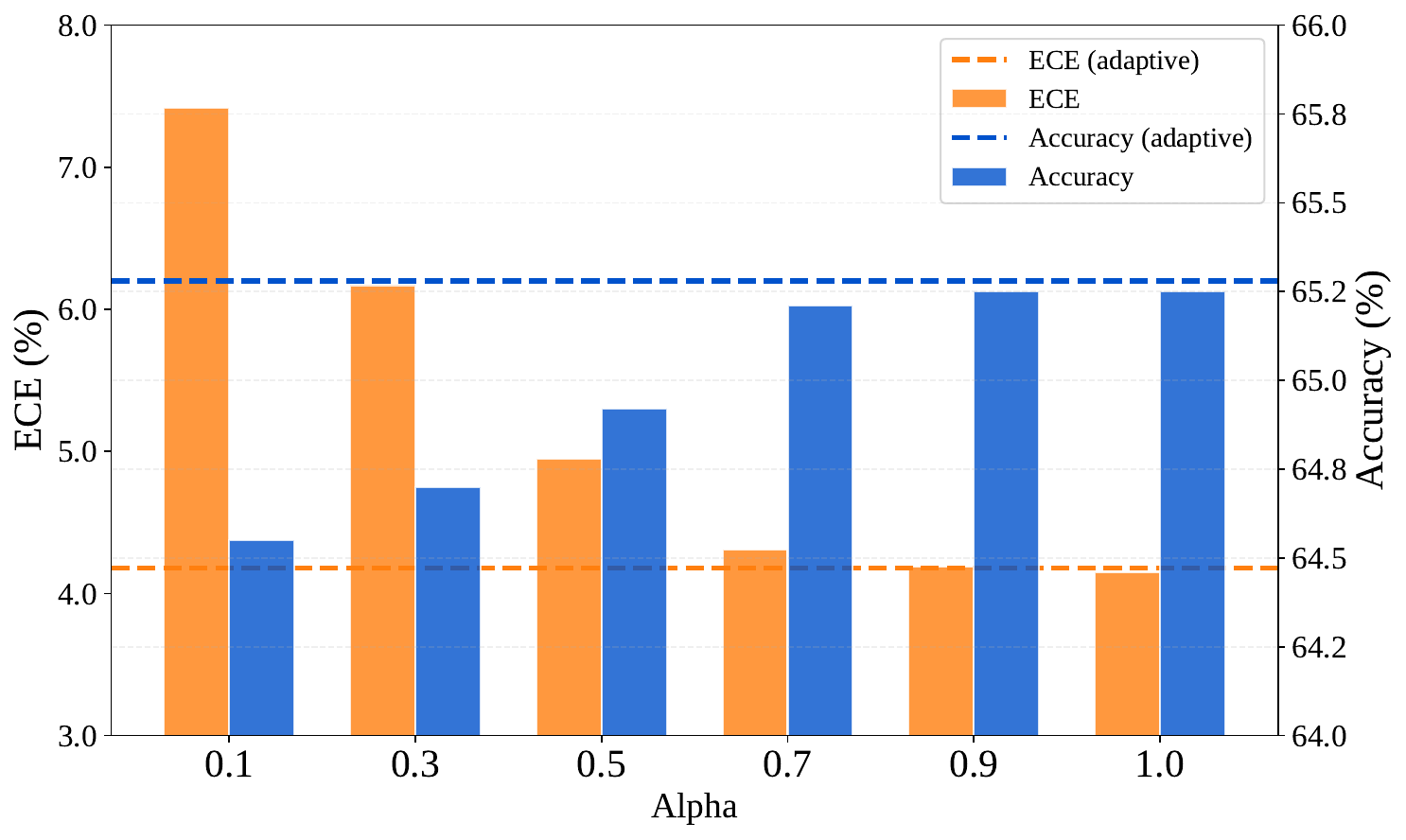}

\captionof{figure}{
Effect of the ensemble weight $\alpha$ on \ecots~using ViT-B/16.
Average accuracy (\%) and ECE (\%) on fine-grained datasets are reported.
}
\label{fig:alpha_fg_imagenet}
\end{minipage}
\hfill
\begin{minipage}[t]{0.46\linewidth}
\centering
\vspace{0pt}

\captionof{table}{
Evaluation of SaLS~\cite{murugesan2024robust}, \cots, and \ecots~on TTL~\cite{imam2025test} and TPS~\cite{sui2025just} using ViT-B/16. Average accuracy (\%) and ECE (\%) on fine-grained datasets and ImageNet variants are reported.
}
\label{tab:ttl_tps}

\setlength{\tabcolsep}{3pt}
\resizebox{\linewidth}{!}{%
\begin{tabular}{llcccc}
\toprule
\multirow{3}{*}{(\%)}
& \multirow{3}{*}{Method}
& \multicolumn{2}{c}{Fine-grained datasets}
& \multicolumn{2}{c}{ImageNet variants} \\
\cmidrule(lr){3-4} \cmidrule(lr){5-6}
& & TTL & TPS & TTL & TPS \\
& & \cite{imam2025test} & \cite{sui2025just}
& \cite{imam2025test} & \cite{sui2025just} \\
\midrule

\multirow{2}{*}{\rotatebox[origin=c]{90}{ACC}}
& base
& 64.48 & 64.99 & 62.46 & 61.69 \\
& \cellcolor{casbg}+ \ecots
& \cellcolor{casbg}64.23
& \cellcolor{casbg}65.11
& \cellcolor{casbg}62.39
& \cellcolor{casbg}61.71 \\

\midrule

\multirow{4}{*}{\rotatebox[origin=c]{90}{ECE}}
& base
& 28.60 & 16.79 & 34.30 & 27.16 \\
& + SaLS~\cite{murugesan2024robust}
& 22.15 & 13.69 & 28.94 & 23.98 \\
& \cellcolor{casbg}+ \cots
& \cellcolor{casbg}9.82
& \cellcolor{casbg}4.87
& \cellcolor{casbg}9.42
& \cellcolor{casbg}6.67 \\
& \cellcolor{casbg}+ \ecots
& \cellcolor{casbg}9.42
& \cellcolor{casbg}5.02
& \cellcolor{casbg}8.66
& \cellcolor{casbg}6.65 \\

\bottomrule
\end{tabular}%
}
\end{minipage}

\end{table*}

\textbf{Generalization to other episodic TTA methods}
We further evaluate \cots~and \ecots~on other episodic TTA methods.
The results in Table~\ref{tab:ttl_tps} show that our method generalizes well beyond TPT-based approaches.
In particular, \cots~consistently brings large reductions in ECE on both datasets, substantially outperforming the corresponding base models and also achieving stronger calibration than SaLS~\cite{murugesan2024robust}.
Furthermore, \ecots~maintains comparable, and sometimes improved accuracy over the base methods.
These results indicate that our approach is not limited to a specific adaptation framework, but can serve as an effective calibration strategy for other episodic TTA methods as well.

\textbf{Overall comparison across backbones.}
We summarize the average performance on fine-grained datasets and ImageNet variants using both ViT-B/16 and ResNet-50 in Table~\ref{tab:backbone_comparison}.
The improvements are consistent across both backbones, demonstrating the robustness of \cots~and \ecots.
Overall, both \cots~and \ecots~provide a superior accuracy-calibration trade-off compared to prior methods.

\begin{table*}[t]
\centering
\caption{Performance comparison of different methods using ViT-B/16 and ResNet-50.
Accuracy (\%) and ECE (\%) are reported on fine-grained datasets and ImageNet variants, respectively.
}
\label{tab:backbone_comparison}
\renewcommand{\arraystretch}{1.12}
\setlength{\tabcolsep}{5pt}
\scriptsize
\resizebox{\linewidth}{!}{
\begin{tabular}{lcccccccc}
\toprule
\multirow{3}{*}{Method}
& \multicolumn{4}{c}{ViT-B/16}
& \multicolumn{4}{c}{ResNet-50} \\
\cmidrule(lr){2-5} \cmidrule(lr){6-9}
& \multicolumn{2}{c}{Fine-grained datasets}
& \multicolumn{2}{c}{ImageNet variants}
& \multicolumn{2}{c}{Fine-grained datasets}
& \multicolumn{2}{c}{ImageNet variants} \\
\cmidrule(lr){2-3} \cmidrule(lr){4-5}
\cmidrule(lr){6-7} \cmidrule(lr){8-9}
& ACC (\%) & ECE (\%) & ACC (\%) & ECE (\%) & ACC (\%) & ECE (\%) & ACC (\%) & ECE (\%) \\
\midrule
Zero-Shot~\cite{radford2021learning} & 63.95 & 4.33  & 57.22 & 4.98  & 56.04 & 5.38  & 40.70 & 7.09  \\
TPT~\cite{shu2022testtime} & 65.19 & 11.30 & 60.74 & 11.90 & 58.06 & 11.31 & 43.83 & 17.55 \\
C-TPT~\cite{yoon2024ctpt} & 64.57 & 5.23  & 59.25 & 6.67  & 57.69 & 6.44  & 42.75 & 13.14 \\
Penalty~\cite{murugesan2024robust} & 64.21 & 8.28  & 60.24 & 10.99 & 57.52 & 8.41  & 43.66 & 17.02 \\
SALS~\cite{murugesan2024robust} & 65.20 & 9.94  & 60.74 & 10.98 & 58.06 & 9.31  & 43.83 & 15.37 \\
O-TPT~\cite{sharifdeen2025otpt} & 64.05 & 4.77  & 57.51 & 4.97  & 57.36 & 5.67  & 40.82 & 6.46  \\
A-TPT~\cite{ahamed2026atpt} & 64.12 & 4.33  & 58.20 & 4.34  & 57.30 & 4.55  & 41.94 & 7.76  \\
SoC~\cite{fillioux2026soc}  & 63.95 & 4.31  & 58.12 & 4.87  & 56.19 & 5.50  & 41.61 & 7.52  \\
\rowcolor{casbg} \texttt{\hspace*{0pt}CoTS}  & 65.19 & 4.28  & 60.74 & 5.30  & 58.06 & 5.04  & 43.83 & 6.87  \\
\rowcolor{casbg} \texttt{\hspace*{0pt}E-CoTS}   & 65.22 & 4.31  & 62.95 & 5.38  & 57.72 & 5.04  & 45.39 & 7.24 \\
\bottomrule
\end{tabular}
}
\end{table*}

\section{Conclusion}
This paper studies the calibration problem in test-time prompt tuning from the perspective of confidence correction.
We observe that existing calibration-oriented methods often reduce ECE at the cost of classification accuracy.
To address this trade-off, we propose \cots, a post-hoc confidence-based temperature scaling method that uses zero-shot CLIP predictions as a calibration anchor.
By only rescaling logits, \cots~improves calibration performance without changing the discriminative benefits of TPT.
We further introduce a weak-strong ensemble strategy and integrate it into \cots, resulting in \ecots, which improves accuracy while maintaining reliable calibration.
Extensive experiments across multiple datasets and backbones validate that our methods effectively mitigate miscalibration without compromising primary accuracy.

{
\small
\bibliographystyle{abbrv}
\bibliography{nips}

@String(IJCV  = {Int. J. Comput. Vis.})

@String(CVPR  = {Proc. CVPR})

@String(ICCV  = {Proc. ICCV})

@String(ECCV  = {Proc. ECCV})

@String(NeurIPS = {Proc. NeurIPS})

@String{IGARSS ={IGARSS} }

@String(NIPS = {Proc. NeurIPS})

@String(ICML  = {Proc. ICML})

@String(ICLR  = {Proc. ICLR})

@String(CVPRW = {Proc. CVPR Workshops})

@String(ICCVW = {Proc. ICCV Workshops})

@String(AAAI  = {Proc. AAAI})

@String(ACL  = {Proc. ACL})

@String(EMNLP = {Proc. EMNLP})

@String(WACV = {Proc. WACV})

@String(ICVGIP = {Proc. ICVGIP})

@String(MICCAI = {Proc. MICCAI})

@inproceedings{sharifdeen2025otpt,
  title={O-tpt: Orthogonality constraints for calibrating test-time prompt tuning in vision-language models},
  author={Sharifdeen, Ashshak and Munir, Muhammad Akhtar and Baliah, Sanoojan and Khan, Salman and Khan, Muhammad Haris},
  booktitle=CVPR,
  pages={19942--19951},
  year={2025}
}

@inproceedings{yoon2024ctpt,
  title={C-{TPT}: Calibrated Test-Time Prompt Tuning for Vision-Language Models via Text Feature Dispersion},
  author={Hee Suk Yoon and Eunseop Yoon and Joshua Tian Jin Tee and Mark A. Hasegawa-Johnson and Yingzhen Li and Chang D. Yoo},
  booktitle=ICLR,
  year={2024}
}

@inproceedings{ahamed2026atpt,
  title={A-{TPT}: Angular Diversity Calibration Properties for Test-Time Prompt Tuning of Vision-Language Models},
  author={Shihab Aaqil Ahamed and Udaya Sampath K. Perera Miriya Thanthrige and Ranga Rodrigo and Muhammad Haris Khan},
  booktitle=ICLR,
  year={2026}
}

@inproceedings{guo2017calibration,
  title={On calibration of modern neural networks},
  author={Guo, Chuan and Pleiss, Geoff and Sun, Yu and Weinberger, Kilian Q},
  booktitle=ICML,
  pages={1321--1330},
  year={2017}
}

@inproceedings{shu2022testtime,
  title={Test-Time Prompt Tuning for Zero-Shot Generalization in Vision-Language Models},
  author={Manli Shu and Weili Nie and De-An Huang and Zhiding Yu and Tom Goldstein and Anima Anandkumar and Chaowei Xiao},
  booktitle=NIPS,
  pages={14274--14289},
  year={2022}
}

@inproceedings{hendrycks2020augmix,
  title={Augmix: A simple data processing method to improve robustness and uncertainty},
  author={Hendrycks, Dan and Mu, Norman and Cubuk, Ekin D and Zoph, Barret and Gilmer, Justin and Lakshminarayanan, Balaji},
  booktitle=ICLR,
  year={2020}
}

@inproceedings{fillioux2026soc,
  title={SoC: Semantic Orthogonal Calibration for Test-Time Prompt Tuning},
  author={Leo Fillioux and Omprakash Chakraborty and Ismail Ben Ayed and Paul-Henry Cournède and Stergios Christodoulidis and Maria Vakalopoulou and Jose Dolz},
  booktitle=CVPR,
  year={2026}
}

@article{liang2025comprehensive,
  title={A comprehensive survey on test-time adaptation under distribution shifts},
  author={Liang, Jian and He, Ran and Tan, Tieniu},
  journal=IJCV,
  volume={133},
  number={1},
  pages={31--64},
  year={2025},
  publisher={Springer}
}

@inproceedings{radford2021learning,
  title={Learning transferable visual models from natural language supervision},
  author={Radford, Alec and Kim, Jong Wook and Hallacy, Chris and Ramesh, Aditya and Goh, Gabriel and Agarwal, Sandhini and Sastry, Girish and Askell, Amanda and Mishkin, Pamela and Clark, Jack and others},
  booktitle=ICML,
  pages={8748--8763},
  year={2021}
}

@inproceedings{murugesan2024robust,
  title={Robust calibration of large vision-language adapters},
  author={Murugesan, Balamurali and Silva-Rodr{\'\i}guez, Julio and Ayed, Ismail Ben and Dolz, Jose},
  booktitle=ECCV,
  pages={147--165},
  year={2024}
}

@article{zhou2022learning,
  title={Learning to prompt for vision-language models},
  author={Zhou, Kaiyang and Yang, Jingkang and Loy, Chen Change and Liu, Ziwei},
  journal=IJCV,
  volume={130},
  number={9},
  pages={2337--2348},
  year={2022},
  publisher={Springer}
}

@inproceedings{dafnis2025testtime,
  title={Test-Time Spectrum-Aware Latent Steering for Zero-Shot Generalization in Vision-Language Models},
  author={Konstantinos M. Dafnis and Dimitris N. Metaxas},
  booktitle=NIPS,
  pages={151169--151194},
  year={2025}
}

@inproceedings{sheng2025r,
  title={R-tpt: Improving adversarial robustness of vision-language models through test-time prompt tuning},
  author={Sheng, Lijun and Liang, Jian and Wang, Zilei and He, Ran},
  booktitle=CVPR,
  pages={29958--29967},
  year={2025}
}

@inproceedings{wang2024open,
  title={Open-vocabulary calibration for fine-tuned CLIP},
  author={Wang, Shuoyuan and Wang, Jindong and Wang, Guoqing and Zhang, Bob and Zhou, Kaiyang and Wei, Hongxin},
  booktitle=ICML,
  pages={51734--51754},
  year={2024}
}

@inproceedings{wang2025understanding,
  title={Understanding and Mitigating Miscalibration in Prompt Tuning for Vision-Language Models},
  author={Wang, Shuoyuan and Li, Yixuan and Wei, Hongxin},
  booktitle=ICML,
  pages={63467--63489},
  year={2025}
}

@inproceedings{minderer2021revisiting,
  title={Revisiting the Calibration of Modern Neural Networks},
  author={Minderer, Matthias and Djolonga, Josip and Romijnders, Rob and Hubis, Frances Ann and Zhai, Xiaohua and Houlsby, Neil and Tran, Dustin and Lucic, Mario},
  booktitle=NIPS,
  pages={15682--15694},
  year={2021}
}

@article{platt1999probabilistic,
  title={Probabilistic outputs for support vector machines and comparisons to regularized likelihood methods},
  author={Platt, John and others},
  journal={Advances in Large Margin Classifiers},
  volume={10},
  number={3},
  pages={61--74},
  year={1999},
  publisher={Cambridge, MA}
}

@inproceedings{liu2022devil,
  title={The devil is in the margin: Margin-based label smoothing for network calibration},
  author={Liu, Bingyuan and Ben Ayed, Ismail and Galdran, Adrian and Dolz, Jose},
  booktitle=CVPR,
  pages={80--88},
  year={2022}
}

@inproceedings{hu2024pseudo,
  title={Pseudo-calibration: Improving predictive uncertainty estimation in unsupervised domain adaptation},
  author={Hu, Dapeng and Liang, Jian and Wang, Xinchao and Foo, Chuan-Sheng},
  booktitle=ICML,
  pages ={19304--19326},
  year={2024}
}

@inproceedings{naeini2015obtaining,
  title={Obtaining well calibrated probabilities using bayesian binning},
  author={Naeini, Mahdi Pakdaman and Cooper, Gregory and Hauskrecht, Milos},
  booktitle=AAAI,
  pages={2901--2907},
  year={2015}
}

@inproceedings{khattak2023maple,
  title={Maple: Multi-modal prompt learning},
  author={Khattak, Muhammad Uzair and Rasheed, Hanoona and Maaz, Muhammad and Khan, Salman and Khan, Fahad Shahbaz},
  booktitle=CVPR,
  pages={19113--19122},
  year={2023}
}

@inproceedings{jia2021scaling,
  title={Scaling up visual and vision-language representation learning with noisy text supervision},
  author={Jia, Chao and Yang, Yinfei and Xia, Ye and Chen, Yi-Ting and Parekh, Zarana and Pham, Hieu and Le, Quoc and Sung, Yun-Hsuan and Li, Zhen and Duerig, Tom},
  booktitle=ICML,
  pages={4904--4916},
  year={2021}
}

@inproceedings{galil2023what,
    title={What Can we Learn From The Selective Prediction And Uncertainty Estimation Performance Of 523 Imagenet Classifiers?},
    author={Ido Galil and Mohammed Dabbah and Ran El-Yaniv},
    booktitle=ICLR,
    year={2023}
}

@article{han2025d,
  title={D-TPT: Dimensional Entropy Maximization for Calibrating Test-Time Prompt Tuning in Vision-Language Models},
  author={Han, Jisu and Hwang, Wonjun},
  journal={arXiv preprint arXiv:2510.09473},
  year={2025}
}

@inproceedings{deng2009imagenet,
  title={ImageNet: A large-scale hierarchical image database},
  author={Deng, Jia and Dong, Wei and Socher, Richard and Li, Li-Jia and Kai Li and Li Fei-Fei},
  booktitle=CVPR,
  pages={248-255},
  year={2009}
}

@article{huang2022unsupervised,
  title={Unsupervised prompt learning for vision-language models},
  author={Huang, Tony and Chu, Jack and Wei, Fangyun},
  journal={arXiv preprint arXiv:2204.03649},
  year={2022}
}

@inproceedings{thulasidasan2019mixup,
  title={On mixup training: improved calibration and predictive uncertainty for deep neural networks},
  author={Thulasidasan, Sunil and Chennupati, Gopinath and Bilmes, Jeff and Bhattacharya, Tanmoy and Michalak, Sarah},
  booktitle=NIPS,
  pages={13911--13922},
  year={2019}
}

@inproceedings{lin2017focal,
  title={Focal loss for dense object detection},
  author={Lin, Tsung-Yi and Goyal, Priya and Girshick, Ross and He, Kaiming and Doll{\'a}r, Piotr},
  booktitle=ICCV,
  pages={2980--2988},
  year={2017}
}

@inproceedings{kull2019beyond,
  title={Beyond temperature scaling: obtaining well-calibrated multiclass probabilities with Dirichlet calibration},
  author={Kull, Meelis and Perello-Nieto, Miquel and K{\"a}ngsepp, Markus and Filho, Telmo Silva and Song, Hao and Flach, Peter},
  booktitle=NIPS,
  pages={12316--12326},
  year={2019}
}

@inproceedings{ding2021local,
  title={Local temperature scaling for probability calibration},
  author={Ding, Zhipeng and Han, Xu and Liu, Peirong and Niethammer, Marc},
  booktitle=ICCV,
  pages={6889--6899},
  year={2021}
}

@article{wang2023calibration,
  title={Calibration in deep learning: A survey of the state-of-the-art},
  author={Wang, Cheng},
  journal={arXiv preprint arXiv:2308.01222},
  year={2023}
}

@inproceedings{zhou2022conditional,
  title={Conditional prompt learning for vision-language models},
  author={Zhou, Kaiyang and Yang, Jingkang and Loy, Chen Change and Liu, Ziwei},
  booktitle=CVPR,
  pages={16816--16825},
  year={2022}
}

@inproceedings{imam2025test,
  title={Test-time low rank adaptation via confidence maximization for zero-shot generalization of vision-language models},
  author={Imam, Raza and Gani, Hanan and Huzaifa, Muhammad and Nandakumar, Karthik},
  booktitle=WACV,
  pages={5449--5459},
  year={2025}
}

@inproceedings{farina2024frustratingly,
  title={Frustratingly Easy Test-Time Adaptation of Vision-Language Models},
  author={Matteo Farina and Gianni Franchi and Giovanni Iacca and Massimiliano Mancini and Elisa Ricci},
  booktitle=NIPS,
  pages={129062--129093},
  year={2024}
}

@inproceedings{zanella2024test,
  title={On the test-time zero-shot generalization of vision-language models: Do we really need prompt learning?},
  author={Zanella, Maxime and Ben Ayed, Ismail},
  booktitle=CVPR,
  pages={23783--23793},
  year={2024}
}

@inproceedings{sui2025just,
  title={Just shift it: Test-time prototype shifting for zero-shot generalization with vision-language models},
  author={Sui, Elaine and Wang, Xiaohan and Yeung-Levy, Serena},
  booktitle=WACV,
  pages={825--835},
  year={2025}
}

@inproceedings{koleilat2025biomedcoop,
  title={Biomedcoop: Learning to prompt for biomedical vision-language models},
  author={Koleilat, Taha and Asgariandehkordi, Hojat and Rivaz, Hassan and Xiao, Yiming},
  booktitle=CVPR,
  pages={14766--14776},
  year={2025}
}

@inproceedings{silva2025few,
  title={Few-shot, now for real: Medical VLMs adaptation without balanced sets or validation},
  author={Silva-Rodr{\'\i}guez, Julio and Shakeri, Fereshteh and Bahig, Houda and Dolz, Jose and Ben Ayed, Ismail},
  booktitle=MICCAI,
  pages={237--247},
  year={2025}
}

@inproceedings{khandelwal2022simple,
  title={Simple but effective: Clip embeddings for embodied ai},
  author={Khandelwal, Apoorv and Weihs, Luca and Mottaghi, Roozbeh and Kembhavi, Aniruddha},
  booktitle=CVPR,
  pages={14829--14838},
  year={2022}
}

@inproceedings{pan2024vlp,
  title={Vlp: Vision language planning for autonomous driving},
  author={Pan, Chenbin and Yaman, Burhaneddin and Nesti, Tommaso and Mallik, Abhirup and Allievi, Alessandro G and Velipasalar, Senem and Ren, Liu},
  booktitle=CVPR,
  pages={14760--14769},
  year={2024}
}

@article{lialin2023scaling,
  title={Scaling down to scale up: A guide to parameter-efficient fine-tuning},
  author={Lialin, Vladislav and Deshpande, Vijeta and Yao, Xiaowei and Rumshisky, Anna},
  journal={arXiv preprint arXiv:2303.15647},
  year={2023}
}

@article{han2024parameter,
  title={Parameter-Efficient Fine-Tuning for Large Models: A Comprehensive Survey},
  author={Han, Zeyu and Gao, Chao and Liu, Jinyang and Zhang, Jeff and Zhang, Sai Qian},
  journal={Transactions on Machine Learning Research},
  year={2024}
}

@inproceedings{lester2021power,
  title={The power of scale for parameter-efficient prompt tuning},
  author={Lester, Brian and Al-Rfou, Rami and Constant, Noah},
  booktitle=EMNLP,
  pages={3045--3059},
  year={2021}
}

@inproceedings{jia2022visual,
  title={Visual prompt tuning},
  author={Jia, Menglin and Tang, Luming and Chen, Bor-Chun and Cardie, Claire and Belongie, Serge and Hariharan, Bharath and Lim, Ser-Nam},
  booktitle=ECCV,
  pages={709--727},
  year={2022}
}

@inproceedings{tanwisuth2023pouf,
  title={Pouf: Prompt-oriented unsupervised fine-tuning for large pre-trained models},
  author={Tanwisuth, Korawat and Zhang, Shujian and Zheng, Huangjie and He, Pengcheng and Zhou, Mingyuan},
  booktitle=ICML,
  pages={33816--33832},
  year={2023}
}

@inproceedings{liang2024realistic,
  title={Realistic unsupervised CLIP fine-tuning with universal entropy optimization},
  author={Liang, Jian and Sheng, Lijun and Wang, Zhengbo and He, Ran and Tan, Tieniu},
  booktitle=ICML,
  pages={29667--29681},
  year={2024}
}

@inproceedings{nixon2019measuring,
  title={Measuring calibration in deep learning.},
  author={Nixon, Jeremy and Dusenberry, Michael W and Zhang, Linchuan and Jerfel, Ghassen and Tran, Dustin},
  booktitle=CVPRW,
  year={2019}
}

@inproceedings{muller2019does,
  title={When does label smoothing help?},
  author={M{\"u}ller, Rafael and Kornblith, Simon and Hinton, Geoffrey},
  booktitle=NeurIPS,
  pages={4694--4703},
  year={2019}
}

@inproceedings{blundell2015weight,
  title={Weight uncertainty in neural network},
  author={Blundell, Charles and Cornebise, Julien and Kavukcuoglu, Koray and Wierstra, Daan},
  booktitle=ICML,
  pages={1613--1622},
  year={2015}
}

@inproceedings{lakshminarayanan2017simple,
  title={Simple and scalable predictive uncertainty estimation using deep ensembles},
  author={Lakshminarayanan, Balaji and Pritzel, Alexander and Blundell, Charles},
  booktitle=NeurIPS,
  pages={6405--6416},
  year={2017}
}

@inproceedings{gal2016dropout,
  title={Dropout as a bayesian approximation: Representing model uncertainty in deep learning},
  author={Gal, Yarin and Ghahramani, Zoubin},
  booktitle=ICML,
  pages={1050--1059},
  year={2016}
}

@inproceedings{yao2023visual,
  title={Visual-language prompt tuning with knowledge-guided context optimization},
  author={Yao, Hantao and Zhang, Rui and Xu, Changsheng},
  booktitle=CVPR,
  pages={6757--6767},
  year={2023}
}

@inproceedings{zhang2022memo,
  title={MEMO: test time robustness via adaptation and augmentation},
  author={Zhang, Marvin and Levine, Sergey and Finn, Chelsea},
  booktitle=NIPS,
  pages={38629--38642},
  year={2022}
}

@inproceedings{sun2020test,
  title={Test-time training with self-supervision for generalization under distribution shifts},
  author={Sun, Yu and Wang, Xiaolong and Liu, Zhuang and Miller, John and Efros, Alexei and Hardt, Moritz},
  booktitle=ICML,
  pages={9229--9248},
  year={2020}
}

@inproceedings{xiao2025dynaprompt,
  title={DynaPrompt: Dynamic Test-Time Prompt Tuning},
  author={Zehao Xiao and Shilin Yan and Jack Hong and Jiayin Cai and Xiaolong Jiang and Yao Hu and Jiayi Shen and Cheems Wang and Cees G. M. Snoek},
  booktitle=ICLR,
  year={2025}
}

@inproceedings{feng2023diverse,
  title={Diverse data augmentation with diffusions for effective test-time prompt tuning},
  author={Feng, Chun-Mei and Yu, Kai and Liu, Yong and Khan, Salman and Zuo, Wangmeng},
  booktitle=ICCV,
  pages={2704--2714},
  year={2023}
}

@inproceedings{ma2023swapprompt,
  title={SwapPrompt: test-time prompt adaptation for vision-language models},
  author={Ma, Xiaosong and Zhang, Jie and Guo, Song and Xu, Wenchao},
  booktitle=NIPS,
  pages={65252--65264},
  year={2023}
}

@inproceedings{hendrycks2021natural,
  title={Natural adversarial examples},
  author={Hendrycks, Dan and Zhao, Kevin and Basart, Steven and Steinhardt, Jacob and Song, Dawn},
  booktitle=CVPR,
  pages={15262--15271},
  year={2021}
}

@inproceedings{recht2019imagenet,
  title={Do imagenet classifiers generalize to imagenet?},
  author={Recht, Benjamin and Roelofs, Rebecca and Schmidt, Ludwig and Shankar, Vaishaal},
  booktitle=ICML,
  pages={5389--5400},
  year={2019}
}

@inproceedings{hendrycks2021many,
  title={The many faces of robustness: A critical analysis of out-of-distribution generalization},
  author={Hendrycks, Dan and Basart, Steven and Mu, Norman and Kadavath, Saurav and Wang, Frank and Dorundo, Evan and Desai, Rahul and Zhu, Tyler and Parajuli, Samyak and Guo, Mike and others},
  booktitle=ICCV,
  pages={8340--8349},
  year={2021}
}

@inproceedings{wang2019learning,
  title={Learning robust global representations by penalizing local predictive power},
  author={Wang, Haohan and Ge, Songwei and Xing, Eric P and Lipton, Zachary C},
  booktitle=NIPS,
  pages={10506--10518},
  year={2019}
}

@inproceedings{fei2004learning,
  title={Learning generative visual models from few training examples: An incremental bayesian approach tested on 101 object categories},
  author={Fei-Fei, Li and Fergus, Rob and Perona, Pietro},
  booktitle=CVPRW,
  pages={178--178},
  year={2004}
}

@inproceedings{nilsback2008automated,
  author={Nilsback, Maria-Elena and Zisserman, Andrew},
  booktitle=ICVGIP,
  title={Automated Flower Classification over a Large Number of Classes},
  year={2008},
  pages={722-729}}

@inproceedings{bossard2014food,
  title={Food-101--mining discriminative components with random forests},
  author={Bossard, Lukas and Guillaumin, Matthieu and Van Gool, Luc},
  booktitle=ECCV,
  pages={446--461},
  year={2014}
}

@inproceedings{xiao2010sun,
  author={Xiao, Jianxiong and Hays, James and Ehinger, Krista A. and Oliva, Aude and Torralba, Antonio},
  booktitle=CVPR,
  title={SUN database: Large-scale scene recognition from abbey to zoo},
  year={2010},
  pages={3485-3492}}

@inproceedings{helber2018introducing,
  author={Helber, Patrick and Bischke, Benjamin and Dengel, Andreas and Borth, Damian},
  booktitle=IGARSS,
  title={Introducing Eurosat: A Novel Dataset and Deep Learning Benchmark for Land Use and Land Cover Classification},
  year={2018},
  pages={204-207}}

@article{soomro2012ucf101,
  title={Ucf101: A dataset of 101 human actions classes from videos in the wild},
  author={Soomro, Khurram and Zamir, Amir Roshan and Shah, Mubarak},
  journal={arXiv preprint arXiv:1212.0402},
  year={2012}
}

@inproceedings{krause20133d,
  title={3d object representations for fine-grained categorization},
  author={Krause, Jonathan and Stark, Michael and Deng, Jia and Fei-Fei, Li},
  booktitle=ICCVW,
  pages={554--561},
  year={2013}
}

@article{maji2013fine,
  title={Fine-grained visual classification of aircraft},
  author={Maji, Subhransu and Rahtu, Esa and Kannala, Juho and Blaschko, Matthew and Vedaldi, Andrea},
  journal={arXiv preprint arXiv:1306.5151},
  year={2013}
}

@inproceedings{parkhi2012cats,
  title={Cats and dogs},
  author={Parkhi, Omkar M and Vedaldi, Andrea and Zisserman, Andrew and Jawahar, CV},
  booktitle=CVPR,
  pages={3498--3505},
  year={2012}
}

@inproceedings{cimpoi2014describing,
  title={Describing textures in the wild},
  author={Cimpoi, Mircea and Maji, Subhransu and Kokkinos, Iasonas and Mohamed, Sammy and Vedaldi, Andrea},
  booktitle=CVPR,
  pages={3606--3613},
  year={2014}
}

@article{brier1950verification,
  title={VERIFICATION OF FORECASTS EXPRESSED IN TERMS OF PROBABILITY},
  author={Glenn W. Brier},
  journal={Monthly Weather Review},
  year={1950},
  volume={78},
  pages={1-3}
}

@article{dong2025adapting,
  title={Adapting vision-language models without labels: A comprehensive survey},
  author={Dong, Hao and Sheng, Lijun and Liang, Jian and He, Ran and Chatzi, Eleni and Fink, Olga},
  journal={arXiv preprint arXiv:2508.05547},
  year={2025}
}

@inproceedings{loshchilov2018decoupled,
  title={Decoupled Weight Decay Regularization},
  author={Ilya Loshchilov and Frank Hutter},
  booktitle=ICLR,
  year={2019}
}

@inproceedings{karmanov2024efficient,
  title={Efficient test-time adaptation of vision-language models},
  author={Karmanov, Adilbek and Guan, Dayan and Lu, Shijian and El Saddik, Abdulmotaleb and Xing, Eric},
  booktitle=CVPR,
  pages={14162--14171},
  year={2024}
}

@inproceedings{zhai2023sigmoid,
  title={Sigmoid loss for language image pre-training},
  author={Zhai, Xiaohua and Mustafa, Basil and Kolesnikov, Alexander and Beyer, Lucas},
  booktitle=ICCV,
  pages={11941--11952},
  year={2023}
}

@inproceedings{yan2026if,
  title={What if consensus lies? selective-complementary reinforcement learning at test time},
  author={Yan, Dong and Liang, Jian and Wang, Yanbo and Lu, Shuo and He, Ran and Tan, Tieniu},
  booktitle=ACL,
  pages={28957--28970},
  year={2026}
}
}

\appendix
\section{Limitations and Broader Impact}
\subsection{Limitations}
While our methods effectively improve calibration for test-time prompt tuning, they are currently evaluated on classification tasks and introduce additional computational overhead.
Extending confidence-oriented temperature scaling to other vision-language tasks, such as semantic segmentation, object detection, or adversarial robustness, remains to be explored.

\subsection{Broader Impact}
This work aims to improve the reliability of vision-language models by mitigating the miscalibration issue in test-time prompt tuning.
The proposed methods achieve a favorable balance between accuracy and calibration error.
Moreover, our approaches can be readily integrated with existing calibration techniques, highlighting their potential as general plug-and-play modules for test-time adaptation.
We hope this work provides useful insights for future research on calibration in vision-language models and encourages the development of more reliable foundation models.

\section{Related Work}
\textbf{Prompt Tuning in Vision-Language Models.}
Vision-Language Models (VLMs) such as CLIP~\cite{radford2021learning} and ALIGN~\cite{jia2021scaling} learn a shared embedding space that aligns visual features with corresponding textual descriptions, enabling zero-shot transfer to a variety of downstream classification tasks.
To efficiently adapt VLMs to diverse downstream tasks, many parameter-efficient fine-tuning (PEFT) methods~\cite{lialin2023scaling,han2024parameter} have proven effective by optimizing only a small subset of parameters while keeping the rest of the model frozen.
One widely used PEFT technique is prompt tuning~\cite{lester2021power,jia2022visual}, which adds trainable tokens either to the input or to intermediate layers.
In particular, CoOp~\cite{zhou2022learning} pioneers prompt learning for VLMs by replacing fixed templates with learnable continuous vectors.
Subsequently, CoCoOp~\cite{zhou2022conditional} incorporates visual cues to generate instance-specific prompts, while MaPLE~\cite{khattak2023maple} further enhances context tuning by applying learnable prompts to both the image and text encoders.
To avoid relying on annotated downstream training data during adaptation, several methods~\cite{huang2022unsupervised,tanwisuth2023pouf,liang2024realistic} explore unsupervised prompt learning with pre-trained VLMs.
TPT~\cite{shu2022testtime} focuses on an interesting unsupervised scenario, where the model is adapted using only a single instance at test time.
Such a challenging paradigm has gained increasing attention~\cite{imam2025test,farina2024frustratingly,zanella2024test,sui2025just,sheng2025r}, and our work also studies this problem but focuses more on calibration performance.

\textbf{Calibration of Deep Neural Networks.}
In real-world safety-critical decision systems, classification networks must not only be accurate but also indicate when they are likely to be incorrect.
Confidence calibration~\cite{guo2017calibration,nixon2019measuring} formalizes this by measuring how well a model's predicted confidence aligns with its actual accuracy.
A recent survey \cite{wang2023calibration} categorizes existing calibration methods into post-hoc, regularization, and uncertainty estimation approaches.
Post-hoc calibration methods \cite{guo2017calibration,kull2019beyond,ding2021local} adjust predictions post-training without modifying parameters.
A popular example is temperature scaling (TS)~\cite{guo2017calibration}, which scales logits using a temperature parameter optimized on a validation set.%
Later methods~\cite{kull2019beyond,ding2021local} extend TS to improve robustness under limited validation data, multi-label tasks, and distribution shifts.
By contrast, regularization-based calibration methods \cite{guo2017calibration,liu2022devil} always incorporate additional regularization objectives during model training.
These include explicit penalties on overconfident predictions \cite{guo2017calibration,liu2022devil} and implicit techniques (\eg, label smoothing \cite{muller2019does}, mixup \cite{thulasidasan2019mixup}, focal loss \cite{lin2017focal}) that improve calibration and generalization via softened targets.
Moreover, uncertainty estimation methods alleviate miscalibration by introducing randomness via Bayesian neural networks~\cite{blundell2015weight}, ensembles~\cite{lakshminarayanan2017simple}, and Monte Carlo dropout~\cite{gal2016dropout}.
Most existing post-hoc calibration methods require a labeled validation set.
One recent approach~\cite{hu2024pseudo} uses mixup synthesis over unlabeled samples to generate pseudo-labeled data for temperature scaling, whereas our method operates using only a single unlabeled instance.
We further provide a hybrid calibration strategy that combines temperature scaling with an output-level ensemble.

\textbf{Calibration of Vision-Language Models.}
Although prompt tuning methods are primarily developed to improve accuracy, recent works \cite{wang2024open,wang2025understanding,murugesan2024robust,yoon2024ctpt} have also examined the calibration performance of VLMs, particularly CLIP~\cite{radford2021learning}.
Specifically, existing few-shot prompt tuning methods are found to lead to a calibration trade-off between base and new classes \cite{wang2024open}, \eg, CoOp~\cite{zhou2022learning} causes overconfidence on new classes, whereas KgCoOp~\cite{yao2023visual} leads to underconfidence on base classes.
Meanwhile, a classic unsupervised prompt tuning method, TPT~\cite{yoon2024ctpt}, improves accuracy but incurs a high calibration error after test-time adaptation.
Since our method belongs to the unsupervised category, we primarily review prior work within this branch of the literature.
C-TPT~\cite{yoon2024ctpt} finds that well-calibrated prompts yield text embeddings with broader class-wise dispersion and, on this basis, proposes a dispersion-based regularization term to encourage embeddings to move away from the class centroid.
To fully exploit the textual feature space, several recent methods (\ie, O-TPT~\cite{sharifdeen2025otpt}, A-TPT~\cite{ahamed2026atpt}, and SoC~\cite{fillioux2026soc}) encourage angular separation between textual features to promote greater dispersion.
Alternatively, D-TPT~\cite{han2025d} identifies and mitigates the influence of dominant feature dimensions to improve calibration performance.
Unlike these regularization-based methods, we follow TS~\cite{guo2017calibration} and propose a simple yet efficient post-hoc calibration method.
The most closely related work to ours is SaLS~\cite{murugesan2024robust}, which adjusts logits within the zero-shot range to maintain reliable confidence.
Both methods are built on the observation that zero-shot predictions are relatively well-calibrated.
However, unlike SaLS~\cite{murugesan2024robust}, our method learns a temperature parameter to bridge the confidence gap before and after adaptation.

\section{Pseudocode}
\label{code}
\definecolor{pycomment}{RGB}{83,132,135}
\definecolor{pykeyword}{RGB}{105,75,190}
\definecolor{pymethod}{RGB}{70,95,190}
\definecolor{pyshape}{RGB}{30,30,30}

\newcommand{\PyComment}[1]{\textcolor{pycomment}{\ttfamily\# #1}}
\newcommand{\PyKey}[1]{\textcolor{pykeyword}{\ttfamily #1}}
\newcommand{\PyMethod}[1]{\textcolor{pymethod}{\ttfamily #1}}
\newcommand{\PyShape}[1]{\textcolor{pyshape}{\ttfamily #1}}

\begin{algorithm}[H]
\caption{PyTorch-style code for \cots~and \ecots}
\label{alg:tpt_cots}
\normalsize
\setlength{\baselineskip}{0.65\baselineskip}
\DontPrintSemicolon
\SetAlgoNoLine
\PyComment{x: input image (C, H, W); $f_\text{clip}$: pretrained CLIP; $\mathcal{A}$: augmentation}\\
\PyComment{N: views nums; $\rho$: selection ratio; T: steps; ensemble: False/ True}\\

\PyKey{def} \texttt{E-CoTS}\ (x, $f_\text{clip}$, $\mathcal{A}$, N, $\rho$, T)\PyKey{:}\\
\Indp
    \PyComment{step 1: augment and select confident views}\\
    views = $\mathcal{A}$\ (x, num\_views=N) \tcp*{\PyShape{(N, C, H, W)}}
    $\text{logits}_\texttt{zs}$ = $f_\text{clip}$(views) \tcp*{\PyShape{(N, K)}}
    idx$_\texttt{zs}$ = \PyMethod{select\_confident\_views}($\text{logits}_\texttt{zs}$, top=$\rho$)\\

    \PyComment{step 2: test-time prompt tuning}\\
    $f_\texttt{tpt}$ = \PyMethod{run\_tpt\_optimization}($f_\text{clip}$, views[idx$_\texttt{zs}$], steps=T)\\
    $\text{logits}_\texttt{tpt}$ = $f_\texttt{tpt}$(views)\\

    \PyComment{step 3: confidence-based temperature scaling (\cots, Eq.\eqref{eq:cots})}\\
    idx$_\texttt{tpt}$ = \PyMethod{select\_confident\_views}($\text{logits}_\texttt{tpt}$[1:], top=$\rho$)\\
    $\tau$ = \PyMethod{optimize\_temperature}($\text{logits}_\texttt{tpt}$[idx$_\texttt{tpt}$], logits$_\texttt{zs}$[idx$_\texttt{tpt}$]) \\
    $\hat{p}^\texttt{tpt}$ = ($\text{logits}_\texttt{tpt}$ / $\tau$).\PyMethod{softmax}(dim=1)\\
    \PyKey{if} ensemble\PyKey{:}     \PyComment{step 4: weak-strong ensemble (\ecots, Eq.\eqref{eq:ecots})}\\
    \Indp
        $\alpha$ = \PyMethod{compute\_average\_similarity}()\\
        $\hat{p}_\texttt{ecots}$ = $\alpha \cdot \hat{p}^\texttt{tpt}$[0]
        + $(1 - \alpha) \cdot \hat{p}^\texttt{tpt}$[idx$_\texttt{tpt}$].\PyMethod{mean}(dim=0)\\
        \PyKey{return} $\hat{p}_\texttt{ecots}$.\PyMethod{argmax}(), $\hat{p}_\texttt{ecots}$.\PyMethod{max}()\\
    \Indm
    \PyKey{else}\PyKey{:}\\
    \Indp
        $\hat{p}_\texttt{cots}$ = $\hat{p}^\texttt{tpt}$[0]\\
        \PyKey{return} $\hat{p}_\texttt{cots}$.\PyMethod{argmax}(), $\hat{p}_\texttt{cots}$.\PyMethod{max}()\\
    \Indm
\end{algorithm}

We provided the pseudocode in Algorithm~\ref{alg:tpt_cots}.

\section{Additional Experimental Results}
\subsection{Computational Efficiency}

\begin{table}[b]
\centering
\caption{Computation time (s) of different methods on DTD and ImageNet-A using ViT-B/16. We report the average inference time per sample.}
\label{tab:time_table_dtd_imageneta}
\scriptsize
\setlength{\tabcolsep}{4pt}
\renewcommand{\arraystretch}{1.15}
\resizebox{0.75\linewidth}{!}{
\begin{tabular}{cccccccc}
\toprule
\multirow{2}{*}{Dataset}
& TPT
& C-TPT
& O-TPT
& A-TPT
& SoC
& SaLS
& \cots~ \\
& {\scriptsize \cite{shu2022testtime}}
& {\scriptsize \cite{yoon2024ctpt}}
& {\scriptsize \cite{sharifdeen2025otpt}}
& {\scriptsize \cite{ahamed2026atpt}}
& {\scriptsize \cite{fillioux2026soc}}
& {\scriptsize \cite{murugesan2024robust}} \\
\midrule
\multirow{1}{*}{DTD}
 & 0.192 & 0.193 & 0.206 & 0.207 & 0.206 & +0.000 & +0.053 \\
\midrule
\multirow{1}{*}{ImageNet-A}
 & 0.223 & 0.224 & 0.258 & 0.258 & 0.256 & +0.000 & +0.054 \\
\bottomrule
\end{tabular}%
}
\end{table}

We measure the computation time of different methods using a single 48GB GPU (NVIDIA RTX A6000).
As shown in Table~\ref{tab:time_table_dtd_imageneta}, integrating~\texttt{CoTS} delivers effective calibration improvement without substantially increasing overall inference cost.
These results show that~\texttt{CoTS} offers an efficient calibration strategy for test-time prompt tuning.

\subsection{Results with ResNet-50}
\label{sec:main_rn50}
\begin{table*}[t]
\centering
\caption{Performance comparison on fine-grained datasets using ResNet-50. Accuracy (\%) and ECE (\%) are reported.}
\label{tab:main_standard_rn50}
\scriptsize
\setlength{\tabcolsep}{3.2pt}
\resizebox{\textwidth}{!}{
\begin{tabular}{llcccccccccccc}
\toprule
(\%) & Method & ImgNet & DTD & Flowers & Food101 & SUN397 & Aircraft & Pets & Caltech & UCF101 & EuroSAT & Cars & Avg. \\
\midrule
\multirow{10}{*}{\rotatebox[origin=c]{90}{Accuracy}}
& Zero-Shot~\cite{radford2021learning}
& 58.17 & 40.37 & 61.67 & 73.94 & 58.84 & 15.75 & 83.54 & 85.88 & 58.84 & 23.67 & 55.78 & 56.04 \\
& TPT~\cite{shu2022testtime}
& 60.68 & 41.51 & 62.57 & 74.99 & 61.36 & 17.69 & 84.43 & 87.87 & 60.67 & 28.42 & 58.47 & 58.06 \\
& C-TPT~\cite{yoon2024ctpt}
& 60.47 & 41.25 & 65.03 & 74.84 & 60.93 & 17.21 & 83.66 & 87.37 & 60.25 & 27.18 & 56.38 & 57.69 \\
& Penalty~\cite{murugesan2024robust}
& 60.64 & 41.31 & 62.66 & 74.70 & 61.25 & 17.36 & 83.94 & 87.56 & 60.08 & 25.67 & 57.56 & 57.52 \\
& SaLS~\cite{murugesan2024robust}
& 60.69 & 41.51 & 62.57 & 74.99 & 61.36 & 17.68 & 84.43 & 87.86 & 60.67 & 28.42 & 58.47 & 58.06 \\
& O-TPT~\cite{sharifdeen2025otpt}
& 58.99 & 41.65 & 65.58 & 74.64 & 59.61 & 17.03 & 83.19 & 87.07 & 59.81 & 27.73 & 55.64 & 57.36 \\
& A-TPT~\cite{ahamed2026atpt}
& 59.91 & 41.86 & 64.20 & 74.29 & 59.31 & 16.00 & 82.68 & 86.88 & 59.81 & 29.14 & 56.17 & 57.30 \\
& SoC~\cite{fillioux2026soc}
& 59.93 & 40.37 & 61.90 & 73.82 & 59.04 & 15.67 & 83.28 & 85.84 & 58.90 & 23.64 & 55.68 & 56.19 \\
& \cellcolor{casbg}\texttt{\hspace*{0pt}CoTS}
& \cellcolor{casbg}60.68 & \cellcolor{casbg}41.51 & \cellcolor{casbg}62.57 & \cellcolor{casbg}74.99 & \cellcolor{casbg}61.36 & \cellcolor{casbg}17.69 & \cellcolor{casbg}84.43 & \cellcolor{casbg}87.87 & \cellcolor{casbg}60.67 & \cellcolor{casbg}28.42 & \cellcolor{casbg}58.47 & \cellcolor{casbg}58.06 \\
& \cellcolor{casbg}\texttt{\hspace*{0pt}E-CoTS}
& \cellcolor{casbg}60.90 & \cellcolor{casbg}41.33 & \cellcolor{casbg}60.71 & \cellcolor{casbg}74.27 & \cellcolor{casbg}61.33 & \cellcolor{casbg}17.73 & \cellcolor{casbg}84.12 & \cellcolor{casbg}87.69 & \cellcolor{casbg}60.47 & \cellcolor{casbg}27.45 & \cellcolor{casbg}58.96 & \cellcolor{casbg}57.72 \\
\midrule
\multirow{10}{*}{\rotatebox[origin=c]{90}{ECE}}

& Zero-Shot~\cite{radford2021learning}
& 2.01  & 9.04  & 3.03  & 2.71  & 3.81  & 6.30  & 5.70  & 4.41  & 3.00  & 14.76 & 4.40  & 5.38 \\
& TPT~\cite{shu2022testtime}
& 11.37 & 25.80 & 13.63 & 5.21  & 9.17  & 15.66 & 3.94  & 3.74  & 11.03 & 21.10 & 3.76  & 11.31 \\
& C-TPT~\cite{yoon2024ctpt}
& 6.75  & 21.65 & 3.81  & 1.76  & 3.01  & 10.69 & 2.79  & 2.61  & 3.02  & 13.37 & 1.33  & 6.44 \\
& Penalty~\cite{murugesan2024robust}
& 11.29 & 16.96 & 11.94 & 3.34  & 8.98  & 13.39 & 3.25  & 3.66  & 10.08 & 6.92  & 2.65  & 8.41 \\
& SaLS~\cite{murugesan2024robust}
& 9.91  & 21.66 & 11.99 & 3.93  & 8.66  & 15.29 & 3.07  & 4.02  & 8.98  & 12.32 & 2.54  & 9.31 \\
& O-TPT~\cite{sharifdeen2025otpt}
& 3.11  & 16.63 & 2.40  & 1.28  & 6.57  & 8.18  & 3.35  & 2.85  & 2.35  & 13.44 & 2.26  & 5.67 \\
& A-TPT~\cite{ahamed2026atpt}
& 2.32  & 15.35 & 2.71  & 1.55  & 4.22  & 8.26  & 2.72  & 4.16  & 2.63  & 4.71  & 1.43  & 4.55 \\
& SoC~\cite{fillioux2026soc}
& 3.02  & 8.95  & 3.34  & 3.63  & 3.89  & 6.36  & 5.61  & 4.53  & 2.37  & 14.79 & 4.02  & 5.50 \\
& \cellcolor{casbg}\texttt{\hspace*{0pt}CoTS}
& \cellcolor{casbg}3.74  & \cellcolor{casbg}7.97  & \cellcolor{casbg}2.62  & \cellcolor{casbg}3.74  & \cellcolor{casbg}5.24  & \cellcolor{casbg}4.88  & \cellcolor{casbg}5.29  & \cellcolor{casbg}5.84  & \cellcolor{casbg}2.79  & \cellcolor{casbg}7.33  & \cellcolor{casbg}6.05  & \cellcolor{casbg}5.04  \\
& \cellcolor{casbg}\texttt{\hspace*{0pt}E-CoTS}
& \cellcolor{casbg}2.36  & \cellcolor{casbg}9.47  & \cellcolor{casbg}6.80  & \cellcolor{casbg}2.18  & \cellcolor{casbg}3.36  & \cellcolor{casbg}5.40  & \cellcolor{casbg}3.89  & \cellcolor{casbg}4.40  & \cellcolor{casbg}3.67  & \cellcolor{casbg}8.54  & \cellcolor{casbg}5.42  & \cellcolor{casbg}5.04 \\
\bottomrule
\end{tabular}
}
\end{table*}

\begin{table}[t]
\centering
\caption{Performance comparison of different methods on ImageNet variants using ResNet-50. Accuracy (\%) and ECE (\%) are reported.}
\label{tab:imagenet_variants_rn50}
\scriptsize
\setlength{\tabcolsep}{5pt}
\renewcommand{\arraystretch}{1.12}
\resizebox{\linewidth}{!}{
\begin{tabular}{lcccccccccc}
\toprule
\multirow{2}{*}{\textbf{Method}}
& \multicolumn{5}{c}{Accuracy (\%)}
& \multicolumn{5}{c}{ECE (\%)} \\
\cmidrule(lr){2-6} \cmidrule(lr){7-11}
& -A & -V & -R & -Sk & Avg.
& -A & -V & -R & -Sk & Avg. \\
\midrule
Zero-Shot~\cite{radford2021learning}
& 21.87 & 51.45 & 56.12 & 33.36 & 40.70 & 21.24 & 3.03  & 0.98  & 3.11  & 7.09 \\
TPT~\cite{shu2022testtime}
& 26.51 & 54.63 & 59.04 & 35.13 & 43.83 & 30.96 & 13.94 & 10.47 & 14.84 & 17.55 \\
C-TPT~\cite{yoon2024ctpt}
& 24.27 & 54.21 & 57.77 & 34.75 & 42.75 & 26.61 & 9.35  & 5.51  & 11.08 & 13.14 \\
Penalty~\cite{murugesan2024robust}
& 26.12 & 54.58 & 58.80 & 35.12 & 43.66 & 29.85 & 13.80 & 9.73  & 14.70 & 17.02 \\
SaLS~\cite{murugesan2024robust}
& 26.52 & 54.63 & 59.04 & 35.13 & 43.83 & 28.10 & 12.27 & 7.79  & 13.32 & 15.37 \\
O-TPT~\cite{sharifdeen2025otpt}
& 21.59 & 52.37 & 56.11 & 33.20 & 40.82 & 19.23 & 2.38  & 1.52  & 2.70  & 6.46 \\
A-TPT~\cite{ahamed2026atpt}
& 22.56 & 53.26 & 57.61 & 34.32 & 41.94 & 20.28 & 4.11  & 0.94  & 5.70  & 7.76 \\
SoC~\cite{fillioux2026soc}
& 22.60 & 53.28 & 56.85 & 33.72 & 41.61 & 19.38 & 4.24  & 2.47  & 3.97  & 7.52 \\
\rowcolor{casbg}\texttt{\hspace*{0pt}CoTS}
& 26.51 & 54.63 & 59.04 & 35.13 & 43.83 & 17.52 & 4.55  & 3.07  & 2.33  & 6.87  \\
\rowcolor{casbg}\texttt{\hspace*{0pt}E-CoTS}
& 31.80 & 55.20 & 58.90 & 35.65 & 45.39 & 17.75 & 4.60  & 1.72  & 4.89  & 7.24 \\
\bottomrule
\end{tabular}
}
\end{table}

We further evaluate different methods with ResNet-50 to verify the robustness of our approach across architectures.
The results in fine-grained datasets and ImageNet variants are shown in Table~\ref{tab:main_standard_rn50} and Table~\ref{tab:imagenet_variants_rn50}.

\subsection{Results with Online TTA Framework}
\label{sec:online_tta}
\begin{table*}[t]
\centering
\caption{Performance comparison of different methods integrated with an online TTA framework (TDA~\cite{karmanov2024efficient}) on fine-grained datasets using ViT-B/16. Accuracy (\%) and ECE (\%) are reported.}
\label{tab:online_tta}
\scriptsize
\setlength{\tabcolsep}{3pt}
\resizebox{\textwidth}{!}{
\begin{tabular}{llcccccccccccr}
\toprule
(\%) & Method & ImgNet & DTD & Flowers & Food101 & SUN397 & Aircraft & Pets & Caltech & UCF101 & EuroSAT & Cars & Average \\
\midrule

\multirow{9}{*}{\rotatebox[origin=c]{90}{Accuracy}}

& TDA~\cite{karmanov2024efficient}
&68.27&46.04&69.75&83.68&65.01&24.06&88.36&94.04&67.99&51.04&66.55&65.65\\

& C-TPT~\cite{yoon2024ctpt}
&69.15&47.22&70.69&83.42&65.52&24.30&88.69&94.08&67.64&47.67&67.01&65.62\\

& Penalty~\cite{murugesan2024robust}
& 68.26 & 46.04 & 69.79 & 83.67 & 65.09 & 24.12
& 88.36 & 94.00 & 67.99 & 51.04 & 66.57 & 65.90 \\

& SaLS~\cite{murugesan2024robust}
&68.27&46.04&69.75&83.68&65.01&24.06&88.36&94.04&67.99&51.04&66.55&65.65\\

& O-TPT~\cite{sharifdeen2025otpt}
&68.49&48.11&70.77&83.15&64.93&24.09&88.58&94.20&66.98&47.06&66.56&65.44\\

& A-TPT~\cite{ahamed2026atpt}
&68.83&48.11&71.09&83.22&65.08&23.97&88.55&93.43&67.01&51.04&66.53&65.80\\

& SoC~\cite{fillioux2026soc}
&68.59&45.21&69.71&83.66&64.82&24.24&88.61&94.20&68.17&51.04&66.55&65.62\\

&\cellcolor{casbg}\texttt{CoTS}
&\cellcolor{casbg}68.27
&\cellcolor{casbg}46.04
&\cellcolor{casbg}69.75
&\cellcolor{casbg}83.68
&\cellcolor{casbg}65.01
&\cellcolor{casbg}24.06
&\cellcolor{casbg}88.36
&\cellcolor{casbg}94.04
&\cellcolor{casbg}67.99
&\cellcolor{casbg}51.04
&\cellcolor{casbg}66.55
&\cellcolor{casbg}65.65\\

&\cellcolor{casbg}\texttt{E-CoTS}
&\cellcolor{casbg}70.25
&\cellcolor{casbg}47.16
&\cellcolor{casbg}69.35
&\cellcolor{casbg}84.58
&\cellcolor{casbg}66.96
&\cellcolor{casbg}25.02
&\cellcolor{casbg}88.80
&\cellcolor{casbg}94.24
&\cellcolor{casbg}69.13
&\cellcolor{casbg}51.52
&\cellcolor{casbg}68.85
&\cellcolor{casbg}66.56\\

\midrule

\multirow{9}{*}{\rotatebox[origin=c]{90}{ECE}}

& TDA~\cite{karmanov2024efficient}
&5.27
&17.02&7.76&2.21&5.78&13.98&2.81&3.32&5.88&7.05&1.95&6.78\\

& C-TPT~\cite{yoon2024ctpt}
&8.80
&20.09&11.35&1.68&7.11&12.54&2.04&3.14&5.41&12.32&4.17&7.98\\

& Penalty~\cite{murugesan2024robust}
& 5.29 & 16.99 & 7.71 & 2.23 & 5.75 & 13.94
& 2.86 & 3.28 & 5.92 & 7.08 & 2.04 & 6.64 \\

& SaLS~\cite{murugesan2024robust}
&3.98
&13.36&5.59&1.48&4.33&12.10&3.38&3.72&4.78&6.75&1.90&5.74\\

& O-TPT~\cite{sharifdeen2025otpt}
&5.80
&15.10&9.71&1.92&3.85&11.70&1.79&3.20&4.60&13.42&3.63&6.89\\

& A-TPT~\cite{ahamed2026atpt}
&5.43
&14.88&10.21&1.47&7.58&14.46&1.26&3.16&4.63&7.05&3.97&6.87\\

& SoC~\cite{fillioux2026soc}
&5.35
&16.31&8.95&1.48&5.08&13.85&1.43&3.64&6.08&7.02&1.85&6.57\\

&\cellcolor{casbg}\texttt{CoTS}
&\cellcolor{casbg}3.47
&\cellcolor{casbg}7.13
&\cellcolor{casbg}3.09
&\cellcolor{casbg}2.54
&\cellcolor{casbg}4.97
&\cellcolor{casbg}4.72
&\cellcolor{casbg}5.02
&\cellcolor{casbg}6.40
&\cellcolor{casbg}3.37
&\cellcolor{casbg}6.66
&\cellcolor{casbg}5.61
&\cellcolor{casbg}4.95\\

&\cellcolor{casbg}\texttt{E-CoTS}
&\cellcolor{casbg}3.05
&\cellcolor{casbg}6.26
&\cellcolor{casbg}2.77
&\cellcolor{casbg}3.02
&\cellcolor{casbg}5.20
&\cellcolor{casbg}3.80
&\cellcolor{casbg}3.27
&\cellcolor{casbg}5.28
&\cellcolor{casbg}3.13
&\cellcolor{casbg}6.78
&\cellcolor{casbg}7.02
&\cellcolor{casbg}4.65\\

\bottomrule
\end{tabular}
}
\vspace{-10pt}
\end{table*}
\begin{table}[t]
\centering
\caption{Performance comparison of different methods integrated with an online TTA framework (TDA~\cite{karmanov2024efficient}) on ImageNet variants using ViT-B/16. Accuracy (\%) and ECE (\%) are reported.}
\label{tab:online_tta_imagenet_variants}
\scriptsize
\setlength{\tabcolsep}{5pt}
\renewcommand{\arraystretch}{1.12}
\resizebox{\linewidth}{!}{
\begin{tabular}{lccccrccccr}
\toprule

\multirow{2}{*}{Method} 
& \multicolumn{5}{c}{Accuracy (\%)} 
& \multicolumn{5}{c}{ECE (\%)} \\

\cmidrule(lr){2-6} 
\cmidrule(lr){7-11}

& -A & -V & -R & -Sk & Average 
& -A & -V & -R & -Sk & Average \\

\midrule

TDA~\cite{karmanov2024efficient}
&50.03&61.87&75.17&48.84&58.98
&11.96&10.06&1.62&10.10&8.44
\\

C-TPT~\cite{yoon2024ctpt}
&52.92&63.32&76.60&49.21&60.51
&12.03&13.06&1.93&14.59&10.40
\\

Penalty~\cite{murugesan2024robust}
& 50.03 & 61.89 & 75.17 & 48.83 & 58.98
& 11.98 & 10.07 & 1.62 & 10.14 & 8.45
\\

SaLS~\cite{murugesan2024robust}
&50.03&61.87&75.17&48.84&58.98
&10.22&8.11&2.23&7.95&7.13
\\

O-TPT~\cite{sharifdeen2025otpt}
&50.65&62.27&75.22&48.98&59.28
&10.57&10.48&1.94&10.92&8.48
\\

A-TPT~\cite{ahamed2026atpt}
&51.44&62.57&75.95&49.14&59.77
&9.45&10.01&1.84&10.43&7.93
\\

SoC~\cite{fillioux2026soc}
&51.45&62.29&75.96&48.73&59.61
&12.03&9.29&2.58&9.07&8.24
\\

\rowcolor{casbg}
\texttt{CoTS}
&\cellcolor{casbg}50.03
&\cellcolor{casbg}61.87
&\cellcolor{casbg}75.17
&\cellcolor{casbg}48.84
&\cellcolor{casbg}58.98
&
\cellcolor{casbg}6.37
&\cellcolor{casbg}3.40
&\cellcolor{casbg}5.43
&\cellcolor{casbg}2.22
&\cellcolor{casbg}4.36
\\

\rowcolor{casbg}
\texttt{E-CoTS}
&\cellcolor{casbg}59.63
&\cellcolor{casbg}64.28
&\cellcolor{casbg}78.36
&\cellcolor{casbg}51.16
&\cellcolor{casbg}63.36
&
\cellcolor{casbg}8.18
&\cellcolor{casbg}3.60
&\cellcolor{casbg}6.24
&\cellcolor{casbg}1.89
&\cellcolor{casbg}4.98
\\

\bottomrule
\end{tabular}
}
\end{table}

To assess the applicability of our approach beyond the standard episodic setting, we integrate ~\texttt{CoTS} and ~\texttt{E-CoTS} with TDA~\cite{karmanov2024efficient}, an online TTA framework.
As shown in Table~\ref{tab:online_tta} and Table~\ref{tab:online_tta_imagenet_variants}, ~\texttt{CoTS} substantially improves calibration, while ~\texttt{E-CoTS} achieves a favorable accuracy--calibration trade-off.

\subsection{Results with SigLIP}
\label{sec:siglip}
\begin{table*}[t]
\centering
\caption{Performance comparison of different methods with SigLIP~\cite{zhai2023sigmoid} on ImageNet and fine-grained datasets. Accuracy (\%) and ECE (\%) are reported.}
\label{tab:siglip_finegrained}
\scriptsize
\setlength{\tabcolsep}{3.2pt}
\resizebox{\textwidth}{!}{
\begin{tabular}{llcccccccccccr}
\toprule
(\%) & Method & ImgNet & DTD & Flowers & Food101 & SUN397 & Aircraft & Pets & Caltech & UCF101 & EuroSAT & Cars & Average \\
\midrule

\multirow{10}{*}{\rotatebox[origin=c]{90}{Accuracy}}

& Zero-Shot
& 75.69 & 62.94 & 84.41 & 87.30 & 69.64 & 40.65 & 93.21 & 97.93 & 70.82 & 41.35 & 90.70 & 74.06 \\

& TPT
& 76.46 & 64.60 & 84.53 & 87.60 & 70.00 & 40.95 & 92.75 & 98.13 & 71.19 & 40.68 & 91.28 & 74.38 \\

& C-TPT
& 76.11 & 63.18 & 84.25 & 87.35 & 69.70 & 40.29 & 92.89 & 97.97 & 70.61 & 40.69 & 90.67 & 73.97 \\

& Penalty
& 76.45 & 64.07 & 84.33 & 87.45 & 69.91 & 40.38 & 92.72 & 96.19 & 70.90 & 41.83 & 91.11 & 74.12 \\

& SaLS
& 76.46 & 64.60 & 84.53 & 87.60 & 70.00 & 40.95 & 92.75 & 98.13 & 71.19 & 40.68 & 91.28 & 74.38 \\

& O-TPT
& 75.69 & 62.41 & 84.33 & 87.32 & 69.56 & 39.78 & 92.86 & 97.93 & 70.18 & 40.68 & 90.69 & 73.77 \\

& A-TPT
& 75.91 & 62.41 & 84.29 & 87.31 & 69.59 & 40.32 & 92.89 & 98.01 & 70.55 & 40.60 & 90.69 & 73.87 \\

& SoC
& 75.97 & 63.00 & 84.33 & 87.35 & 69.68 & 40.65 & 93.13 & 97.93 & 70.82 & 41.35 & 90.75 & 74.09 \\

& \cellcolor{casbg}\texttt{CoTS}
& \cellcolor{casbg}76.46
& \cellcolor{casbg}64.60
& \cellcolor{casbg}84.53
& \cellcolor{casbg}87.60
& \cellcolor{casbg}70.00
& \cellcolor{casbg}40.95
& \cellcolor{casbg}92.75
& \cellcolor{casbg}98.13
& \cellcolor{casbg}71.19
& \cellcolor{casbg}40.68
& \cellcolor{casbg}91.28
& \cellcolor{casbg}74.38 \\

& \cellcolor{casbg}\texttt{E-CoTS}
& \cellcolor{casbg}77.86
& \cellcolor{casbg}65.19
& \cellcolor{casbg}84.78
& \cellcolor{casbg}87.73
& \cellcolor{casbg}70.04
& \cellcolor{casbg}39.60
& \cellcolor{casbg}92.72
& \cellcolor{casbg}98.22
& \cellcolor{casbg}71.64
& \cellcolor{casbg}38.74
& \cellcolor{casbg}91.38
& \cellcolor{casbg}74.35 \\

\midrule

\multirow{10}{*}{\rotatebox[origin=c]{90}{ECE}}

& Zero-Shot
& 4.54 & 6.29 & 3.83 & 2.04 & 3.89 & 7.80 & 2.54 & 1.53 & 6.94 & 15.99 & 0.69 & 5.10 \\

& TPT
& 6.57 & 11.70 & 4.08 & 3.34 & 7.04 & 12.90 & 1.87 & 0.81 & 9.52 & 19.90 & 1.87 & 7.24 \\

& C-TPT
& 5.81 & 9.40 & 4.15 & 2.57 & 5.35 & 11.48 & 2.12 & 1.30 & 8.35 & 18.39 & 0.91 & 6.35 \\

& Penalty
& 6.49 & 9.68 & 4.21 & 2.69 & 6.90 & 12.90 & 2.60 & 1.63 & 8.30 & 14.35 & 1.43 & 6.47 \\

& SaLS
& 6.29 & 10.43 & 4.12 & 3.05 & 6.63 & 12.61 & 1.87 & 0.82 & 9.07 & 18.95 & 1.75 & 6.87 \\

& O-TPT
& 4.96 & 8.76 & 3.87 & 2.42 & 4.63 & 11.02 & 2.18 & 1.38 & 8.11 & 18.51 & 0.74 & 6.05 \\

& A-TPT
& 5.35 & 8.36 & 3.99 & 2.48 & 4.87 & 11.28 & 2.16 & 1.29 & 8.23 & 17.84 & 0.83 & 6.06 \\

& SoC
& 5.30 & 6.30 & 3.89 & 2.30 & 4.36 & 7.83 & 2.33 & 1.21 & 7.40 & 16.04 & 1.03 & 5.27 \\

& \cellcolor{casbg}\texttt{CoTS}
& \cellcolor{casbg}4.18
& \cellcolor{casbg}4.67
& \cellcolor{casbg}4.18
& \cellcolor{casbg}2.20
& \cellcolor{casbg}3.68
& \cellcolor{casbg}7.55
& \cellcolor{casbg}2.27
& \cellcolor{casbg}1.37
& \cellcolor{casbg}6.65
& \cellcolor{casbg}15.12
& \cellcolor{casbg}1.18
& \cellcolor{casbg}4.82 \\

& \cellcolor{casbg}\texttt{E-CoTS}
& \cellcolor{casbg}4.22
& \cellcolor{casbg}5.22
& \cellcolor{casbg}4.67
& \cellcolor{casbg}1.95
& \cellcolor{casbg}3.95
& \cellcolor{casbg}6.64
& \cellcolor{casbg}2.34
& \cellcolor{casbg}1.44
& \cellcolor{casbg}6.26
& \cellcolor{casbg}16.27
& \cellcolor{casbg}0.84
& \cellcolor{casbg}4.89 \\

\bottomrule
\end{tabular}
}
\end{table*}
\begin{table}[t]
\centering
\caption{Performance comparison of different methods with SigLIP~\cite{zhai2023sigmoid} on ImageNet variants. Accuracy (\%) and ECE (\%) are reported.}
\label{tab:siglip_imagenet_variants}
\scriptsize
\setlength{\tabcolsep}{5pt}
\renewcommand{\arraystretch}{1.12}
\resizebox{\linewidth}{!}{
\begin{tabular}{lccccrccccr}
\toprule

\multirow{2}{*}{Method}
& \multicolumn{5}{c}{Accuracy (\%)}
& \multicolumn{5}{c}{ECE (\%)} \\

\cmidrule(lr){2-6}
\cmidrule(lr){7-11}

& -A & -V & -R & -Sk & Average
& -A & -V & -R & -Sk & Average \\

\midrule

Zero-Shot
&45.07&68.43&89.31&66.59&67.35
&15.90&7.26&1.42&7.43&8.00\\

TPT
&46.73&69.23&89.89&67.08&68.23
&17.25&9.45&0.68&10.37&9.43\\

C-TPT
&45.84&68.94&89.57&66.88&67.81
&17.08&8.65&0.88&9.57&9.04\\

Penalty
&46.01&69.19&89.70&67.08&68.00
&16.98&9.40&0.78&10.28&9.36\\

SaLS
&46.73&69.23&89.89&67.08&68.23
&16.69&9.11&0.86&9.98&9.16\\

O-TPT
&45.25&68.41&89.34&66.56&67.39
&16.54&7.79&1.05&7.99&8.34\\

A-TPT
&45.45&68.78&89.46&66.80&67.62
&16.71&8.10&0.96&8.69&8.62\\

SoC
&45.72&68.75&89.51&66.85&67.71
&16.39&8.05&1.02&8.39&8.46\\

\rowcolor{casbg}
\texttt{CoTS}
&\cellcolor{casbg}46.73
&\cellcolor{casbg}69.23
&\cellcolor{casbg}89.89
&\cellcolor{casbg}67.08
&\cellcolor{casbg}68.23
&\cellcolor{casbg}13.91
&\cellcolor{casbg}6.72
&\cellcolor{casbg}1.82
&\cellcolor{casbg}7.14
&\cellcolor{casbg}7.40\\

\rowcolor{casbg}
\texttt{E-CoTS}
&\cellcolor{casbg}56.35
&\cellcolor{casbg}70.70
&\cellcolor{casbg}90.95
&\cellcolor{casbg}67.44
&\cellcolor{casbg}71.36
&\cellcolor{casbg}10.64
&\cellcolor{casbg}6.42
&\cellcolor{casbg}2.44
&\cellcolor{casbg}7.43
&\cellcolor{casbg}6.73\\

\bottomrule
\end{tabular}
}
\end{table}

We also investigate whether the proposed methods remain effective with a different vision-language model by conducting experiments with SigLIP~\cite{zhai2023sigmoid}.
Table~\ref{tab:siglip_finegrained} and Table~\ref{tab:siglip_imagenet_variants} show that ~\texttt{CoTS} and ~\texttt{E-CoTS} consistently provide competitive accuracy and improved calibration.

\subsection{Standard Deviations}
We report standard deviations of accuracy and ECE across three seeds in Table~\ref{tab:std_vit_fg_imagenet} and Table~\ref{tab:std_imagenet_variants}.
On fine-grained datasets and ImageNet, \cots~achieves consistently low variance, indicating stable performance.
On ImageNet variants, both \cots~and \ecots~remain robust across seeds, with \ecots~maintaining strong calibration despite minor variability.
Overall, the results confirm that the proposed methods are reliable and stable while improving calibration under different runs.
\begin{table*}[t]
\centering
\caption{Standard deviations across three seeds on fine-grained datasets and ImageNet using ViT-B/16. Accuracy(\%) and ECE(\%) are reported in percentage points.}
\label{tab:std_vit_fg_imagenet}
\scriptsize
\setlength{\tabcolsep}{2pt}
\renewcommand{\arraystretch}{1.12}
\resizebox{\textwidth}{!}{
\begin{tabular}{llccccccccccc}
\toprule
(\%) & Method & ImgNet & DTD & Flowers & Food101 & SUN397 & Aircraft & Pets & Caltech & UCF101 & EuroSAT & Cars \\
\midrule
\multirow{10}{*}{\rotatebox[origin=c]{90}{Accuracy}}
& TPT~\cite{shu2022testtime} & 0.04 & 0.11 & 0.08 & 0.02 & 0.06 & 0.16 & 0.11 & 0.22 & 0.27 & 0.09 & 0.08 \\
& C-TPT~\cite{yoon2024ctpt}  & 0.01 & 0.15 & 0.12 & 0.03 & 0.11 & 0.09 & 0.02 & 0.10 & 0.05 & 0.02 & 0.11 \\
& Penalty~\cite{murugesan2024robust} & 0.02 & 0.10 & 0.23 & 0.01 & 0.06 & 0.35 & 0.19 & 0.18 & 0.15 & 0.06 & 0.07 \\
& SALS~\cite{murugesan2024robust} & 0.04 & 0.11 & 0.08 & 0.02 & 0.05 & 0.19 & 0.11 & 0.22 & 0.27 & 0.09 & 0.09 \\
& O-TPT~\cite{sharifdeen2025otpt} & 0.02 & 0.11 & 0.04 & 0.04 & 0.04 & 0.01 & 0.05 & 0.10 & 0.06 & 0.06 & 0.12 \\
& A-TPT~\cite{ahamed2026atpt} & 0.02 & 0.03 & 0.07 & 0.03 & 0.04 & 0.35 & 0.01 & 0.16 & 0.13 & 0.00 & 0.10 \\
& SoC~\cite{fillioux2026soc} & 0.03 & 0.05 & 0.12 & 0.03 & 0.04 & 0.01 & 0.12 & 0.11 & 0.07 & 0.01 & 0.05 \\
& \cellcolor{casbg}\texttt{\hspace*{0pt}CoTS} & \cellcolor{casbg}0.04 & \cellcolor{casbg}0.11 & \cellcolor{casbg}0.08 & \cellcolor{casbg}0.02 & \cellcolor{casbg}0.06 & \cellcolor{casbg}0.16 & \cellcolor{casbg}0.11 & \cellcolor{casbg}0.22 & \cellcolor{casbg}0.27 & \cellcolor{casbg}0.09 & \cellcolor{casbg}0.08 \\
& \cellcolor{casbg}\texttt{\hspace*{0pt}E-CoTS} & \cellcolor{casbg}0.04 & \cellcolor{casbg}0.28 & \cellcolor{casbg}0.06 & \cellcolor{casbg}0.03 & \cellcolor{casbg}0.09 & \cellcolor{casbg}0.20 & \cellcolor{casbg}0.26 & \cellcolor{casbg}0.12 & \cellcolor{casbg}0.18 & \cellcolor{casbg}0.14 & \cellcolor{casbg}0.13 \\
\midrule
\multirow{10}{*}{\rotatebox[origin=c]{90}{ECE}}
& TPT~\cite{shu2022testtime} & 0.07 & 0.10 & 0.07 & 0.04 & 0.14 & 0.12 & 0.10 & 0.12 & 0.38 & 0.04 & 0.07 \\
& C-TPT~\cite{yoon2024ctpt} & 0.04 & 0.17 & 0.10 & 0.04 & 0.15 & 0.07 & 0.10 & 0.12 & 0.22 & 0.26 & 0.35 \\
& Penalty~\cite{murugesan2024robust} & 0.04 & 0.15 & 0.25 & 0.07 & 0.11 & 0.40 & 0.18 & 0.14 & 0.17 & 0.15 & 0.17 \\
& SALS~\cite{murugesan2024robust} & 0.06 & 0.18 & 0.11 & 0.01 & 0.03 & 0.11 & 0.09 & 0.06 & 0.36 & 0.38 & 0.08 \\
& O-TPT~\cite{sharifdeen2025otpt} & 0.02 & 0.14 & 0.14 & 0.04 & 0.04 & 0.13 & 0.07 & 0.10 & 0.07 & 0.10 & 0.15 \\
& A-TPT~\cite{ahamed2026atpt} & 0.02 & 0.25 & 0.13 & 0.03 & 0.08 & 0.31 & 0.10 & 0.16 & 0.08 & 0.00 & 0.10 \\
& SoC~\cite{fillioux2026soc} & 0.06 & 0.02 & 0.04 & 0.02 & 0.02 & 0.05 & 0.16 & 0.09 & 0.06 & 0.01 & 0.05 \\
& \cellcolor{casbg}\texttt{\hspace*{0pt}CoTS} & \cellcolor{casbg}0.08 & \cellcolor{casbg}0.61 & \cellcolor{casbg}0.12 & \cellcolor{casbg}0.03 & \cellcolor{casbg}0.19 & \cellcolor{casbg}0.28 & \cellcolor{casbg}0.07 & \cellcolor{casbg}0.22 & \cellcolor{casbg}0.29 & \cellcolor{casbg}0.27 & \cellcolor{casbg}0.07 \\
& \cellcolor{casbg}\texttt{\hspace*{0pt}E-CoTS} & \cellcolor{casbg}0.03 & \cellcolor{casbg}0.86 & \cellcolor{casbg}0.26 & \cellcolor{casbg}0.00 & \cellcolor{casbg}0.26 & \cellcolor{casbg}0.13 & \cellcolor{casbg}0.09 & \cellcolor{casbg}0.16 & \cellcolor{casbg}0.36 & \cellcolor{casbg}0.13 & \cellcolor{casbg}0.15 \\
\bottomrule
\end{tabular}
}
\end{table*}

\begin{table}[t]
\centering
\caption{Standard deviations across three seeds on ImageNet variants using ViT-B/16. Accuracy(\%) and ECE(\%) are reported in percentage points.}
\label{tab:std_imagenet_variants}
\scriptsize
\setlength{\tabcolsep}{10pt}
\renewcommand{\arraystretch}{1.12}
\resizebox{\linewidth}{!}{
\begin{tabular}{lcccccccc}
\toprule
\multirow{2}{*}{\textbf{Method}}
& \multicolumn{4}{c}{Accuracy (\%)}
& \multicolumn{4}{c}{ECE (\%)} \\
\cmidrule(lr){2-5} \cmidrule(lr){6-9}
& -A & -V & -R & -Sk
& -A & -V & -R & -Sk \\
\midrule
TPT~\cite{shu2022testtime} & 0.07 & 0.14 & 0.04 & 0.03 & 0.15 & 0.24 & 0.03 & 0.05 \\
C-TPT~\cite{yoon2024ctpt} & 0.13 & 0.16 & 0.06 & 0.02 & 0.18 & 0.05 & 0.05 & 0.02 \\
Penalty~\cite{murugesan2024robust} & 0.13 & 0.16 & 0.03 & 0.02 & 0.10 & 0.22 & 0.09 & 0.04 \\
SALS~\cite{murugesan2024robust} & 0.07 & 0.14 & 0.04 & 0.03 & 0.04 & 0.24 & 0.13 & 0.03 \\
O-TPT~\cite{sharifdeen2025otpt} & 0.01 & 0.02 & 0.03 & 0.03 & 0.03 & 0.16 & 0.03 & 0.05 \\
A-TPT~\cite{ahamed2026atpt} & 0.04 & 0.19 & 0.06 & 0.04  & 0.14 & 0.18 & 0.07 & 0.06 \\
SoC~\cite{fillioux2026soc} & 0.10 & 0.08 & 0.03 & 0.03 & 0.09 & 0.11 & 0.04 & 0.02 \\
\rowcolor{casbg} \texttt{\hspace*{0pt}CoTS} & 0.07 & 0.14 & 0.04 & 0.03 & 0.29 & 0.18 & 0.05 & 0.08 \\
\rowcolor{casbg} \texttt{\hspace*{0pt}E-CoTS} & 0.24 & 0.11 & 0.08 & 0.07 & 0.02 & 0.16 &0.08 & 0.08 \\
\bottomrule
\end{tabular}
}
\end{table}

\subsection{Calibration Metrics}
\begin{table*}[t]
\centering
\caption{Brier scores (\%) of different methods on fine-grained datasets with ViT-B/16.}
\label{tab:main_brier}
\scriptsize
\setlength{\tabcolsep}{2pt}
\resizebox{\textwidth}{!}{
\begin{tabular}{llcccccccccccr}
\toprule
(\%) & Method & ImgNet & DTD & Flowers & Food101 & SUN397 & Aircraft & Pets & Caltech & UCF101 & EuroSAT & Cars & Average \\
\midrule
\multirow{10}{*}{\rotatebox[origin=c]{90}{Brier score}}
& Zero-Shot~\cite{radford2021learning}
& 16.44 & 18.45 & 14.22 & 9.92  & 18.33 & 14.77 & 7.47 & 6.11 & 14.55 & 19.36 & 16.44 & 14.19 \\
& TPT~\cite{shu2022testtime}
& 18.20 & 24.82 & 16.83 & 10.14 & 20.15 & 18.49 & 8.78 & 5.76 & 16.01 & 24.81 & 16.54 & 16.41 \\
& C-TPT~\cite{yoon2024ctpt}
& 16.88 & 21.12 & 13.95 & 10.29 & 19.36 & 14.40 & 7.20 & 5.86 & 15.17 & 22.36 & 16.61 & 14.84 \\
& Penalty~\cite{murugesan2024robust}
& 18.23 & 23.46	& 16.28	& 10.43	& 20.03	& 17.37	& 10.21	& 6.57 & 16.13 & 19.26 & 16.47 & 15.86 \\
& SaLS~\cite{murugesan2024robust}
& 17.94 & 23.36 & 16.23 & 9.98  & 19.90 & 17.61 & 8.60 & 5.83 & 15.75 & 22.66 & 16.34 & 15.84 \\
& O-TPT~\cite{sharifdeen2025otpt}
& 16.16 & 19.06 & 13.38 & 10.56 & 19.46 & 14.11 & 7.26 & 5.71 & 15.37 & 22.73 & 16.57 & 14.58 \\
& A-TPT~\cite{ahamed2026atpt}
& 16.47 & 19.71 & 13.45 & 10.27 & 19.00 & 15.04 & 7.44 & 6.34 & 15.65 & 19.36 & 16.53 & 14.48 \\
& SoC~\cite{fillioux2026soc}
& 16.63 & 18.61 & 13.84 & 10.10 & 18.79 & 14.79 & 7.00 & 6.02 & 14.45 & 19.37 & 16.59 & 14.20 \\
& \cellcolor{casbg} \texttt{\hspace*{-2pt}CoTS}
& \cellcolor{casbg}16.61 & \cellcolor{casbg}19.51 & \cellcolor{casbg}14.24 & \cellcolor{casbg}9.73  & \cellcolor{casbg}18.34 & \cellcolor{casbg}14.14 & \cellcolor{casbg}7.43 & \cellcolor{casbg}6.62 & \cellcolor{casbg}14.36 & \cellcolor{casbg}21.62 & \cellcolor{casbg}16.22 & \cellcolor{casbg}14.44 \\
& \cellcolor{casbg} \texttt{\hspace*{-2pt}E-CoTS}
& \cellcolor{casbg}16.30 & \cellcolor{casbg}19.72 & \cellcolor{casbg}14.36 & \cellcolor{casbg}9.60  & \cellcolor{casbg}17.99 & \cellcolor{casbg}14.30 & \cellcolor{casbg}7.45 & \cellcolor{casbg}5.58 & \cellcolor{casbg}14.03 & \cellcolor{casbg}21.43 & \cellcolor{casbg}15.96 & \cellcolor{casbg}14.25 \\
\bottomrule
\end{tabular}
}
\end{table*}

\begin{table*}[t]
\centering
\caption{Brier scores (\%) of different methods on ImageNet variants with ViT-B/16.}
\label{tab:main_brier2}
\scriptsize
\setlength{\tabcolsep}{8pt}
\resizebox{0.7\textwidth}{!}{
\begin{tabular}{llccccr}
\toprule
(\%) & Method & -A & -V & -R & -Sk & Average \\
\midrule
\multirow{10}{*}{\rotatebox[origin=c]{90}{Brier score}} & Zero-Shot~\cite{radford2021learning} & 20.68 & 17.58 & 12.45 & 17.91 & 17.16 \\
& TPT~\cite{shu2022testtime} & 23.95 & 20.10 & 12.82 & 22.12 & 19.75 \\
& C-TPT~\cite{yoon2024ctpt} & 21.64 & 18.56 & 12.47 & 19.83 & 18.13 \\
& Penalty~\cite{murugesan2024robust} & 23.87 & 20.00 & 12.97 & 22.13 & 19.74\\
& SaLS~\cite{murugesan2024robust} & 23.34 & 19.71 & 12.50 & 21.68 & 19.30 \\
& O-TPT~\cite{sharifdeen2025otpt} & 20.32 & 17.41 & 12.40 & 17.98 & 17.03 \\
& A-TPT~\cite{ahamed2026atpt} & 20.51 & 17.69 & 12.35 & 18.25 & 17.20 \\
& SoC~\cite{fillioux2026soc} & 20.63 & 17.66 & 12.25 & 18.09 & 17.16 \\
& \cellcolor{casbg} \texttt{\hspace*{-2pt}CoTS} & \cellcolor{casbg}21.40  & \cellcolor{casbg}17.97  & \cellcolor{casbg}12.67  & \cellcolor{casbg}18.33  & \cellcolor{casbg}17.59 \\
& \cellcolor{casbg} \texttt{\hspace*{-2pt}E-CoTS} & \cellcolor{casbg}21.93  & \cellcolor{casbg}17.75  & \cellcolor{casbg}12.10  & \cellcolor{casbg}18.39  & \cellcolor{casbg}17.54 \\
\bottomrule
\end{tabular}
}
\end{table*}

\begin{table*}[t]
\centering
\caption{Class-wise ECE (\%)~\cite{kull2019beyond} of different methods on fine-grained datasets with ViT-B/16.}
\label{tab:cece_fg}
\scriptsize
\setlength{\tabcolsep}{2pt}
\resizebox{\textwidth}{!}{
\begin{tabular}{llcccccccccccr}
\toprule
(\%) & Method & ImgNet & DTD & Flowers & Food101 & SUN397 & Aircraft & Pets & Caltech & UCF101 & EuroSAT & Cars & Average \\
\midrule

\multirow{10}{*}{\rotatebox[origin=c]{90}{CECE}}

& Zero-Shot~\cite{radford2021learning}
&0.04&1.42&0.62&0.21&0.13&0.57&0.70&0.24&0.52&6.59&0.23&1.03\\

& TPT~\cite{shu2022testtime}
&0.04&1.49&0.53&0.19&0.12&0.71&0.62&0.17&0.47&6.95&0.22&1.05\\

& C-TPT~\cite{yoon2024ctpt}
&0.04&1.39&0.54&0.24&0.12&0.61&0.60&0.22&0.55&7.37&0.23&1.08\\

& Penalty~\cite{murugesan2024robust}
&0.04&1.43&0.54&0.19&0.12&0.66&0.93&0.23&0.49&5.96&0.22&0.98\\

& SaLS~\cite{murugesan2024robust}
&0.04&1.46&0.53&0.18&0.12&0.69&0.62&0.18&0.47&6.38&0.22&0.99\\

& O-TPT~\cite{sharifdeen2025otpt}
&0.04&1.36&0.55&0.25&0.14&0.60&0.60&0.23&0.58&7.39&0.23&1.09\\

& A-TPT~\cite{ahamed2026atpt}
&0.04&1.38&0.54&0.24&0.13&0.60&0.63&0.25&0.57&6.59&0.23&1.02\\

& SoC~\cite{fillioux2026soc}
&0.04&1.44&0.58&0.23&0.13&0.57&0.62&0.25&0.51&6.58&0.23&1.02\\

&\cellcolor{casbg}\texttt{CoTS}
&\cellcolor{casbg}0.04
&\cellcolor{casbg}1.25
&\cellcolor{casbg}0.58
&\cellcolor{casbg}0.20
&\cellcolor{casbg}0.12
&\cellcolor{casbg}0.55
&\cellcolor{casbg}0.72
&\cellcolor{casbg}0.25
&\cellcolor{casbg}0.50
&\cellcolor{casbg}6.07
&\cellcolor{casbg}0.24
&\cellcolor{casbg}0.96\\

&\cellcolor{casbg}\texttt{E-CoTS}
&\cellcolor{casbg}0.04
&\cellcolor{casbg}1.28
&\cellcolor{casbg}0.58
&\cellcolor{casbg}0.20
&\cellcolor{casbg}0.12
&\cellcolor{casbg}0.56
&\cellcolor{casbg}0.70
&\cellcolor{casbg}0.22
&\cellcolor{casbg}0.50
&\cellcolor{casbg}6.02
&\cellcolor{casbg}0.24
&\cellcolor{casbg}\textbf{0.95}\\

\bottomrule
\end{tabular}
}
\end{table*}

\begin{table*}[t]
\centering
\caption{Class-wise ECE (\%)~\cite{kull2019beyond} of different methods on ImageNet variants with ViT-B/16.}
\label{tab:cece_imagenet}
\scriptsize
\setlength{\tabcolsep}{8pt}
\resizebox{0.7\textwidth}{!}{
\begin{tabular}{llccccr}
\toprule
(\%) & Method & -A & -V & -R & -Sk & Average\\
\midrule

\multirow{10}{*}{\rotatebox[origin=c]{90}{CECE}}

& Zero-Shot~\cite{radford2021learning}
&0.30&0.06&0.18&0.06&0.15\\

& TPT~\cite{shu2022testtime}
&0.28&0.06&0.14&0.06&0.14\\

& C-TPT~\cite{yoon2024ctpt}
&0.29&0.06&0.16&0.06&0.14\\

& Penalty~\cite{murugesan2024robust}
&0.28&0.06&0.14&0.06&0.14\\

& SaLS~\cite{murugesan2024robust}
&0.28&0.06&0.14&0.06&0.14\\

& O-TPT~\cite{sharifdeen2025otpt}
&0.30&0.06&0.18&0.06&0.15\\

& A-TPT~\cite{ahamed2026atpt}
&0.30&0.06&0.17&0.06&0.15\\

& SoC~\cite{fillioux2026soc}
&0.29&0.06&0.17&0.06&0.15\\

&\cellcolor{casbg}\texttt{CoTS}
&\cellcolor{casbg}0.27
&\cellcolor{casbg}0.06
&\cellcolor{casbg}0.17
&\cellcolor{casbg}0.06
&\cellcolor{casbg}0.14\\

&\cellcolor{casbg}\texttt{E-CoTS}
&\cellcolor{casbg}0.26
&\cellcolor{casbg}0.06
&\cellcolor{casbg}0.16
&\cellcolor{casbg}0.06
&\cellcolor{casbg}\textbf{0.13}\\

\bottomrule
\end{tabular}
}
\end{table*}
\begin{table*}[t]
\centering
\caption{Adaptive ECE (\%)~\cite{nixon2019measuring} of different methods on fine-grained datasets with ViT-B/16.}
\label{tab:aece_fg}
\scriptsize
\setlength{\tabcolsep}{2pt}
\resizebox{\textwidth}{!}{
\begin{tabular}{llcccccccccccr}
\toprule
(\%) & Method & ImgNet & DTD & Flowers & Food101 & SUN397 & Aircraft & Pets & Caltech & UCF101 & EuroSAT & Cars & Average \\
\midrule

\multirow{10}{*}{\rotatebox[origin=c]{90}{AECE}}

& Zero-Shot~\cite{radford2021learning}
&0.03&1.33&0.48&0.18&0.10&0.55&0.54&0.17&0.41&6.57&0.12&0.95\\

& TPT~\cite{shu2022testtime}
&0.02&1.37&0.45&0.15&0.08&0.66&0.54&0.13&0.38&6.93&0.13&0.98\\

& C-TPT~\cite{yoon2024ctpt}
&0.02&1.24&0.43&0.19&0.09&0.59&0.48&0.15&0.43&7.33&0.13&1.01\\

& Penalty~\cite{murugesan2024robust}
&0.02&1.35&0.46&0.16&0.08&0.63&0.80&0.16&0.40&5.97&0.13&0.92\\

& SaLS~\cite{murugesan2024robust}
&0.02&1.35&0.46&0.15&0.08&0.64&0.55&0.14&0.38&6.40&0.13&0.94\\

& O-TPT~\cite{sharifdeen2025otpt}
&0.03&1.28&0.43&0.20&0.09&0.60&0.48&0.15&0.46&7.34&0.14&1.02\\

& A-TPT~\cite{ahamed2026atpt}
&0.03&1.28&0.43&0.20&0.10&0.57&0.49&0.17&0.46&6.57&0.13&0.95\\

& SoC~\cite{fillioux2026soc}
&0.03&1.36&0.46&0.18&0.09&0.55&0.49&0.18&0.41&6.56&0.12&0.95\\

&\cellcolor{casbg}\texttt{CoTS}
&\cellcolor{casbg}0.03
&\cellcolor{casbg}1.29
&\cellcolor{casbg}0.48
&\cellcolor{casbg}0.17
&\cellcolor{casbg}0.09
&\cellcolor{casbg}0.56
&\cellcolor{casbg}0.60
&\cellcolor{casbg}0.18
&\cellcolor{casbg}0.41
&\cellcolor{casbg}6.11
&\cellcolor{casbg}0.12
&\cellcolor{casbg}\textbf{0.91}\\

&\cellcolor{casbg}\texttt{E-CoTS}
&\cellcolor{casbg}0.03
&\cellcolor{casbg}1.30
&\cellcolor{casbg}0.49
&\cellcolor{casbg}0.17
&\cellcolor{casbg}0.09
&\cellcolor{casbg}0.56
&\cellcolor{casbg}0.60
&\cellcolor{casbg}0.17
&\cellcolor{casbg}0.41
&\cellcolor{casbg}6.05
&\cellcolor{casbg}0.13
&\cellcolor{casbg}0.91\\

\bottomrule
\end{tabular}
}
\end{table*}

\begin{table*}[t]
\centering
\caption{Adaptive ECE (\%)~\cite{nixon2019measuring} of different methods on ImageNet variants with ViT-B/16.}
\label{tab:aece_imagenet}
\scriptsize
\setlength{\tabcolsep}{8pt}
\resizebox{0.7\textwidth}{!}{
\begin{tabular}{llccccr}
\toprule
(\%) & Method & -A & -V & -R & -Sk & Average\\
\midrule

\multirow{10}{*}{\rotatebox[origin=c]{90}{AECE}}

& Zero-Shot~\cite{radford2021learning}
&0.24&0.03&0.16&0.05&0.12\\

& TPT~\cite{shu2022testtime}
&0.21&0.03&0.12&0.05&\textbf{0.10}\\

& C-TPT~\cite{yoon2024ctpt}
&0.23&0.03&0.14&0.05&0.11\\

& Penalty~\cite{murugesan2024robust}
&0.21&0.03&0.12&0.05&0.10\\

& SaLS~\cite{murugesan2024robust}
&0.21&0.03&0.12&0.05&\textbf{0.10}\\

& O-TPT~\cite{sharifdeen2025otpt}
&0.25&0.03&0.16&0.05&0.12\\

& A-TPT~\cite{ahamed2026atpt}
&0.25&0.03&0.15&0.05&0.12\\

& SoC~\cite{fillioux2026soc}
&0.24&0.03&0.14&0.05&0.11\\

&\cellcolor{casbg}\texttt{CoTS}
&\cellcolor{casbg}0.22
&\cellcolor{casbg}0.03
&\cellcolor{casbg}0.14
&\cellcolor{casbg}0.05
&\cellcolor{casbg}0.11\\

&\cellcolor{casbg}\texttt{E-CoTS}
&\cellcolor{casbg}0.21
&\cellcolor{casbg}0.03
&\cellcolor{casbg}0.14
&\cellcolor{casbg}0.05
&\cellcolor{casbg}0.11\\

\bottomrule
\end{tabular}
}
\end{table*}
\textbf{Calibration error.}
To quantify model calibration~\cite{naeini2015obtaining,guo2017calibration}, we primarily adopt the Expected Calibration Error (ECE)~\cite{naeini2015obtaining}, which measures the discrepancy between predicted probabilities and observed frequencies.
Specifically, the predictions are partitioned into $M$ confidence bins $\{B_1, B_2, \dots, B_M\}$ based on the confidence values, where $B_m$ contains all samples whose predicted confidence falls into the $m$-th interval.
The ECE is then computed as a weighted average of the absolute difference between accuracy and confidence within each bin:
\begin{equation}
    \text{ECE} = \sum\limits_{m=1}^{M} \frac{|B_m|}{S} \left| \ \text{acc}(B_m) - \text{conf}(B_m) \right|,
\end{equation}
where $S$ is the total number of samples, $|B_m|$ denotes the number of samples in bin $B_m$, and $\text{acc}(B_m)$ and $\text{conf}(B_m)$ represent the fraction of correctly predicted samples and the mean predicted confidence for the $m$-th bin, respectively.
ECE provides a scalar measure of miscalibration: smaller values indicate better alignment between predicted confidences and empirical accuracies.

\textbf{Brier score.}
We also report the Brier Score (BS)~\cite{brier1950verification} to evaluate the quality of probabilistic predictions.
Different from ECE, which measures the confidence-accuracy discrepancy after binning, the BS directly computes the squared error between the predicted probability distribution and the one-hot ground-truth label.
For a multi-class classification problem with $K$ classes, it is formulated as:
\begin{equation}
    \text{BS} = \frac{1}{S} \sum_{i=1}^{S} \sum_{c=1}^{K}
    \left( p_{i,c} - y_{i,c} \right)^2,
\end{equation}
where $S$ denotes the total number of samples, $p_{i,c}$ represents the predicted probability of sample $i$ for class $c$, and $y_{i,c}$ is the corresponding one-hot ground-truth label.
A lower BS indicates that the predicted probability distribution is closer to the true label distribution.
The BS values of different methods with ViT-B/16 are shown in Table~\ref{tab:main_brier} and Table~\ref{tab:main_brier2}. Our methods achieve competitive Brier scores across datasets.

\textbf{Class-wise ECE.}
We additionally report the Class-wise Expected Calibration Error (CECE)~\cite{kull2019beyond}, which evaluates calibration across all classes rather than only the predicted class.
It is computed as:
\begin{equation}
    \text{CECE}
    =
    \sum_{m=1}^{M}
    \sum_{c=1}^{K}
    \frac{|B_{m,c}|}{SK}
    \left|
    \text{acc}_{c}(B_{m,c})
    -
    \text{conf}_{c}(B_{m,c})
    \right|,
\end{equation}
where $B_{m,c}$ denotes the $m$-th confidence bin for class $c$.
Lower CECE indicates better class-wise calibration.
The results are reported in Table~\ref{tab:cece_fg} and Table~\ref{tab:cece_imagenet}.

\textbf{Adaptive ECE.}
We also report Adaptive Expected Calibration Error (AECE)~\cite{nixon2019measuring}, which uses adaptive bins with approximately equal numbers of samples to reduce the sensitivity to fixed-width binning.
It is defined as:
\begin{equation}
    \text{AECE}
    =
    \sum_{r=1}^{R}
    \sum_{c=1}^{K}
    \frac{1}{RK}
    \left|
    \text{acc}_{c}(B_{r,c})
    -
    \text{conf}_{c}(B_{r,c})
    \right|.
\end{equation}
Lower AECE indicates better calibration.
The results are reported in Table~\ref{tab:aece_fg} and Table~\ref{tab:aece_imagenet}.

\section{Additional Analysis}
\subsection{Sensitivity to Hand-crafted Prompt Initialization}
\label{sec:sens_to_prompt}

\begin{table}[t]
\centering
\caption{Performance comparison under different initial prompts on ImageNet variants. Accuracy (\%) and ECE (\%) are reported.}
\label{tab:prompt_imagenet_variants}
\scriptsize
\setlength{\tabcolsep}{5pt}
\renewcommand{\arraystretch}{1.12}
\resizebox{\linewidth}{!}{
\begin{tabular}{lcccccc>{\columncolor{casbg}}c>{\columncolor{casbg}}c}
\toprule
\multirow{2}{*}{Prompt template} & Metric & Zero-Shot & TPT & SaLS & A-TPT & SoC & \texttt{CoTS} & \texttt{E-CoTS} \\
& (\%) & \cite{radford2021learning} & \cite{shu2022testtime} & \cite{murugesan2024robust} & \cite{ahamed2026atpt} & \cite{fillioux2026soc} & & \\
\midrule
\multirow{2}{*}{\texttt{"a\_bad\_photo\_of\_a"}}
& ACC & 58.60 & 61.61 & 61.61 & 57.58 & 59.03 & 61.61 & 63.66 \\
& ECE & 4.47  & 12.07 & 10.61 & 5.13  & 5.73  & 5.16  & 5.12  \\
\midrule
\multirow{2}{*}{\texttt{"a\_black\_and\_white\_photo\_of\_a"}}
& ACC & 55.46 & 59.30 & 59.30 & 56.84 & 57.09 & 59.30 & 60.69 \\
& ECE & 6.50  & 18.07 & 16.67 & 7.88  & 8.06  & 5.95  & 7.23  \\
\midrule
\multirow{2}{*}{\texttt{"a\_blurry\_photo\_of\_a"}}
& ACC & 57.72 & 61.11 & 61.11 & 58.35 & 59.10 & 61.11 & 62.82 \\
& ECE & 4.87  & 15.67 & 12.72 & 10.27 & 8.74  & 5.32  & 5.25  \\
\midrule
\multirow{2}{*}{\texttt{ "a\_good\_photo\_of\_a"}}
& ACC & 57.67 & 61.34 & 61.34 & 58.48 & 59.47 & 61.34 & 63.21 \\
& ECE & 4.67  & 14.90 & 12.84 & 8.14  & 7.86  & 5.35  & 5.57 \\
\midrule
\multirow{2}{*}{\texttt{"a\_high\_contrast\_photo\_of\_a"}}
& ACC & 56.35 & 60.20 & 60.20 & 57.21 & 57.79 & 60.20 & 61.59 \\
& ECE & 5.49  & 19.30 & 16.57 & 8.84  & 9.62  & 5.43  & 6.20  \\
\midrule
\multirow{2}{*}{\texttt{"a\_low\_contrast\_photo\_of\_a"}}
& ACC & 57.63 & 61.28 & 61.28 & 56.57 & 58.28 & 61.28 & 62.82 \\
& ECE & 5.21  & 17.34 & 8.12  & 8.60  & 14.98 & 5.31  & 5.55 \\
\midrule
\multirow{2}{*}{\texttt{"a\_photo\_of\_a\_small"}}
& ACC & 56.25 & 60.10 & 60.10 & 58.88 & 58.15 & 60.10 & 61.67 \\
& ECE & 4.19  & 14.97 & 12.61 & 5.10  & 6.93  & 5.07  & 4.80 \\
\bottomrule
\end{tabular}
}
\end{table}

We evaluate the performance of different methods under multiple initial prompt templates on ImageNet variants.
Table~\ref{tab:prompt_imagenet_variants} reports ACC (\%) and ECE (\%) for each method, showing how \cots~and \ecots~consistently maintain low ECE while achieving high accuracy across prompts.

\subsection[Ensemble Weight alpha in \ecots]{Ensemble Weight $\alpha$ in \ecots}
\label{sec:alpha_ig}
\begin{figure}[t]
    \centering
    \includegraphics[width=0.6\linewidth]{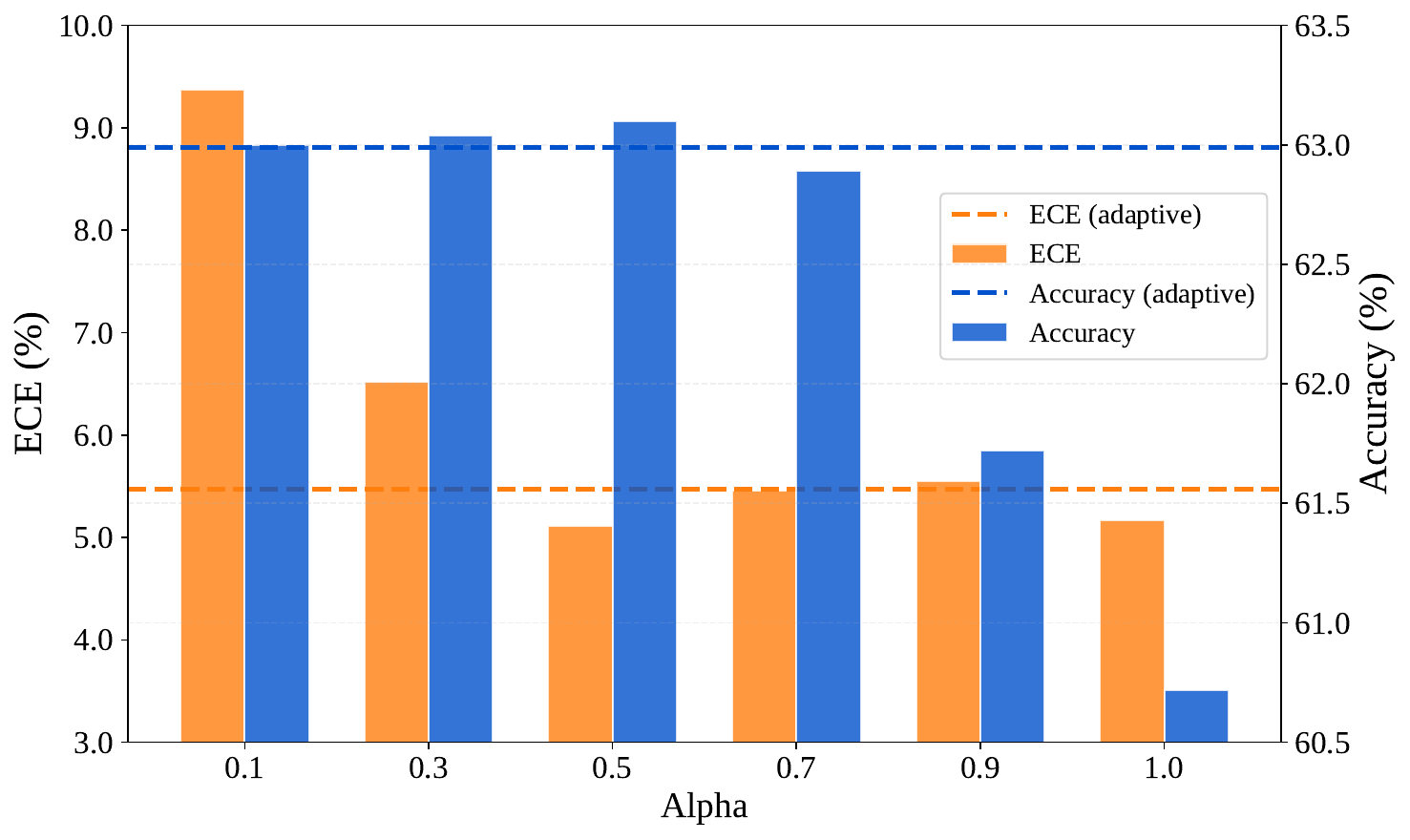}
    \caption{Effect of ensemble weight $\alpha$ on ImageNet variants using ViT-B/16. Accuracy (\%) and ECE (\%) are reported.}
    \label{fig:alpha_imagenet_variants}
\end{figure}

As shown in Fig.~\ref{fig:alpha_imagenet_variants}, the adaptive weighting strategy achieves a better balance between accuracy and calibration, avoiding the need to tune $\alpha$ for different settings.

\begin{figure}[t]
    \centering
    \includegraphics[width=0.6\linewidth]{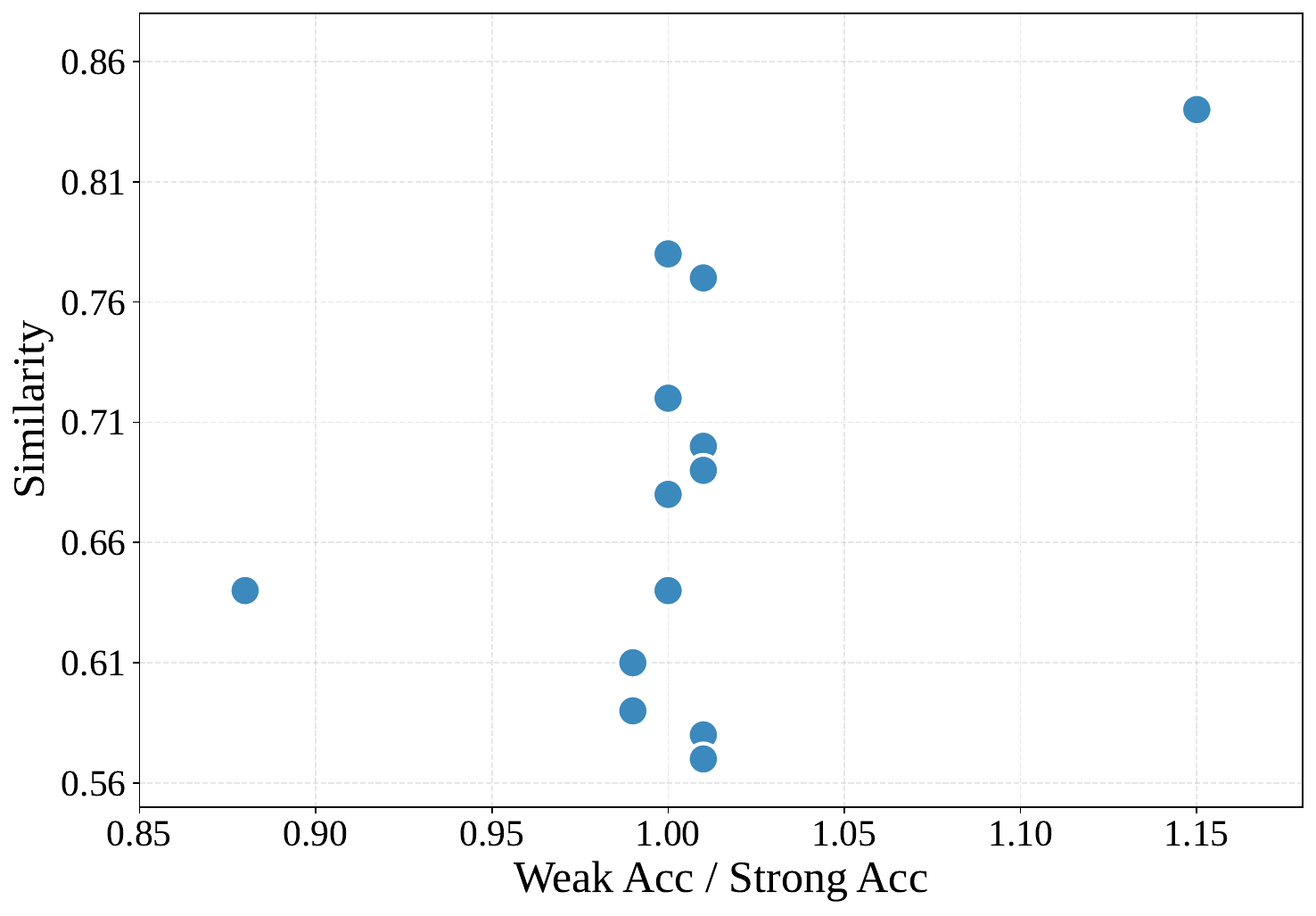}
    \caption{Weak-strong accuracy ratio versus dataset-level similarity across different datasets.}
    \label{fig:sim_scatter}
\end{figure}

\subsection{Similarity-based Analysis}
Figure~\ref{fig:sim_scatter} shows the weak-strong accuracy ratio plotted against dataset-level similarity.
This result demonstrates the effectiveness of the proposed weak-strong ensemble strategy.

\subsection{Step-wise Analysis}
\begin{figure}[t]
    \centering
    \begin{subfigure}[b]{0.48\linewidth}
        \centering
        \includegraphics[width=\linewidth]{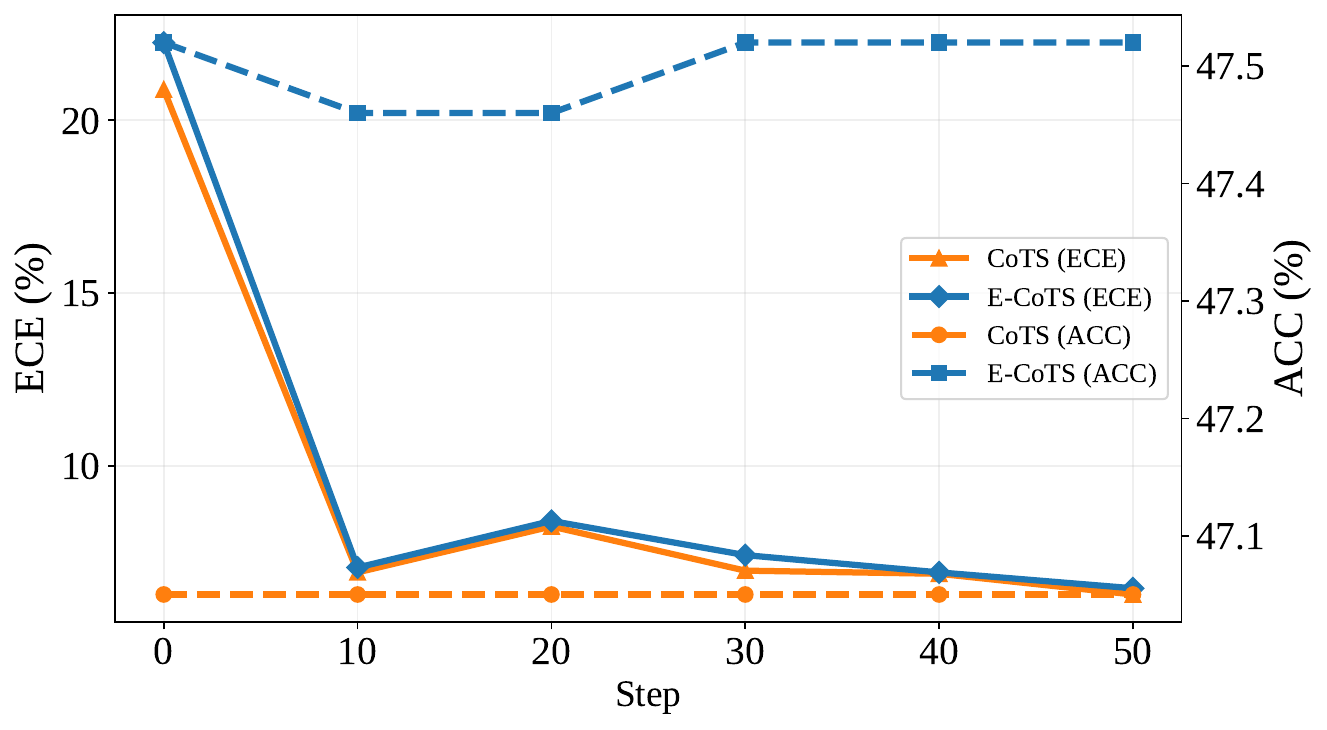}
        \caption{DTD}
        \label{fig:iter_dtd}
    \end{subfigure}
    \hfill
    \begin{subfigure}[b]{0.48\linewidth}
        \centering
        \includegraphics[width=\linewidth]{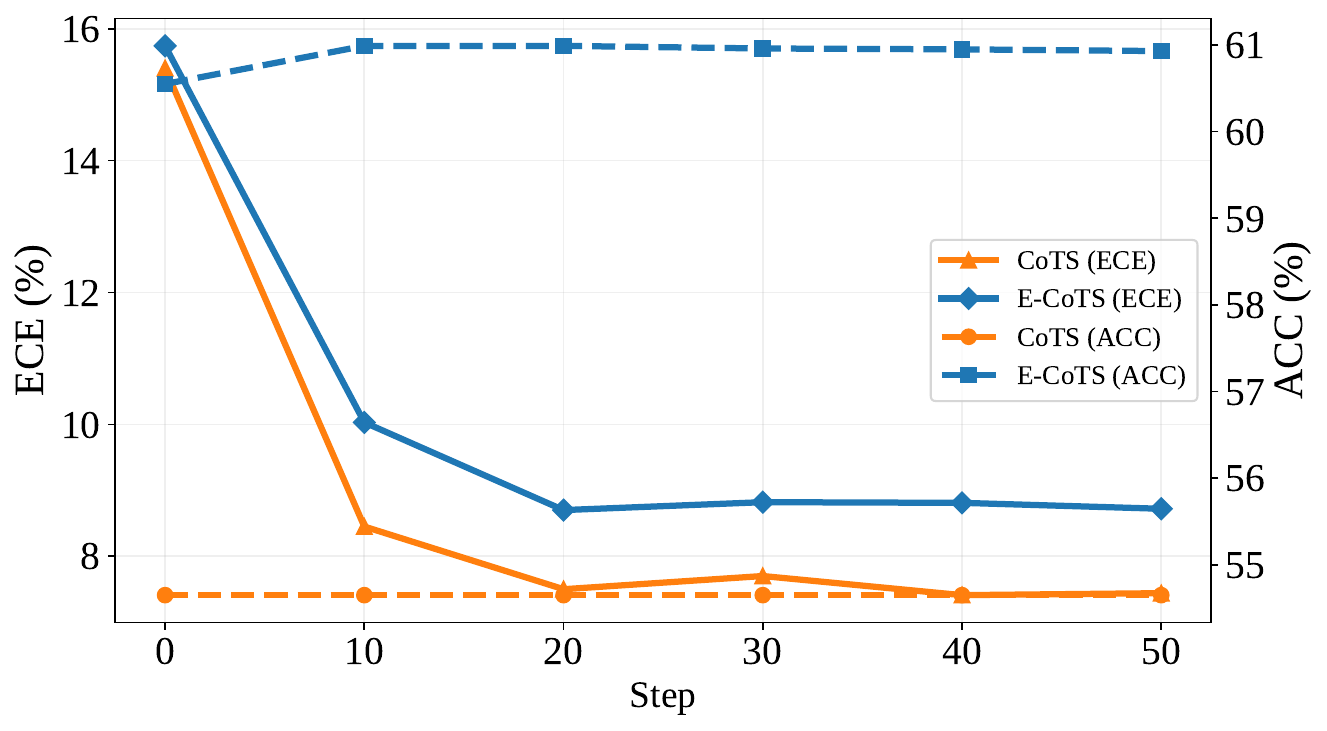}
        \caption{ImageNet-A}
        \label{fig:iter_A}
    \end{subfigure}

    \caption{ACC(\%) and ECE(\%) curves of \cots~and \ecots~on DTD and ImageNet-A over iteration steps using ViT-B/16.}
    \label{fig:step}
\end{figure}

Figure~\ref{fig:step} presents ECE and ACC of \cots~and \ecots~over iteration steps on DTD and ImageNet-A.
As the number of steps increases, ECE tends to converge while ACC remains stable.

\subsection{Sensitivity to \texorpdfstring{$K$ and $T_0$}{K and T0}}

\begin{figure}[t]
    \centering
    \includegraphics[width=\linewidth]{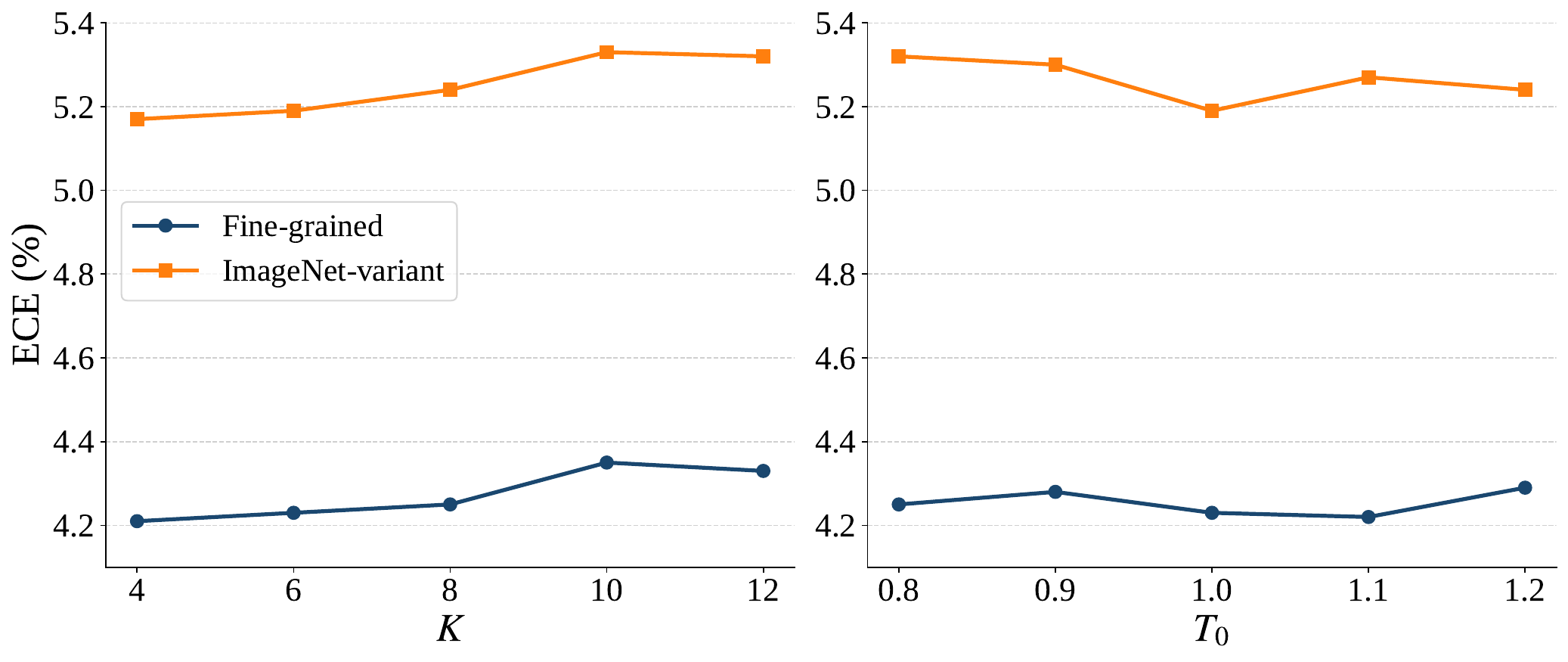}
    \caption{Sensitivity analysis of \texttt{CoTS} to the number of selected strong views $K$ and the temperature initialization $T_0$. ECE (\%) is reported on fine-grained datasets and ImageNet variants.}
    \label{fig:hyper_ablation_K_T0}
\end{figure}

We further investigate the sensitivity to the number of selected strong views $K$ and the temperature initialization $T_0$. Figure~\ref{fig:hyper_ablation_K_T0} show that ~\texttt{CoTS} is generally robust to variations in both hyperparameters. We set $K=6$ and $T_0=1.0$ by default throughout the experiments.

\subsection{Comparison between Ensemble-Only and \ecots}
\label{sec:ensemble}
We further compare base methods, ensemble-only, and \ecots~across different TPT-based calibration methods (Table~\ref{tab:ens_ecots_ablation}).
Ensemble-only often improves accuracy but can sometimes increase ECE.
By integrating \cots~into the ensemble prediction, \ecots~reduces ECE while retaining the accuracy gains from ensembling.

\subsection{Comparison with ZERO and Our Ensemble Strategy}
Table~\ref{tab:zero_ensemble_comparison} compares ZERO~\cite{farina2024frustratingly} with our ensemble strategy under both zero-shot and TPT settings. Our ensemble achieves a better accuracy--calibration trade-off than ZERO, while ~\texttt{E-CoTS} further substantially reduces ECE and maintains competitive accuracy.

\begin{table*}[t]
\centering
\caption{Performance of base, direct ensemble, and \texttt{E-CoTS} across different TPT-based methods using ViT-B/16. Accuracy (\%) and ECE (\%) are reported on fine-grained datasets and ImageNet variants.}
\label{tab:ens_ecots_ablation}
\renewcommand{\arraystretch}{1.1}
\setlength{\tabcolsep}{4.2pt}
\scriptsize
\resizebox{\linewidth}{!}{
\begin{tabular}{llcccccccccc}
\toprule
\multirow{3}{*}{(\%)}
& \multirow{3}{*}{Method}
& \multicolumn{5}{c}{Fine-grained datasets}
& \multicolumn{5}{c}{ImageNet variants} \\
\cmidrule(lr){3-7} \cmidrule(lr){8-12}
& & TPT & C-TPT & O-TPT & A-TPT & SoC
& TPT & C-TPT & O-TPT & A-TPT & SoC \\
& & \cite{shu2022testtime} & \cite{yoon2024ctpt} & \cite{sharifdeen2025otpt} & \cite{ahamed2026atpt} & \cite{fillioux2026soc} & \cite{shu2022testtime} & \cite{yoon2024ctpt} & \cite{sharifdeen2025otpt} & \cite{ahamed2026atpt} & \cite{fillioux2026soc} \\
\midrule
\multirow{3}{*}{\rotatebox[origin=c]{90}{ACC}}
& base
& 65.25 & 64.57 & 64.04 & 64.15 & 63.96 & 60.72 & 59.26 & 57.51 & 58.18 & 58.17 \\
& Ensemble
& 65.32 & 65.27 & 64.99 & 65.12 & 65.11 & 62.90 & 62.66 & 62.03 & 62.41 & 61.84 \\
& \texttt{E-CoTS}
& 65.28 & 65.24 & 64.95 & 65.09 & 65.11 & 62.99 & 62.67 & 61.97 & 62.28 & 61.81 \\
\midrule
\multirow{3}{*}{\rotatebox[origin=c]{90}{ECE}}
& base
& 11.25 & 5.24  & 4.80  & 4.27  & 4.30  & 12.06 & 6.74  & 4.99  & 4.37  & 4.83  \\
& Ensemble
& 12.28 & 5.35  & 4.90  & 4.00  & 4.12  & 13.36 & 7.86  & 5.36  & 5.38  & 4.08  \\
& \texttt{E-CoTS}
& 4.18  & 4.19  & 4.29  & 3.86  & 4.18  & 5.47  & 5.18  & 4.51  & 4.84  & 4.66 \\
\bottomrule
\end{tabular}
}
\end{table*}

\begin{table}[t]
\centering
\caption{Comparison of ZERO~\cite{farina2024frustratingly} and our ensemble strategy using ViT-B/16. 
Accuracy (\%) and ECE (\%) are reported on fine-grained datasets and ImageNet variants.}
\label{tab:zero_ensemble_comparison}
\renewcommand{\arraystretch}{1.00}
\setlength{\tabcolsep}{5pt}
\scriptsize
\resizebox{0.8\linewidth}{!}{
\begin{tabular}{lcccc}
\toprule
\multirow{2}{*}{Method}
& \multicolumn{2}{c}{Fine-grained datasets}
& \multicolumn{2}{c}{ImageNet variants} \\
\cmidrule(lr){2-3}
\cmidrule(lr){4-5}
& ACC (\%) & ECE (\%) & ACC (\%) & ECE (\%) \\
\midrule

Zero-Shot~\cite{radford2021learning}
&63.95&4.33&57.22&\textbf{4.98}\\

w/ ZERO~\cite{farina2024frustratingly}
&64.41&18.24&\textbf{62.42}&19.54\\

\rowcolor{casbg}
w/ Ensemble (Ours)
&\textbf{65.21}&\textbf{4.00}&61.86&5.03\\

\midrule

TPT~\cite{shu2022testtime}
&65.19&11.30&60.74&11.90\\

w/ ZERO~\cite{farina2024frustratingly}
&64.65&27.26&\textbf{62.97}&28.34\\

\rowcolor{casbg}
w/ Ensemble (Ours)
&\textbf{65.32}&12.28&62.90&13.36\\

\rowcolor{casbg}
w/ \texttt{E-CoTS}
&65.22&\textbf{4.31}&62.95&\textbf{5.38}\\

\bottomrule
\end{tabular}
}
\end{table}

\clearpage
\section{Reliability Diagram}
\begin{figure*}[t]
    \centering

    \begin{subfigure}[t]{0.21\textwidth}
        \centering
        \includegraphics[width=\linewidth]{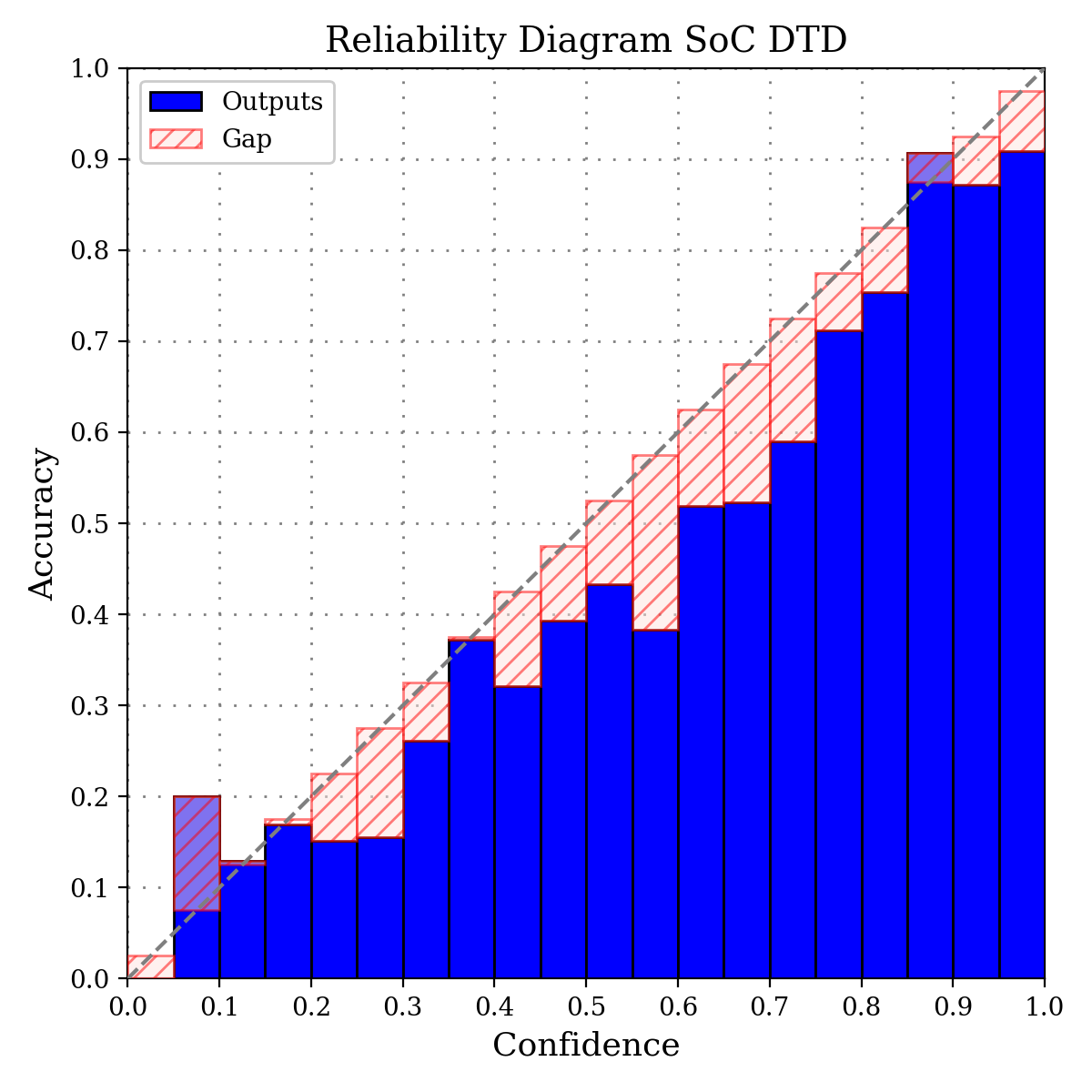}
        \caption{SoC~\cite{fillioux2026soc}: DTD}
        \label{fig:soc_dtd}
    \end{subfigure}
    \hfill
    \begin{subfigure}[t]{0.21\textwidth}
        \centering
        \includegraphics[width=\linewidth]{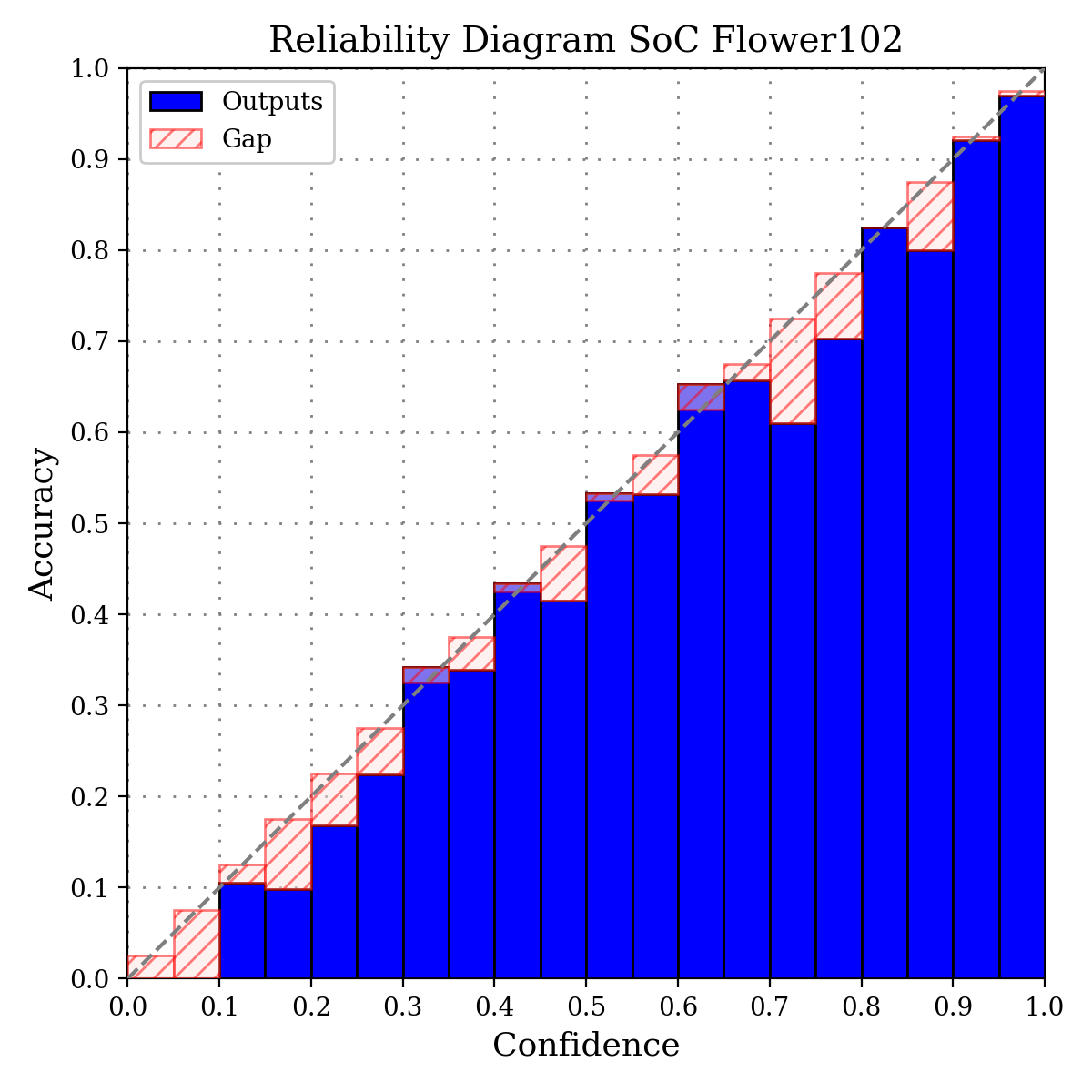}
        \caption{SoC~\cite{fillioux2026soc}: Flowers}
        \label{fig:soc_flower102}
    \end{subfigure}
    \hfill
    \begin{subfigure}[t]{0.21\textwidth}
        \centering
        \includegraphics[width=\linewidth]{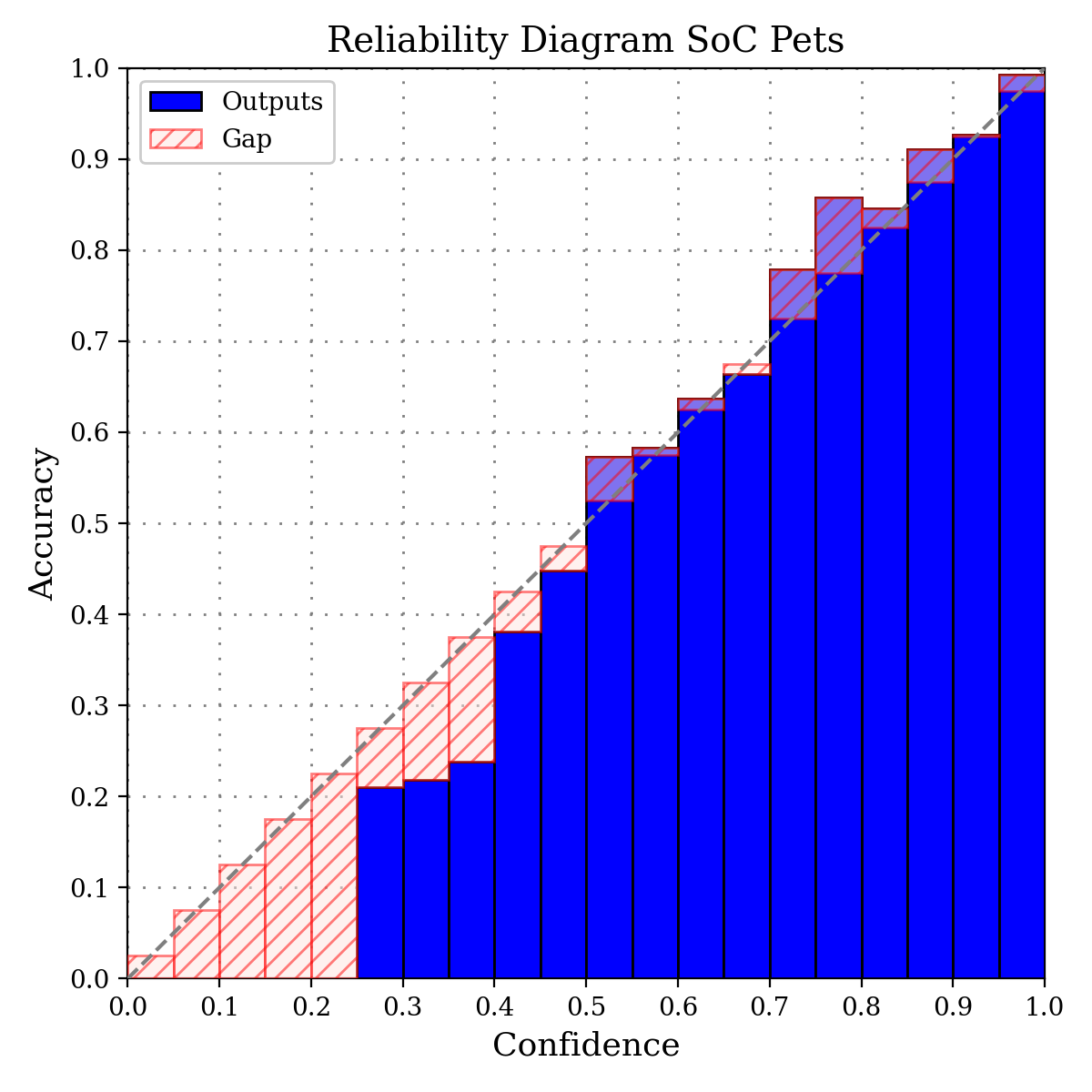}
        \caption{SoC~\cite{fillioux2026soc}: Pets}
        \label{fig:soc_pets}
    \end{subfigure}
    \hfill
    \begin{subfigure}[t]{0.21\textwidth}
        \centering
        \includegraphics[width=\linewidth]{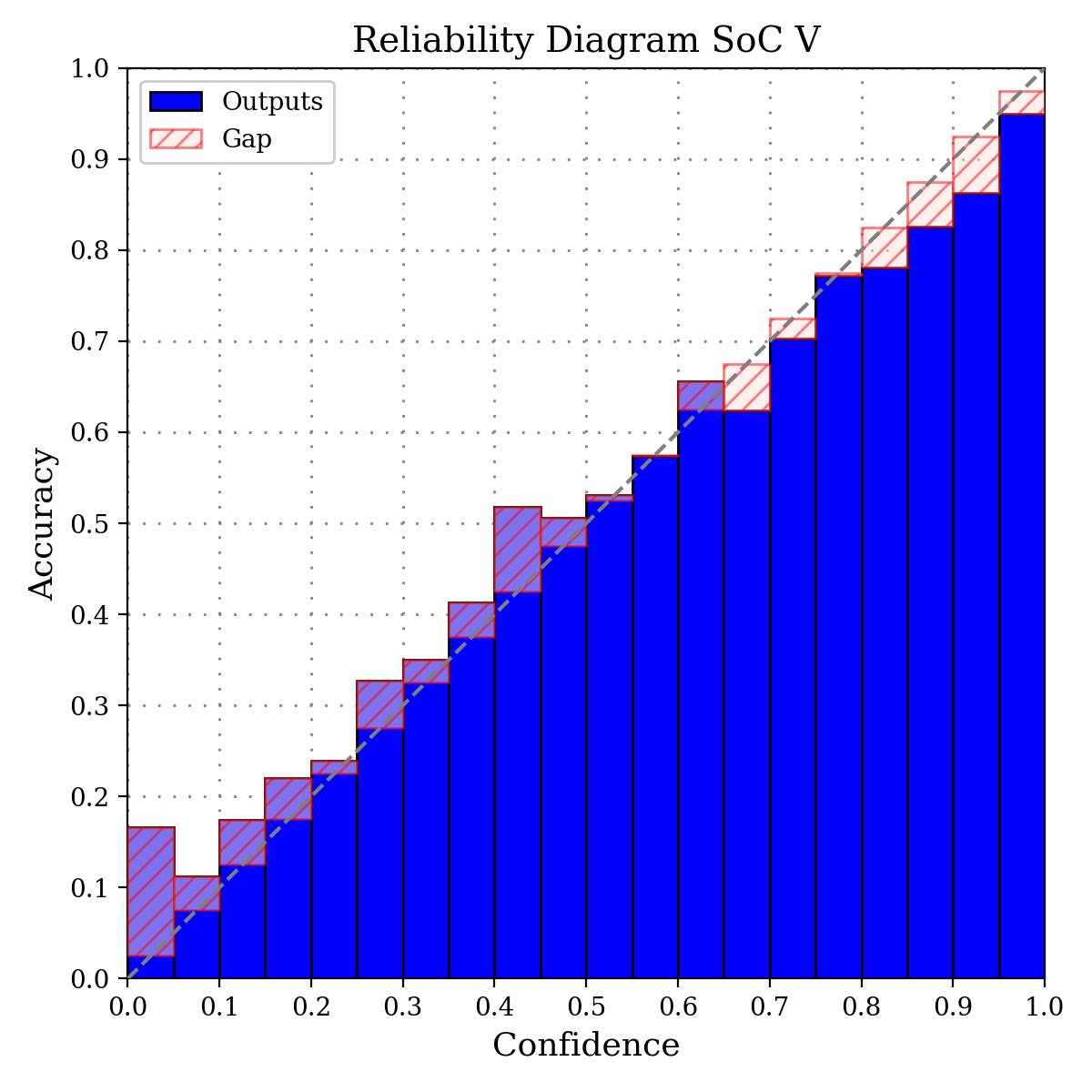}
        \caption{SoC~\cite{fillioux2026soc}: IN-V}
        \label{fig:soc_v}
    \end{subfigure}

    \vspace{0.15em}

    \begin{subfigure}[t]{0.21\textwidth}
        \centering
        \includegraphics[width=\linewidth]{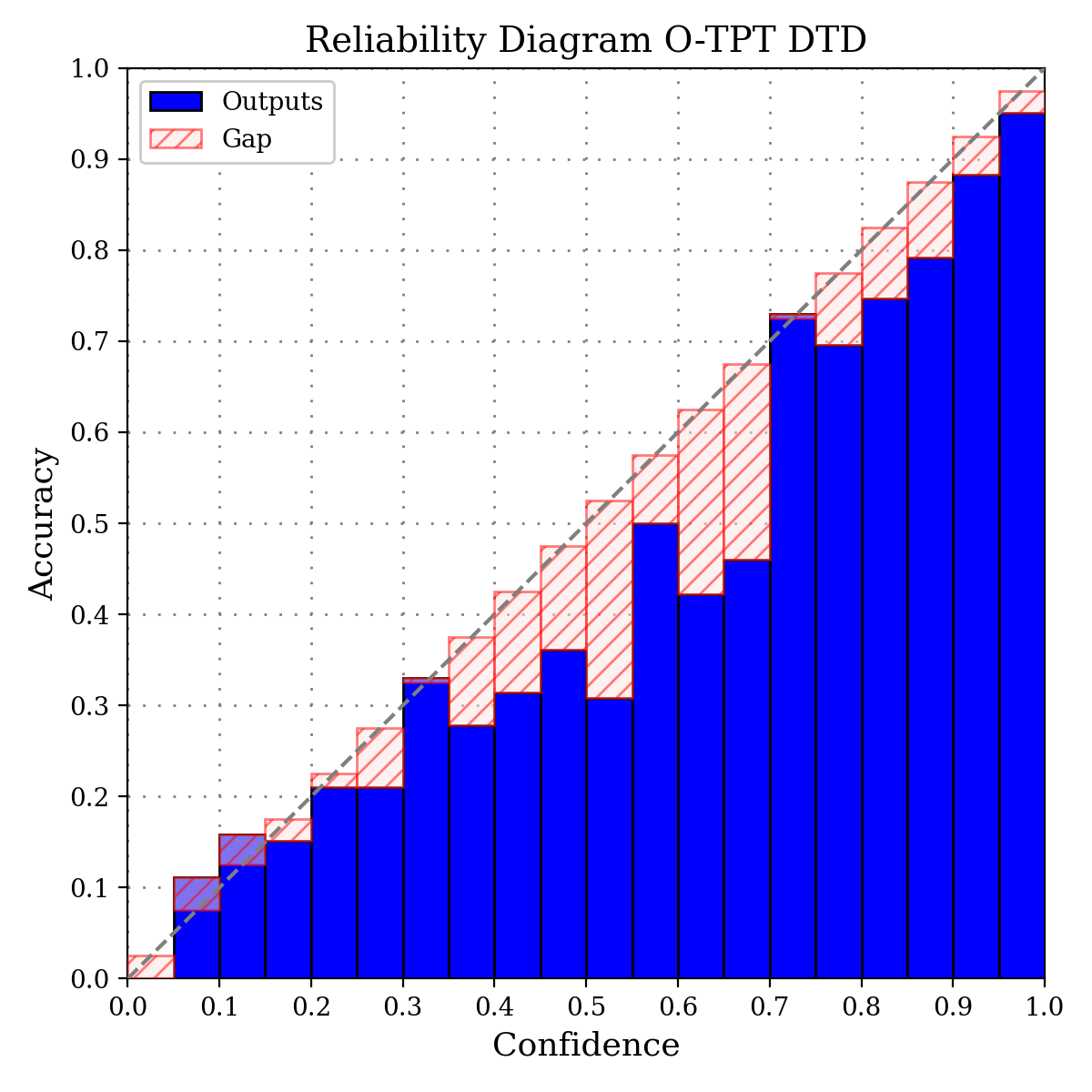}
        \caption{O-TPT~\cite{sharifdeen2025otpt}: DTD}
        \label{fig:otpt_dtd}
    \end{subfigure}
    \hfill
    \begin{subfigure}[t]{0.21\textwidth}
        \centering
        \includegraphics[width=\linewidth]{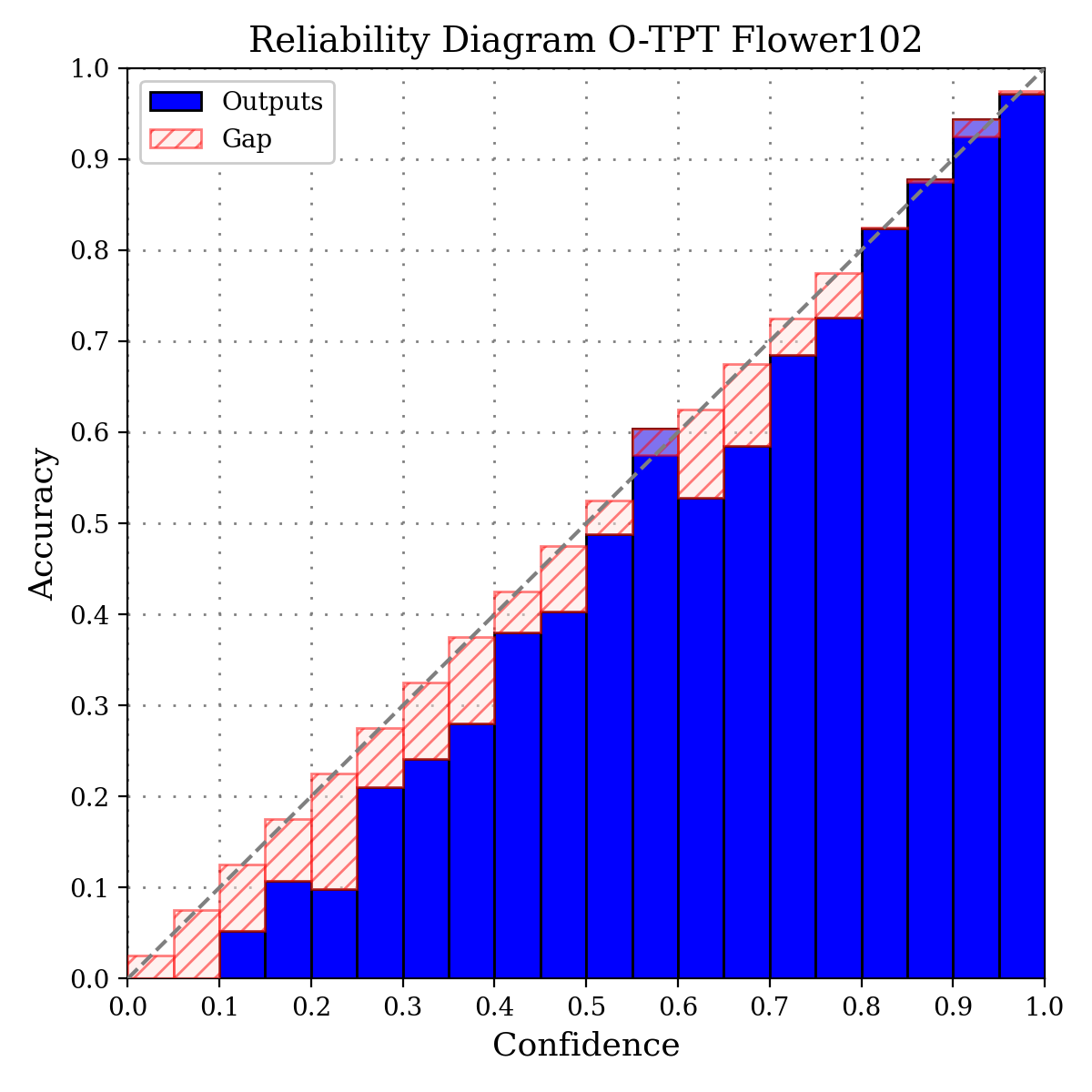}
        \caption{O-TPT~\cite{sharifdeen2025otpt}: Flowers}
        \label{fig:otpt_flower102}
    \end{subfigure}
    \hfill
    \begin{subfigure}[t]{0.21\textwidth}
        \centering
        \includegraphics[width=\linewidth]{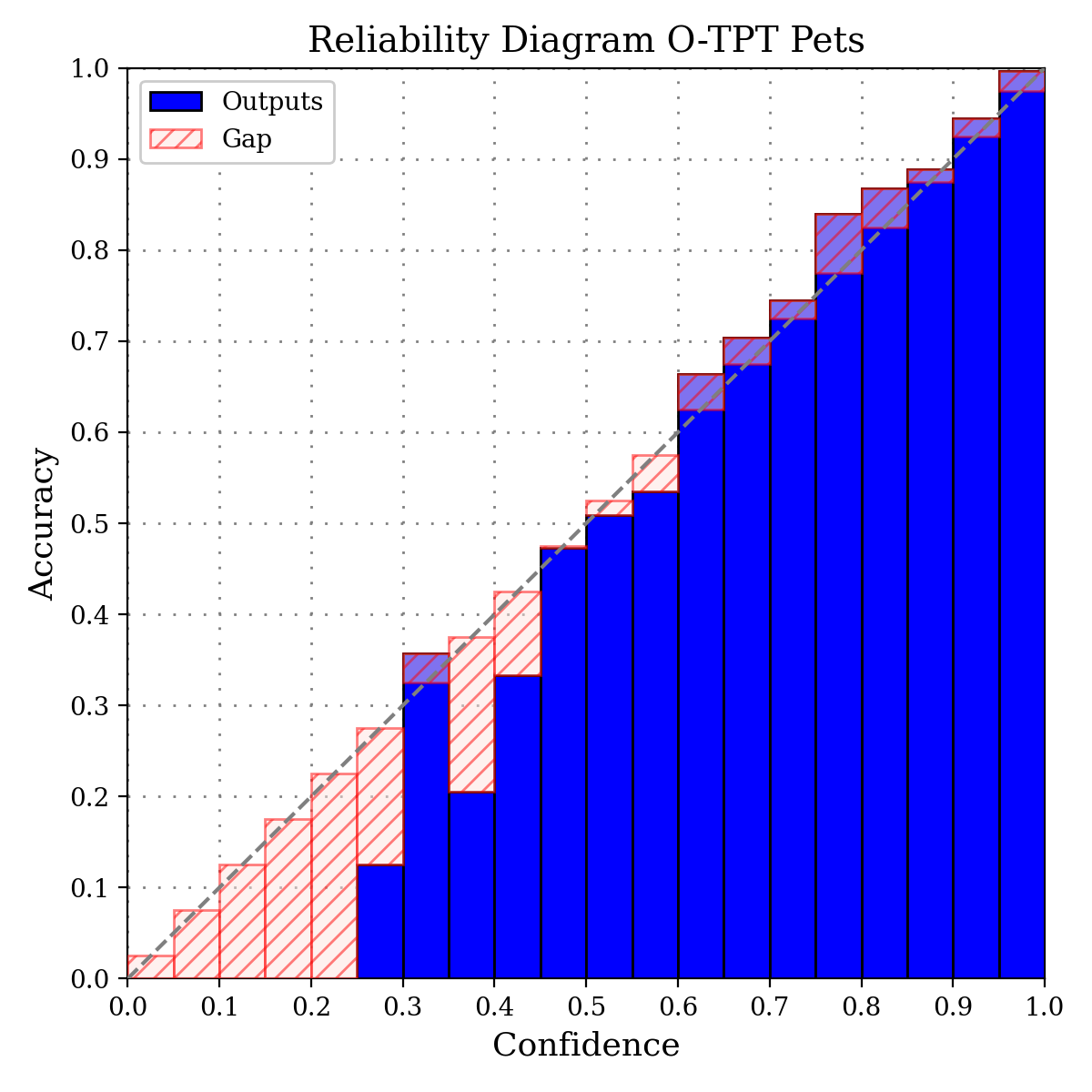}
        \caption{O-TPT~\cite{sharifdeen2025otpt}: Pets}
        \label{fig:otpt_pets}
    \end{subfigure}
    \hfill
    \begin{subfigure}[t]{0.21\textwidth}
        \centering
        \includegraphics[width=\linewidth]{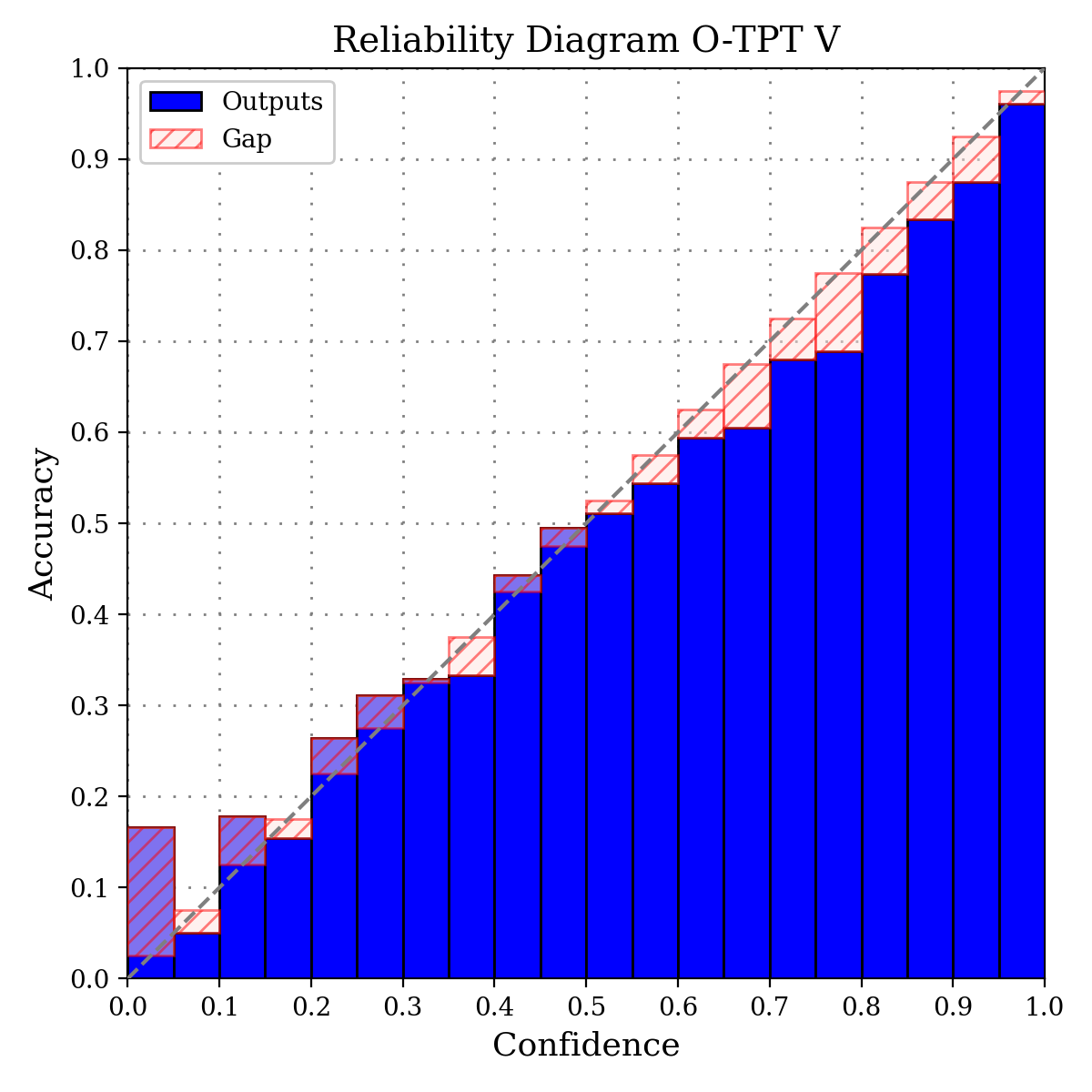}
        \caption{O-TPT~\cite{sharifdeen2025otpt}: IN-V}
        \label{fig:otpt_v}
    \end{subfigure}

    \vspace{0.15em}

    \begin{subfigure}[t]{0.21\textwidth}
        \centering
        \includegraphics[width=\linewidth]{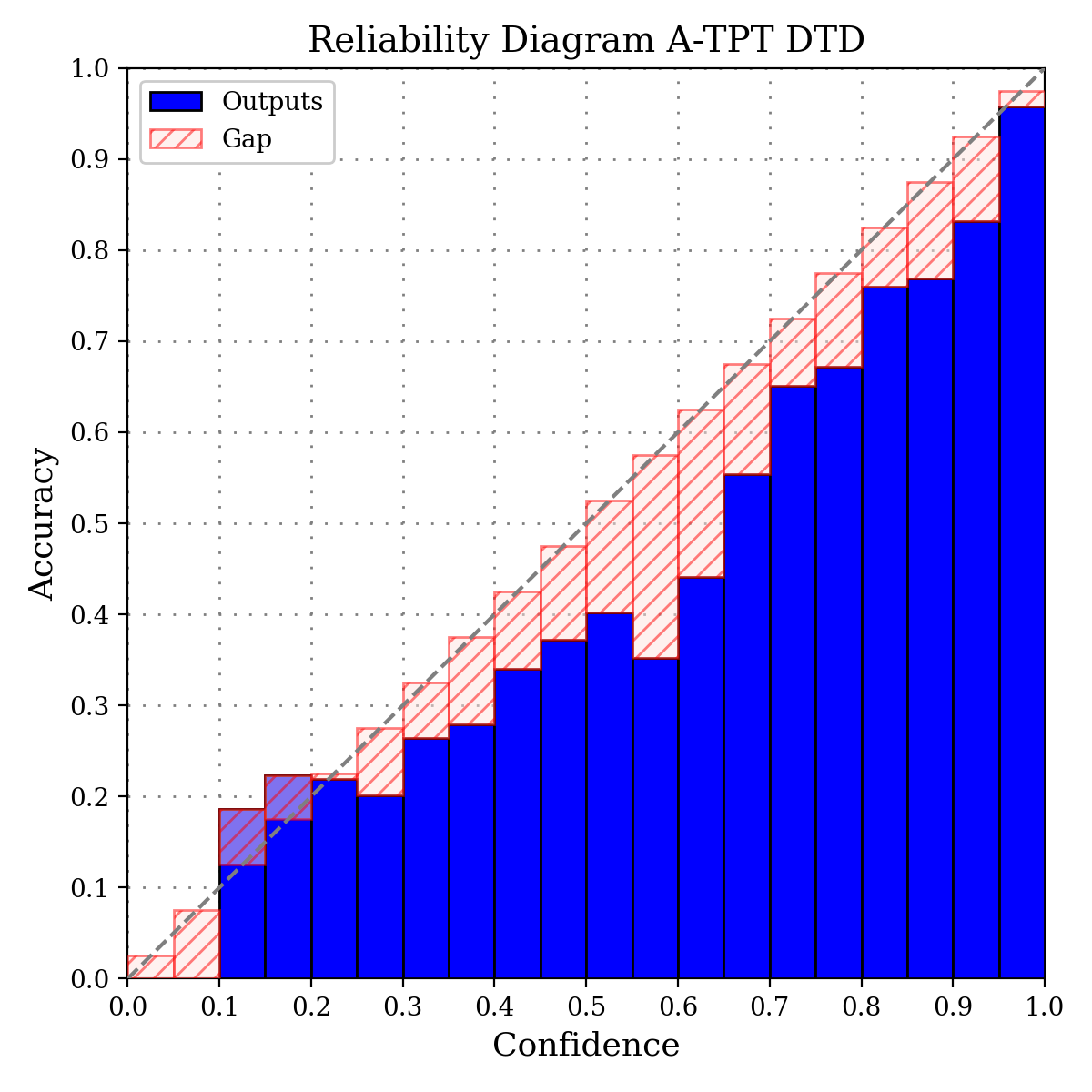}
        \caption{A-TPT~\cite{ahamed2026atpt}: DTD}
        \label{fig:atpt_dtd}
    \end{subfigure}
    \hfill
    \begin{subfigure}[t]{0.21\textwidth}
        \centering
        \includegraphics[width=\linewidth]{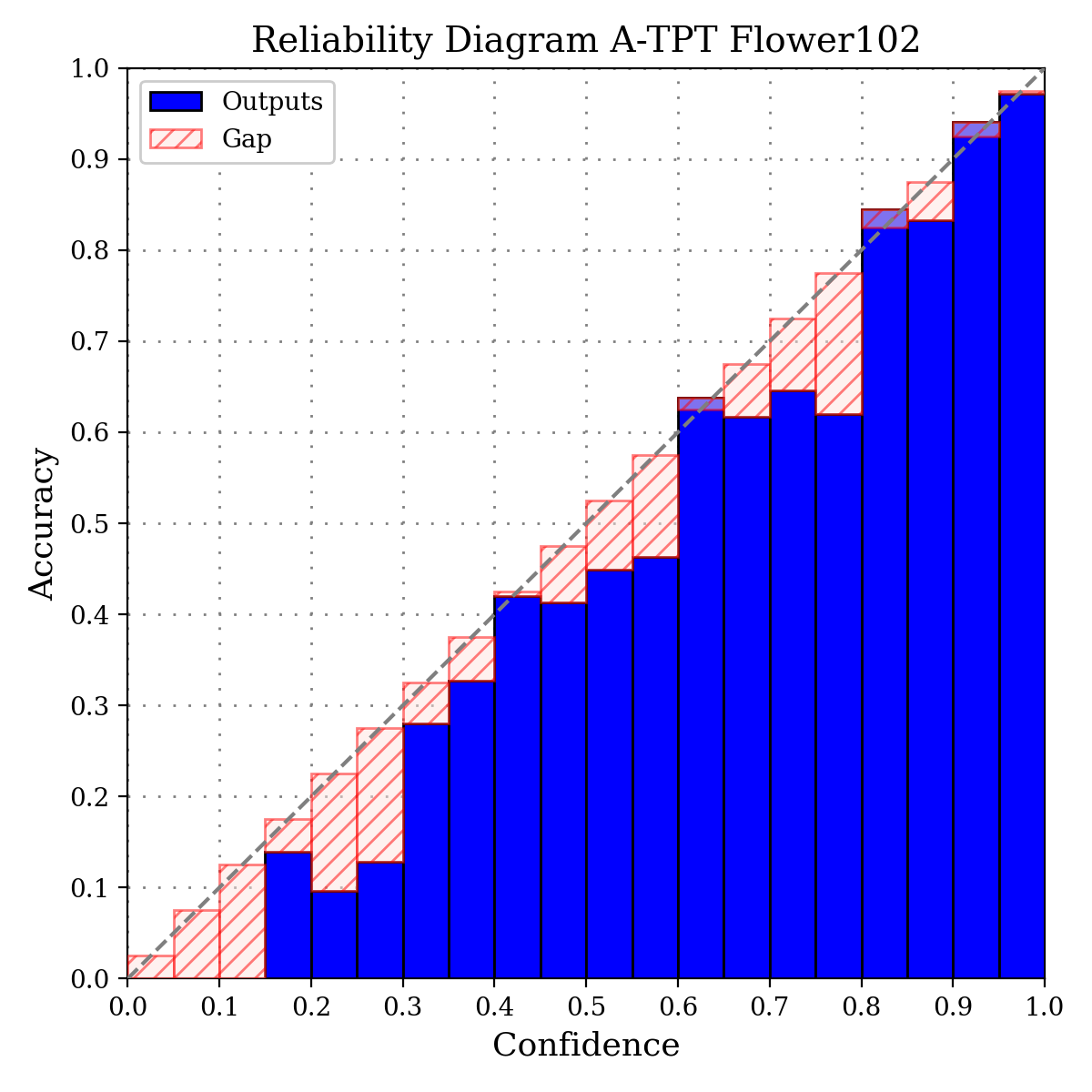}
        \caption{A-TPT~\cite{ahamed2026atpt}: Flowers}
        \label{fig:atpt_flower102}
    \end{subfigure}
    \hfill
    \begin{subfigure}[t]{0.21\textwidth}
        \centering
        \includegraphics[width=\linewidth]{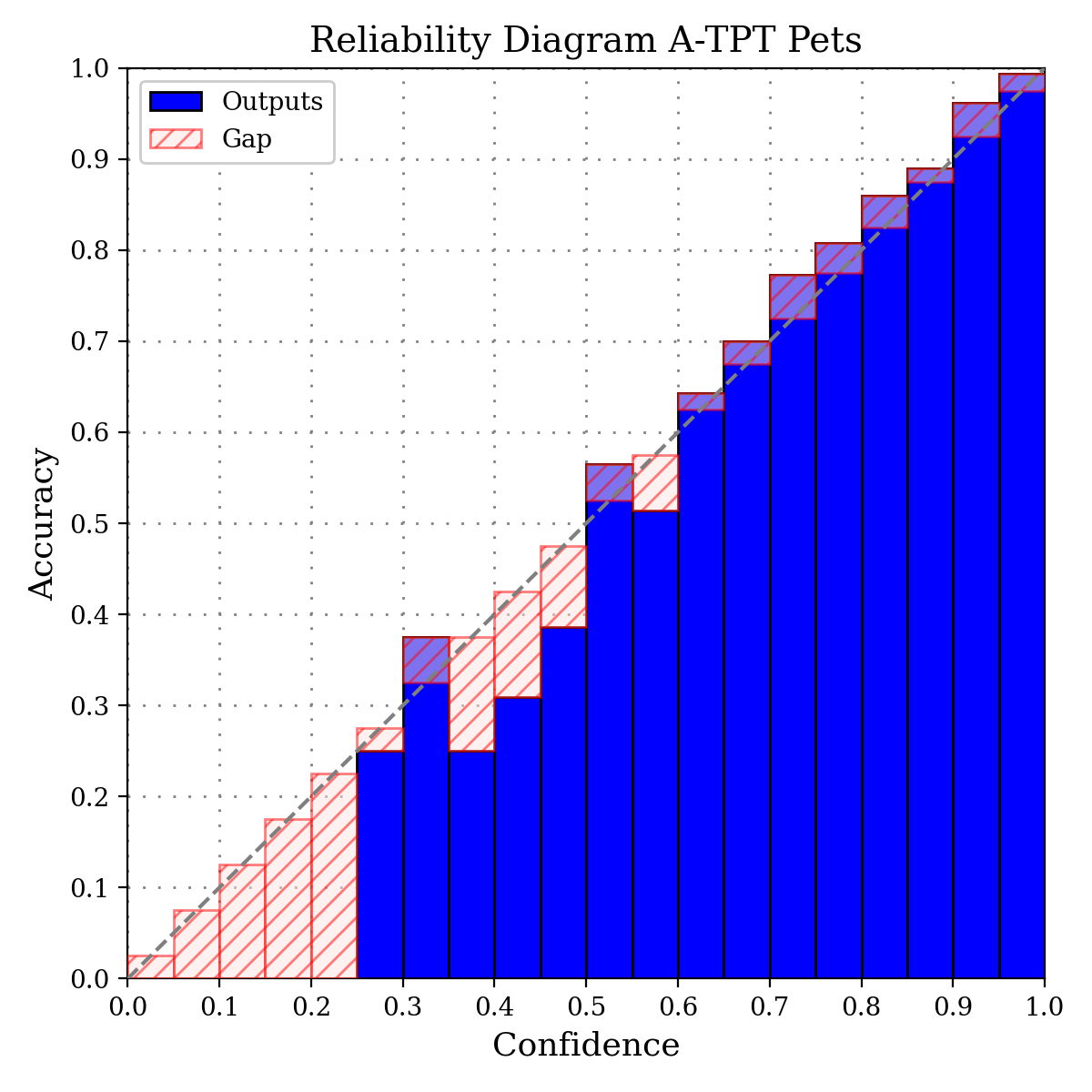}
        \caption{A-TPT~\cite{ahamed2026atpt}: Pets}
        \label{fig:atpt_pets}
    \end{subfigure}
    \hfill
    \begin{subfigure}[t]{0.21\textwidth}
        \centering
        \includegraphics[width=\linewidth]{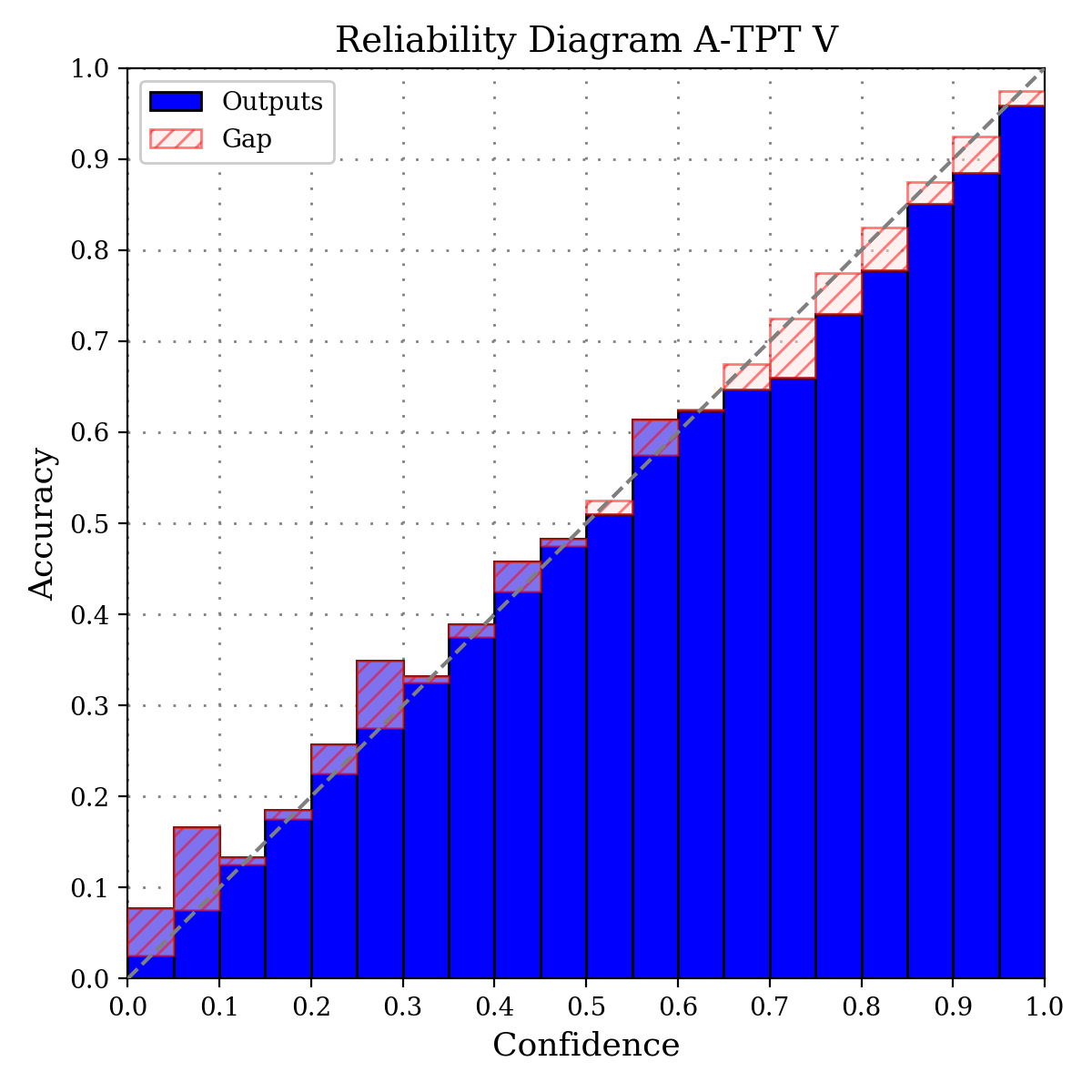}
        \caption{A-TPT~\cite{ahamed2026atpt}: IN-V}
        \label{fig:atpt_v}
    \end{subfigure}

    \vspace{0.15em}

    \begin{subfigure}[t]{0.21\textwidth}
        \centering
        \includegraphics[width=\linewidth]{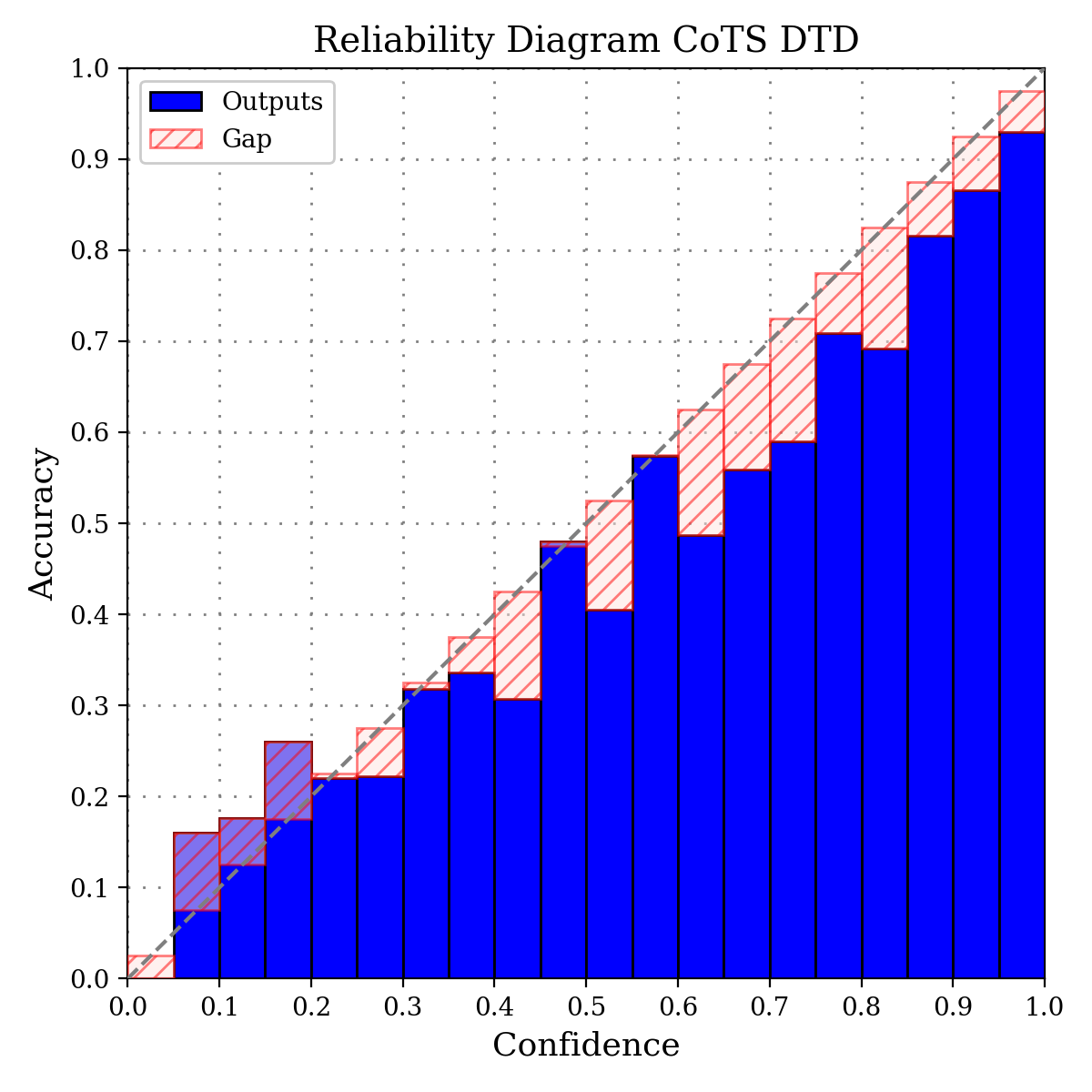}
        \caption{\cots: DTD}
        \label{fig:cots_dtd}
    \end{subfigure}
    \hfill
    \begin{subfigure}[t]{0.21\textwidth}
        \centering
        \includegraphics[width=\linewidth]{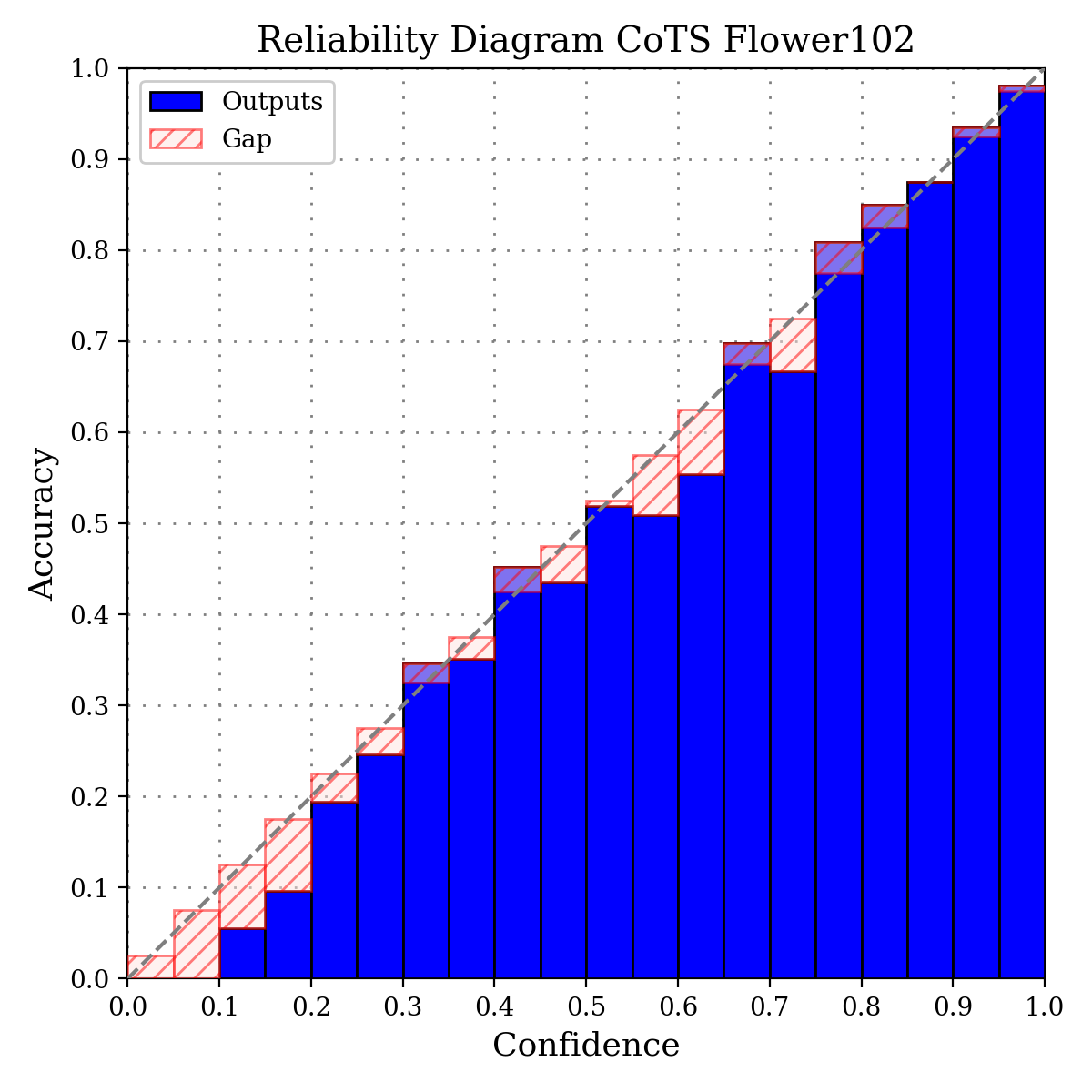}
        \caption{\cots: Flowers}
        \label{fig:cots_flower102}
    \end{subfigure}
    \hfill
    \begin{subfigure}[t]{0.21\textwidth}
        \centering
        \includegraphics[width=\linewidth]{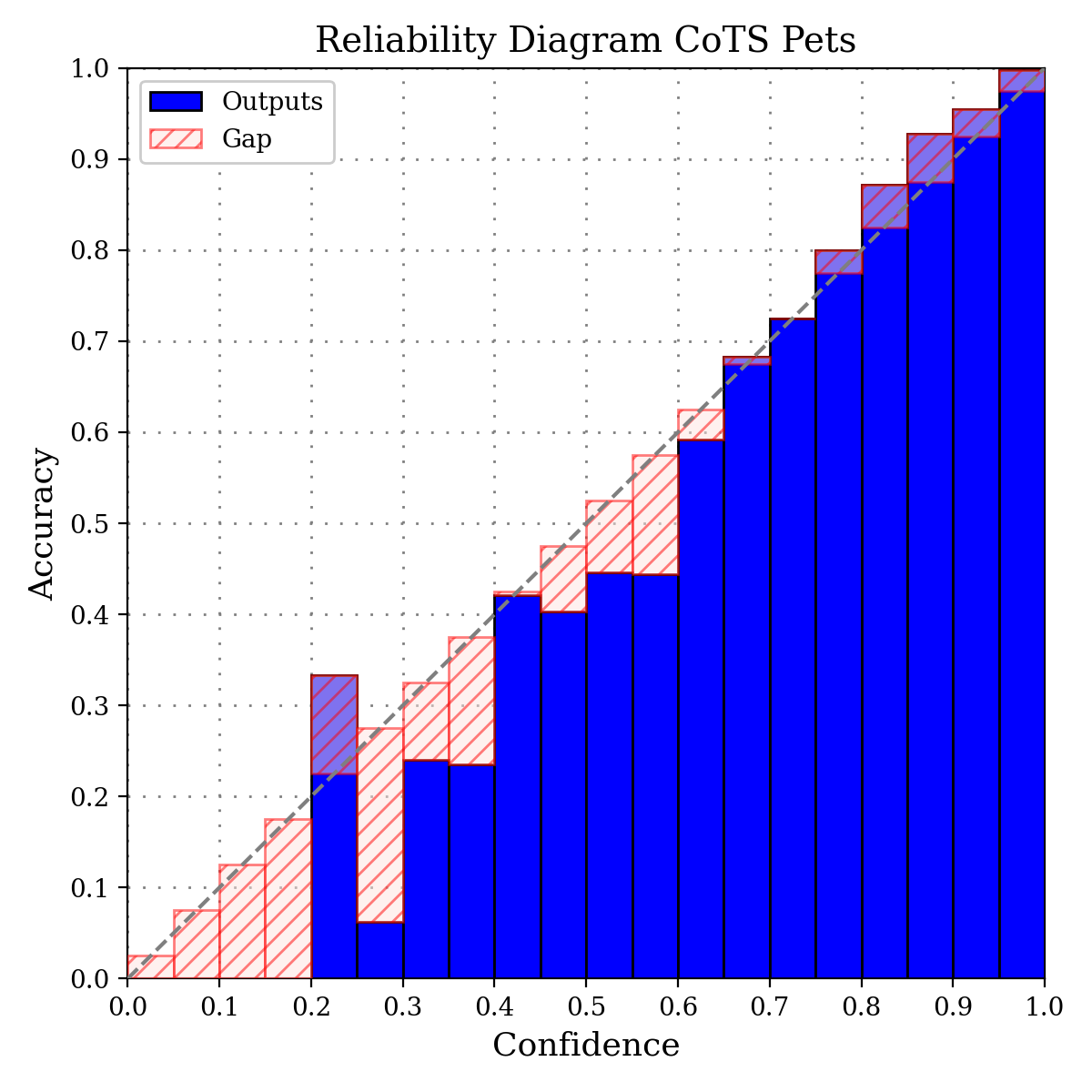}
        \caption{\cots: Pets}
        \label{fig:cots_pets}
    \end{subfigure}
    \hfill
    \begin{subfigure}[t]{0.21\textwidth}
        \centering
        \includegraphics[width=\linewidth]{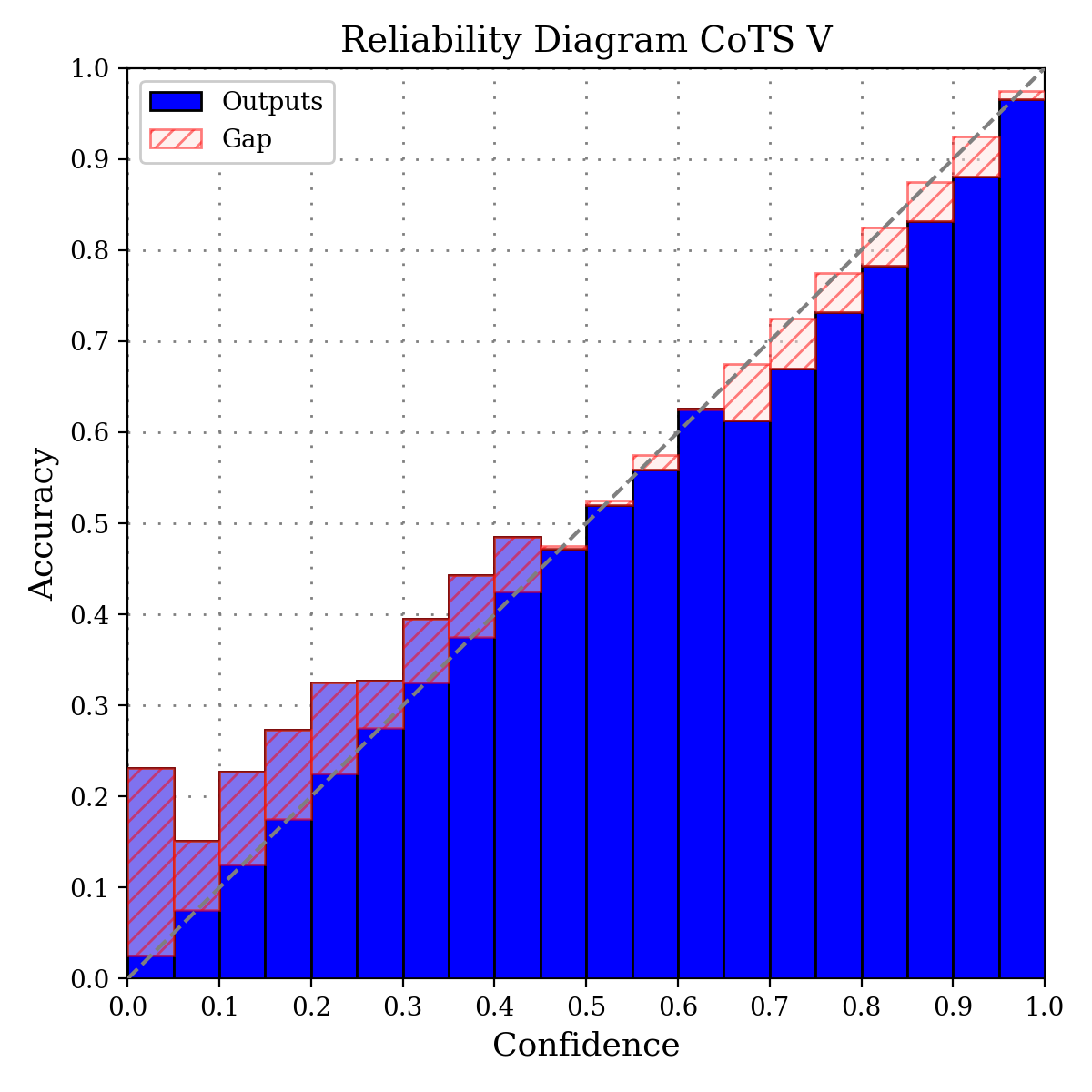}
        \caption{\cots: IN-V}
        \label{fig:cots_v}
    \end{subfigure}

    \begin{subfigure}[t]{0.21\textwidth}
        \centering
        \includegraphics[width=\linewidth]{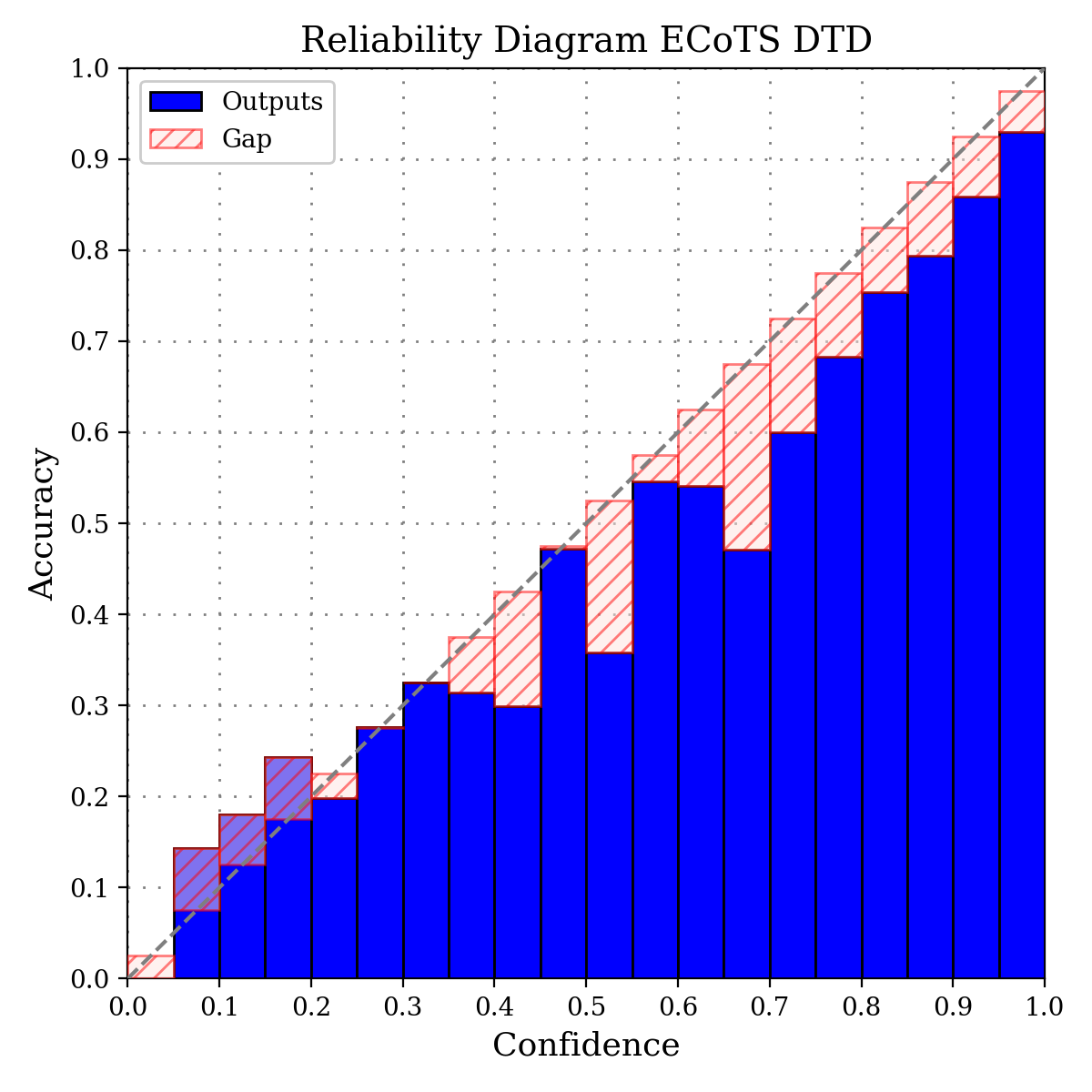}
        \caption{\ecots: DTD}
    \end{subfigure}
    \hfill
    \begin{subfigure}[t]{0.21\textwidth}
        \centering
        \includegraphics[width=\linewidth]{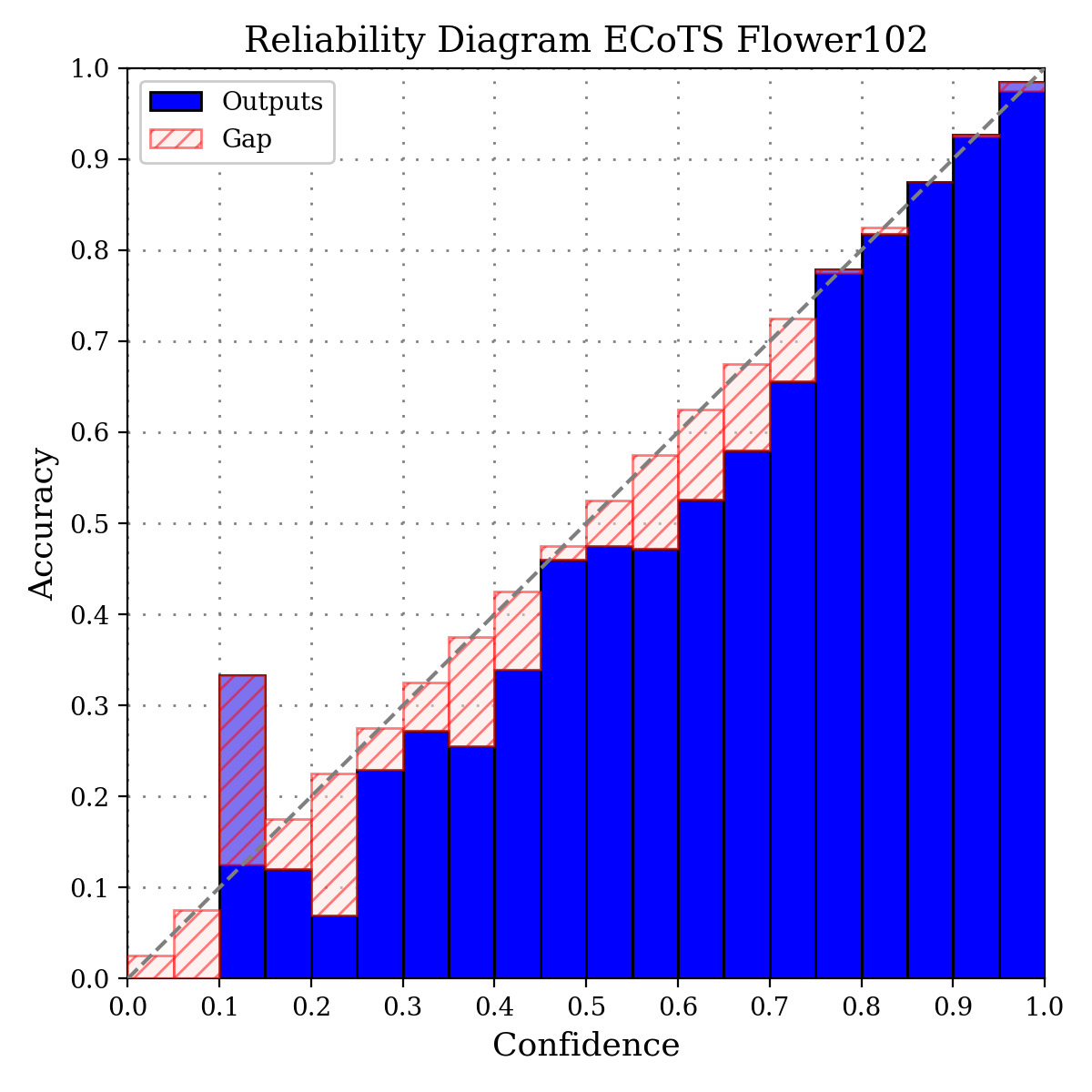}
        \caption{\ecots: Flowers}
    \end{subfigure}
    \hfill
    \begin{subfigure}[t]{0.21\textwidth}
        \centering
        \includegraphics[width=\linewidth]{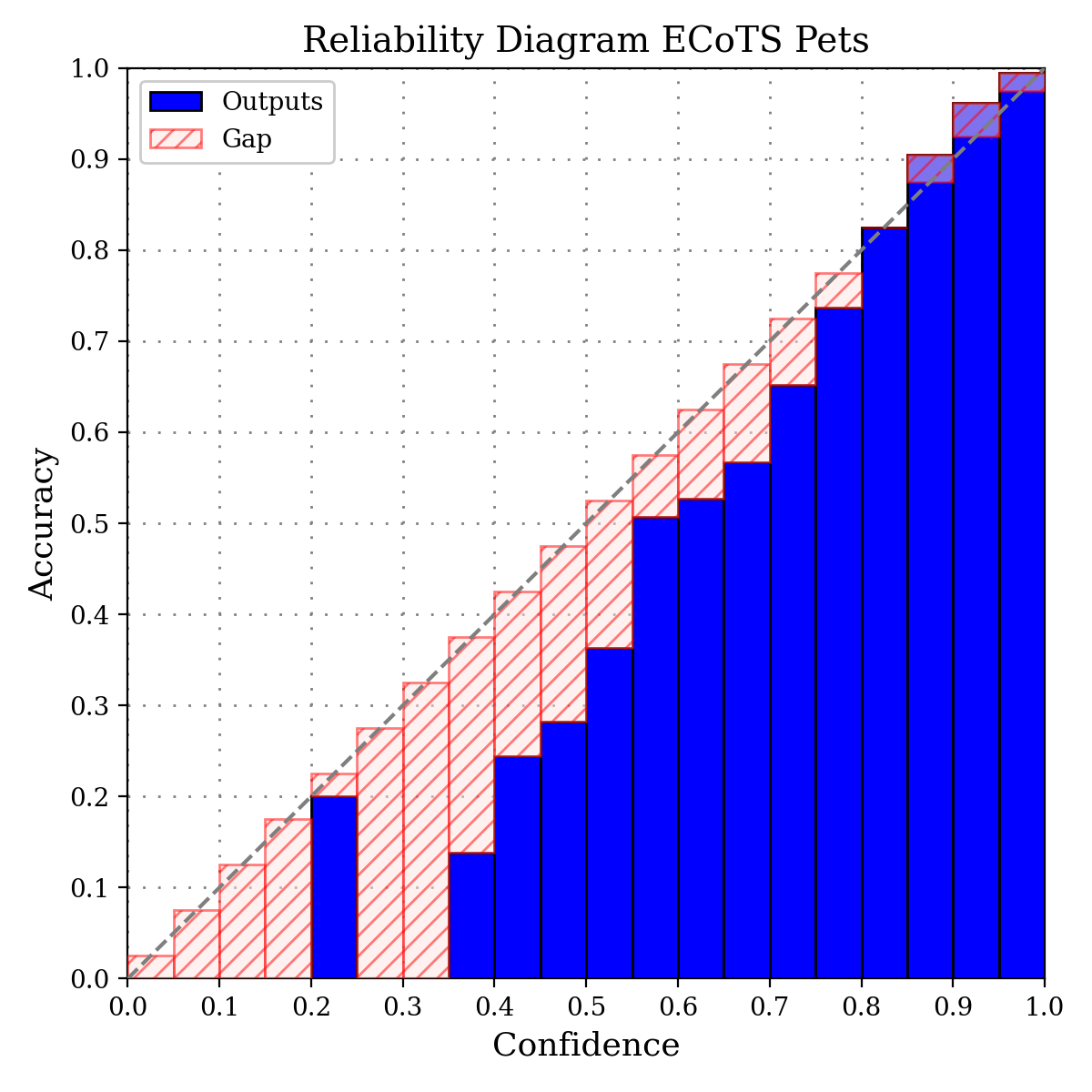}
        \caption{\ecots: Pets}
    \end{subfigure}
    \hfill
    \begin{subfigure}[t]{0.21\textwidth}
        \centering
        \includegraphics[width=\linewidth]{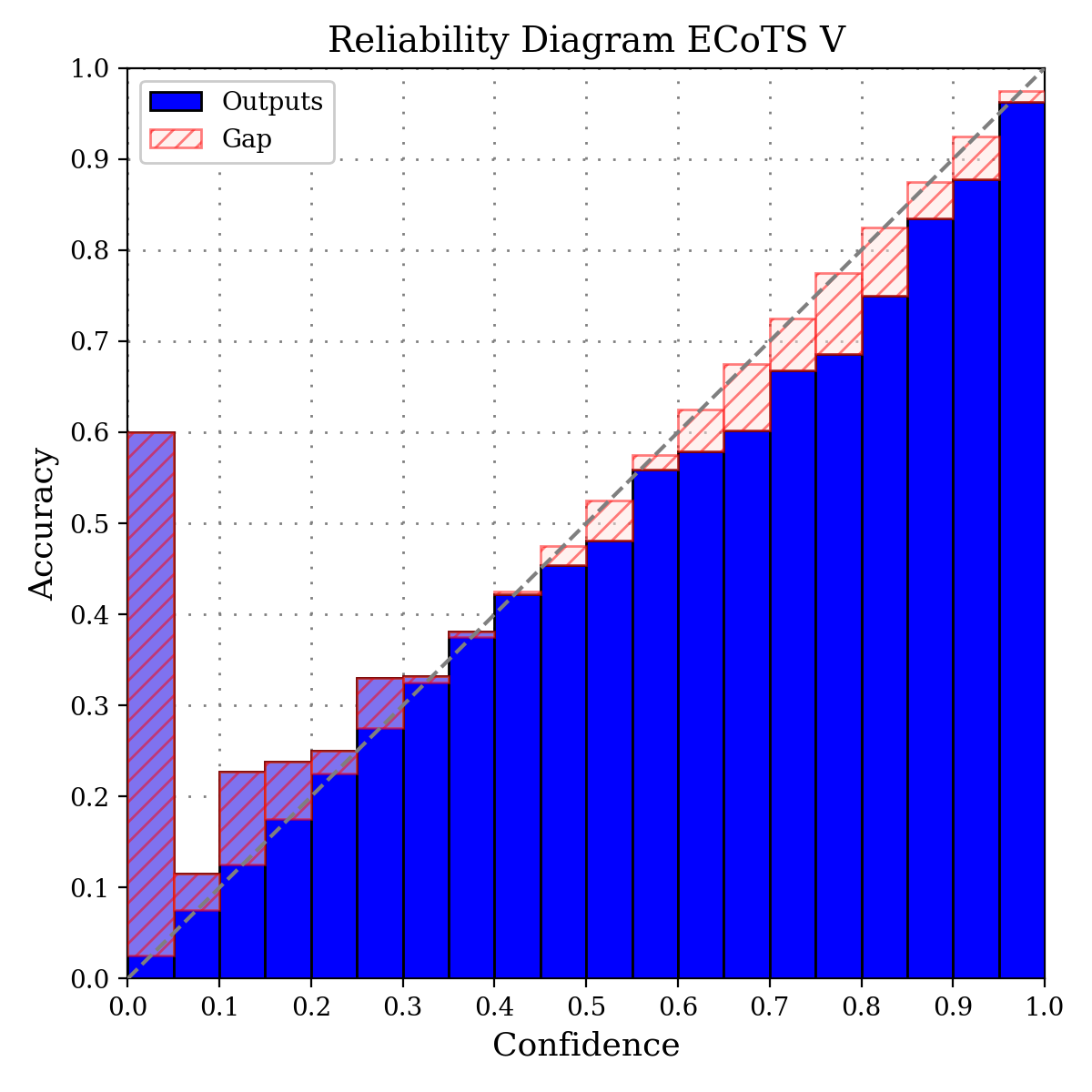}
        \caption{\ecots: IN-V}
    \end{subfigure}

    \caption{
    Reliability diagrams of SoC~\cite{fillioux2026soc}, O-TPT~\cite{sharifdeen2025otpt}, A-TPT~\cite{ahamed2026atpt}, \cots~, and \ecots~on DTD, Flowers, Pets, and ImageNet-V using ViT-B/16.
    }
    \label{fig:reliability_diagrams_4x4}
\end{figure*}

As shown in Fig.~\ref{fig:reliability_diagrams_4x4}, \cots~and \ ecots~achieve small accuracy-confidence gaps across datasets, indicating reliable calibration.

\end{document}